\documentclass[sigconf,screen]{acmart}

\AtBeginDocument{%
  \providecommand\BibTeX{{%
    \normalfont B\kern-0.5em{\scshape i\kern-0.25em b}\kern-0.8em\TeX}}}

\acmConference[]{}{}{}
\setcopyright{none}
\copyrightyear{2021}
\acmYear{2021}

\usepackage[utf8]{inputenc} 
\usepackage[T1]{fontenc}    
\usepackage{hyperref}       
\usepackage{url}            

\usepackage{booktabs}       
\usepackage{amsfonts}       
\usepackage{nicefrac}       
\usepackage{microtype}      
\usepackage{algorithmic}
\usepackage{multirow}
\usepackage{amsmath}
\usepackage{graphicx}
\usepackage{subcaption}
\usepackage{tabularx}
\usepackage{colortbl}
\usepackage{epstopdf}
\usepackage[linesnumbered,noend]{algorithm2e}

\usepackage{enumitem}

\usepackage{cleveref}

\usepackage{color}

\newcommand{\baselines}{seven\xspace}
\newcommand{\baselinesPlusOurs}{eight\xspace}

\newcommand{\vbarmc}[2]{\multicolumn{#1}{|c}{#2}}
\newcommand{\functionname}[1]{\texttt{#1}}
\newcommand{\feature}[1]{\textit{#1}}
\newcommand{\technique}[1]{\textit{#1}}
\newcommand{\var}[1]{\textsf{#1}}
\newcommand{\threshold}{\lambda}
\newcommand{\Themis}{\texttt{Themis}\xspace}

\newcommand{\RQ}[1]{\textbf{RQ#1}\xspace}

\graphicspath{{plots/}}

\author{Sahil Verma}
\affiliation{
        \institution{Department of Computer Science and Engineering\\ University of Washington\\ Seattle, USA}
}
\email{vsahil@cs.washington.edu}

\author{Michael Ernst}
\affiliation{
  \institution{Department of Computer Science and Engineering\\ University of Washington\\ Seattle, USA}
}
\email{mernst@cs.washington.edu}

\author{Rene Just}
\affiliation{
  \institution{Department of Computer Science and Engineering\\ University of Washington\\ Seattle, USA}
}
\email{rjust@cs.washington.edu}

\begin{document}
\title{Removing biased data to improve fairness and accuracy}

\setlength{\leftmargini}{.75\leftmargini}
\setlength{\leftmarginii}{.75\leftmarginii}
\setlength{\leftmarginiii}{.75\leftmarginiii}

\begin{abstract}
  Machine learning systems are often trained using data collected from historical decisions. 
  If past decisions were biased, then automated systems that learn from
  historical data will also be biased. 
  We propose a black-box approach to identify and remove biased training data. 
  Machine learning models trained on such debiased
  data (a subset of the original training data) have low individual
  discrimination, often 0\%.
  These models also have greater accuracy and lower statistical disparity than models trained on the full historical data. 
  We evaluated our methodology in experiments using 6 real-world datasets. 
  Our approach outperformed \baselines previous approaches in terms of individual discrimination and accuracy. 
\end{abstract}

\maketitle

\section{Introduction}
\label{sec:intro}

Automated decisions can be faster, cheaper, and less subjective than manual ones.
In order to automate decisions, an organization can train a machine learning model on historical decisions or other manually-labeled data and use it to make decisions in the future. 
However, if the training data is biased, that bias will be reflected in
the model's decisions~\cite{garbageinout,biased_data_1}. 
This problem has even been noted by the US White House~\cite{white_house}.

Bias in models can have far-reaching societal consequences, like worsening wealth inequality~\cite{biased-loan1}, 
difference in employment rate across genders~\cite{biased-hiring1,biased-hiring2}, 
and difference in incarceration rates across races~\cite{unfair_arrest_2}. 
Discrimination has been reported in machine-produced decisions for
real-life scenarios like parole~\cite{propublica-main}, credit
cards~\cite{criticism_1}, hiring~\cite{biased-hiring1}, and predictive
policing~\cite{NYC_policing}.
These problems are exacerbated by humans' unwarranted confidence in
machine-produced decisions: people generally deem such decisions
fair~\cite{weakened_appeal}.
a
Data bias happens mostly due to two phenomena: \emph{label bias} and \emph{selection bias}. 
\emph{Label bias} occurs when the training labels (which are mostly
generated manually) are afflicted by human bias. 
For example, loan applications and job applications from minority
communities have been more frequently
denied~\cite{Steil:2017-redlining,redliningwiki,redlining-thinkprogress}.
Training on historical data would perpetuate that injustice.
\emph{Selection bias} occurs when
selecting samples from a demographic group inadvertently introduces
undesired correlations between the features pertaining to that demographic
group and the training labels~\cite{Blum2019RecoveringFB,ImageNet:2020,trade_off_2019}, 
e.g. in the selected subsample for a group, most of their loan requests were denied. 
The propagation of bias has raised significant concerns related to use of
machine learning models in critical decision making like the ones mentioned above. 


Bias in machine learning systems is undesirable
because it can produce unfair decisions~\cite{biased-hiring1,biased-loan3},
because biased decisions are less effective and less profitable~\cite{less_effective1,less_effective2}, and
because bias attracts lawsuits and widespread criticism~\cite{criticism_1,criticism_2}. 
Biased decisions can be challenged on the basis of disparate treatment and disparate impact laws in the US~\cite{Barocas2016BigDD,jolly-ryan-have,Selbst2017DisparateII,Skeem2016RISKRA,Winrow2010TheDB}. 
Similar laws exist in other countries~\cite{australia_act}. 
Laws define \emph{sensitive attributes} (e.g., race, sex, religion) that are illegal to use as the basis of any decision. 


We use both the common statistical disparity, and also \emph{individual discrimination} as metrics to measure discrimination, in particular for machine learning classifiers. 
(Other definitions of fairness exist; \cref{sec:related} justifies our choice.) 
A model is individually fair~\cite{dwork_fairness_2011} if it yields
similar predictions for similar individuals. 
A \emph{similar pair} consists of two individuals $\langle X, Y \rangle$ such that
$X \ne Y$ and $X$ is similar to $Y$. 
Similarity among the individuals and among the predictions is measured by defining a distance function in the input space and the output space, respectively. 
A \emph{similar pair} of individuals are two individuals with distance lower than a specified threshold. 
Ideally, for ensuring fairness, the input space similarity metric should not take sensitive features into
account, so two individuals who differ only in the sensitive features should
always be similar.
A \emph{discriminatory pair} is a similar pair such that the model makes
dissimilar predictions for the two similar individuals.
The \emph{individual discrimination} of a model is the proportion of similar pairs of individuals that
receive dissimilar predictions, i.e., the percentage of discriminatory pairs~\cite{galhotra_fairness_2017,Udeshi-testing:2018,Aggarwal-testing:2019}. 
An estimate of individual discrimination uses a pool of similar
pairs, which may be synthetically generated or sampled from the training data.
To be fair, a model's individual discrimination should be zero or close to it. 


Previous approaches to reducing discrimination (see \cref{sec:related}) modify the training data's features or
labels (pre-processing), or add fairness regularizers while training
(in-processing), or perform post-hoc corrections in the model's predictions
(post-processing).



Our approach is a pre-processing approach that identifies and removes biased datapoints from the training data, leaving all other datapoints unchanged. 
We conjectre that some training data is more biased than the rest and consequently has more influence on the predictions of a learned model. 
Influence functions~\cite{koh_understanding_2017} measure the influence of
a training datapoint on a particular prediction.
Our approach
(1) generates discriminatory pairs (similar individuals who receive dissimilar predictions) and
(2) uses influence functions to sort
the training datapoints in order of most to least influential for
the dissimilar predictions received by the discriminatory pairs. 
We hypothesize that the datapoints that are most responsible for the
dissimilar predictions are the most biased datapoints. 
Removing the most influential datapoints yields a \emph{debiased dataset}. 
A model trained on the debiased dataset is fairer with less 
individual discrimination --- often 0\% --- than a model trained on the full
dataset.

Our approach works for black-box classification models (for which only \functionname{train} and \functionname{predict} functions are available)~\cite{blackbox1,blackbox2}, and therefore proprietary models can be debiased as long as their training data is accessible. 
We performed 8 experiments using 6 real-world datasets (some datasets contain more than one sensitive attribute). 
Our approach outperforms \baselines previous approaches in terms of
individual discrimination and accuracy and is near the average in terms of
statistical disparity.


In summary, our contributions are: 
\begin{itemize}[noitemsep,topsep=-1pt]
 \item We propose a novel black-box approach for improving individual fairness: identify and remove biased decisions in historical data. 
 \item In two sets of experiments using 6 real-world datasets, the classifiers
 trained on debiased datasets exhibit nearly 0\% individual discrimination. Our
 approach outperforms \baselines previous approaches in terms of
 individual discrimination and accuracy, and it always improves (reduces)
 statistical disparity.
 \item To the best of our knowledge, we are the first to empirically demonstrate 
 an increase in test accuracy (better generalization) in a supervised learning setup with real datasets
 while reducing discrimination, compared to the case when no fairness technique 
 is used~\cite{trade_off_2019,Blum2019RecoveringFB}. 
 \item Our implementation and experimental scripts are open source. 
\end{itemize}



\section{Motivating Example}
\label{sec:example}


\begin{table}
  \caption{A hypothetical dataset of past loan decisions. 
  The second datapoint is a biased decision because a black person in high range income was 
  denied a loan (0), whereas all white people with high range income were given a loan (1). 
  }
  \label{tab:biased-decision}
  \vspace{-10pt}
  \centering
  \begin{tabular}{cccrr}
  \toprule
  Id & Income & Wealth & Race & Decision \\
  \midrule
  \#1 & 1.0 & 0.1 & White & 1 \\
  \rowcolor{lightgray}
  \#2 & 0.9 & 0.7 & Black & 0 \\
  \#3 & 0.8 & 0.3 & White & 1 \\
  \#4 & 0.1 & 0.7 & Black & 0 \\
  \#5 & 0.1 & 0.5 & White & 0 \\
  \#6 & 0.5 & 0.9 & Black & 0 \\
  \#7 & 1.0 & 0.8 & Black & 1 \\
  \bottomrule
  \end{tabular}
 \end{table}
 

\setlength{\belowcaptionskip}{-8pt}
\begin{figure*}
  \includegraphics[width=\textwidth]{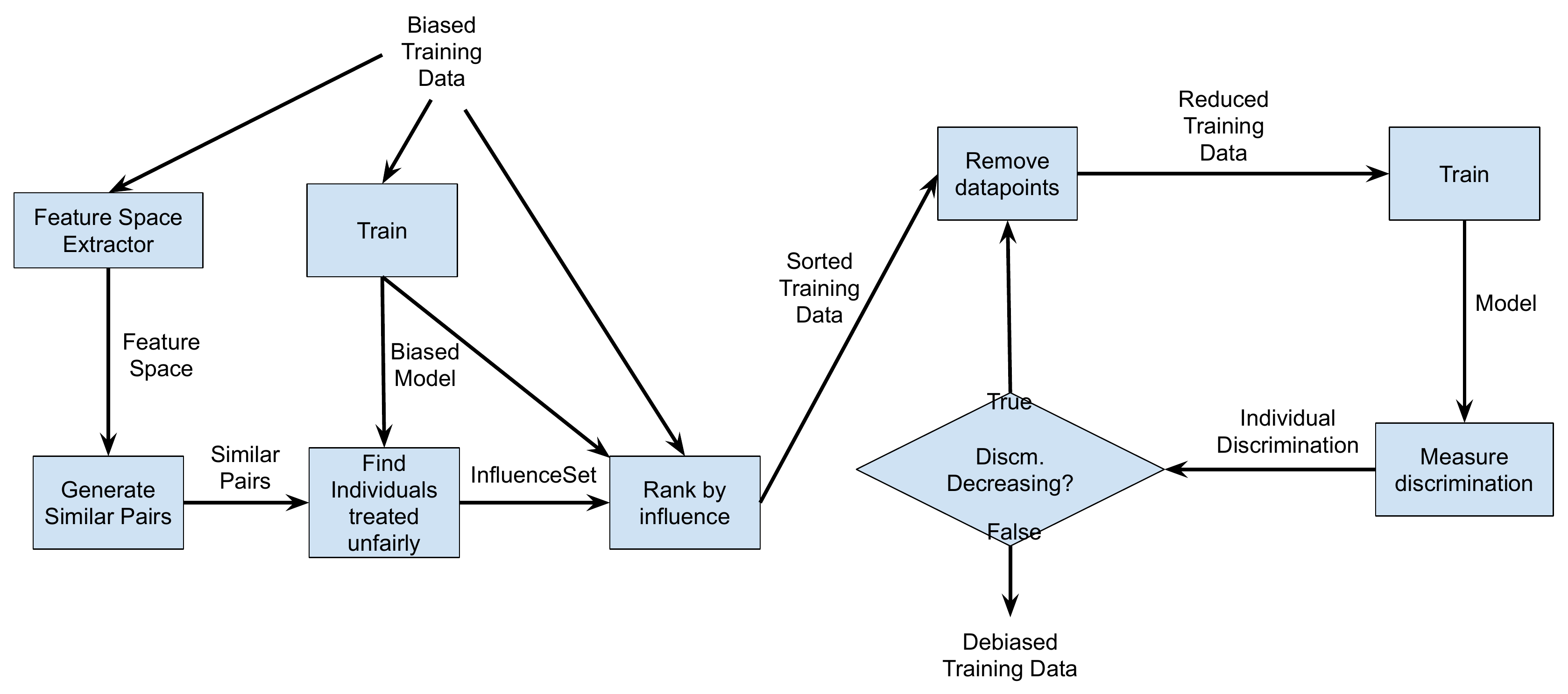}
  \vspace{-20pt}
  \caption{Flowchart with the steps in our approach. The left portion shows the steps in~\Cref{alg:rankpoints} 
  and the right portion shows the steps in~\Cref{alg:debiaseddata}. 
  The output of the algorithm is a debiased training data which can be used to 
  train a debiased model. \looseness=-1}
 \label{fig:algo-plot}
\end{figure*}

Suppose that a bank wishes to automate the process of loan approval.
The bank could train a machine learning model on historical loan decisions,
then use the model to improve speed, reduce costs, and reduce subjectivity.
The financial sector is a prominent user of machine learning~\cite{finance_1,finance_2}. 

\Cref{tab:biased-decision} shows a hypothetical dataset consisting of historical
loan decisions. The bank collects 3 features from each applicant: \feature{income}, \feature{wealth}, and \feature{race}. 
As shown in the table, \feature{income} and \feature{wealth} are numeric features with values lying between 0 and 1, after normalization. 
\feature{Race} is a binary feature with `white' and `black' as the two possible values. 
The outcome is also binary with `1' indicating that a loan was offered and `0' indicating a denial. 

Loan approvals should ideally not depend on the applicant's
race, but past loan approvers might have been consciously or unconsciously biased. 
In fact, that was the case at this bank: \#2 was a biased decision in which a black person with income in high
range (0.9) was denied a loan, whereas all white people with high range income ($>$ 0.66) were given a loan. 
A classifier trained on a dataset that contains biased datapoints is
likely to exhibit individual discrimination. 

The bank would like to train a classifier using the unbiased parts of the
historical data. Most of the decisions were probably sound (otherwise, the bank would have been out-competed by its rivals). 
It would be too expensive to filter biased decisions or create a new dataset manually, 
and those manual activities would also be prone to conscious and unconscious bias.

\subsection{Measuring individual discrimination}

Estimating a model's individual discrimination requires the distance
functions for the input and output space, and a pool of similar pairs.
We consider two individuals similar if their \feature{income} and \feature{wealth} are the same, regardless of their \feature{race}. 
We consider two decisions similar if they are the same decision.
We randomly generate 700 similar pairs (see \Cref{sec:algo} for details). 
This is 100 times as large as the training dataset; the value ``100'' is arbitrary.

We trained a model on the dataset of \cref{tab:biased-decision} and
evaluated it on the 1400 individuals who form the 700 pairs of similar individuals. 
The model predicted different outcomes for 26\% of the similar pairs (181 discriminatory pairs out of the 700 similar pairs).


\subsection{Finding biased datapoints}

Our goal is to find biased training datapoints, such that removing them
reduces the model's individual discrimination. 
This also improves the model's accuracy and statistical parity, which is discussed in~\Cref{sec:eval}. 
Our approach contains three high-level steps:
\begin{enumerate}[leftmargin=0pt]
\item
  Find unfair decisions made by a learned model. In particular, first
  generate a pool of pairs of similar
  individuals and find the discriminating pairs within them.
  Then, for each discriminating pair, heuristically determine which of the two
  individuals was not treated fairly.

\item
 Rank the training datapoints according to their contribution to the unfair
 decisions.
\item
 Remove some of the highest-ranked training datapoints and retrain a model
 with a lower individual discrimination.
\end{enumerate}

\paragraph{Identify unfair decisions}
There are 181 discriminatory pairs --- pairs of similar individuals with dissimilar outcomes.
One individual from each pair has been treated unfairly.
The biased treatment might be in their favor (getting an undeserved loan) or against them (denied a loan they deserved).
%
%
Our heuristic is that the individual whose classification confidence (i.e., the
probability with which an individual is classified) is lower is the one
who was treated unfairly. 
Note that these are the individuals in the generated pool of similar individuals, not in the training dataset. 


\paragraph{Identify training datapoints responsible for unfair decisions}
We use influence functions~\cite{koh_understanding_2017} to find the training datapoints 
that were most responsible for producing different predictions for the discriminatory pairs. 
An influence function sorts the training datapoints of a model in order of most to 
least responsible for a single prediction from the model. 
An influence function can also sort the datapoints for a set of predictions
(the 181 unfair decisions, in our case), by measuring the average
influence over all the relevant predictions. 

We hypothesize that influence functions rank the training datapoints in order of 
most to least biased datapoints. 
Note that, the discriminatory pairs are only used to measure discrimination and
to sort datapoints in the original training dataset; the discriminatory pairs
themselves never occur in the training data.

For the dataset of \cref{tab:biased-decision}, the ranked datapoints are: \#2, \#7, \#3, \#1, \#4, \#6, \#5. 
According to this ranking, \#2 is the most biased decision, followed by \#7 and so on. 

\paragraph{Remove biased datapoints}

When \#2 is removed, and the model is retrained, it results in only 1 discriminatory pair 
out of the same 700 similar pairs of individuals, i.e., 0.14\% remaining individual discrimination. 
When both \#2 and \#7 are removed, the discrimination is 16\%. 
Therefore, removing only \#2 is the local minimum for discrimination. 
Our algorithm returns the dataset with \#2 removed. 
This reduced dataset is called the ``debiased dataset''. 
Note that our approach removes biased datapoints from the training data. 
The similar pairs are synthetically generated and are only used to estimate individual 
discrimination and identify biased training points. 
The similar pairs can not be removed since they don't occur in the training dataset. 

The bank should train its model on the fair decisions in the debiased dataset, 
rather than on the whole dataset. 
We choose to remove the biased datapoint and not modify its label because the bias 
could arise either due to labeling bias or due to selection bias. 
Note that our approach can remove a datapoint belonging to any demographic
group, not only the disadvantaged group as happened in this hypothetical dataset.

All the steps of our approach are shown in~\Cref{fig:algo-plot}.

Our approach requires access to sensitive attribute. 
\citet{removingSensitive1,removingSensitive2} have argued about the necessity of having access to 
the sensitive features in order to identify and address bias in automated models. 
After identifying and removing the biased decisions, the bank can train their model by 
omitting the sensitive feature(s) (\feature{race} in this case) to avoid disparate treatment~\cite{jolly-ryan-have}.

Our approach empowers the bank to make less biased decisions in the
future.
(As with any automated decision-making process, the bank should include
avenues for challenge and redress.)
This avoids violating anti-discrimination laws and is the morally right thing to do.



\section{Algorithm}
\label{sec:algo}


\SetKwProg{Fn}{Function}{}{}
\RestyleAlgo{algoruled}
\newcommand\mycommfont[1]{\small\ttfamily\textcolor{blue}{#1}}
\SetCommentSty{mycommfont}

\newcommand{\xAlCapSty}[1]{\small\sffamily\bfseries\MakeUppercase{#1}}
\SetAlCapSty{xAlCapSty}


\newcommand\mynlfont[1]{\scriptsize\sffamily{#1}}
\SetNlSty{mynlfont}{}{} 

\SetSideCommentLeft


\begin{algorithm}[h]
 \DontPrintSemicolon 
 \SetKwInOut{KwIn}{Input}
 \SetKwInOut{KwOut}{Output}
 \SetKwData{Ma}{M}
 \SetKwData{Sa}{S}
 \SetKwData{Da}{D}
 \SetKwData{Size}{Size}
 \SetKwData{SD}{SD}
 \KwIn {Training dataset \Da,
   Sensitive attribute \Sa,
   Binary classification model \Ma trained on \Da,
   Input space similarity threshold $\threshold$,
} 
 \KwOut {\Da sorted in decreasing order of contribution to bias}
 
 \caption{\label{alg:rankpoints}Sort the training data in decreasing order of bias}
 \SetKwFunction{GenerateSimilarPairs}{GenerateSimilarPairs}
 \SetKwFunction{themis}{\Themis}
 \SetKwData{PSI}{PSI}
 \SetKwData{DP}{DP}
 
 \SetKwData{Discm}{Discm}
 \SetKwData{PrevDiscm}{PrevDiscm}
 \SetKwData{InfluenceSet}{InfluenceSet}
 \SetKwData{SimilarPairs}{SimilarPairs}
 \SetKwFunction{InfluenceRanker}{RankByInfluence}
 \SetKwFunction{FairModelPrediction}{FairModelPrediction}
 \SetKwFunction{SortDataset}{SortDataset}
 \SetKwFunction{Split}{Split}
 \SetKwFunction{Train}{Train}
 \SetKwFunction{Loss}{Loss}
 \SetKwFunction{Sample}{Sample}
 \SetKwFunction{Probability}{Probability}
 \SetKwFunction{SimilarIndividual}{SimilarIndividual}
 \SetKw{break}{break}
 \SetKwData{PMa}{M'}
 \SetKwFunction{MinimumRemainingDiscrimination}{MinimumRemainingDiscrimination}
 
 \Fn{$\mathit{\SortDataset(\Da, \Sa, \Ma, \threshold})$ }{ 
  $\mathit{\PSI \leftarrow \GenerateSimilarPairs(\Da, \Sa, \threshold)}$ \label{line:call-GenerateSimilarPairs} \\
  \tcp{Discriminating pairs: similar individuals who received dissimilar outcomes} 
  $\mathit{\DP \leftarrow \{\langle i_1, i_2 \rangle \in \PSI : \Ma(i_1) \neq \Ma(i_2)\}}$ \label{line:discm1}\\
  \tcp{Individuals that M discriminates against} 
  $\mathit{\InfluenceSet \leftarrow \{\}}$ \label{line:influenceset-start}\\
  \For{$\mathit{\langle i_1, i_2 \rangle \in \DP}$} {
  \tcp{Add to \InfluenceSet the individual with lower classification confidence}
  \eIf{$\mathit{\Probability(i_1) < \Probability(i_2)}$ \label{line:influ1}} 
    {$\mathit{\InfluenceSet \leftarrow \InfluenceSet \cup \langle i_1, \Ma(i_1)\rangle}$} 
    {$\mathit{\InfluenceSet \leftarrow \InfluenceSet \cup \langle i_2, \Ma(i_2)\rangle}$} \label{line:influ2} 
    }  \label{line:influenceset-end}
  \Return \InfluenceRanker(\InfluenceSet, \Da) \label{line:influencer}
 }
 \tcp{Randomly generates pairs of similar individuals}

 \Fn{$\mathit{\GenerateSimilarPairs(\Da, \Sa, \threshold)}$} {\label{line:generatesimilar}
 $\mathit{\SimilarPairs \leftarrow \{\}}$ \\ 
 \For{$\mathit{i\leftarrow 1 \ \KwTo\ 100 * \vert D\vert}$}{
 $\mathit{A1} \leftarrow \Sample(\Da)$ \tcp{Sample from feature space}
 $\mathit{A2} \leftarrow \SimilarIndividual(\Da, A1, \Sa, \threshold)$
   \tcp{Generate A2 s.t. distance(A1, A2) $< \threshold$} \label{line:findsimilar}
 $\mathit{\SimilarPairs \leftarrow \SimilarPairs \cup \langle A1, A2\rangle}$\\
 }
 \Return $\SimilarPairs$
 }

 \tcp{Ranks datapoints in \Da
 responsible for discrimination, sorted in decreasing
 order of influence}
 \Fn{$\mathit{\InfluenceRanker(\InfluenceSet, \Da)}$ }{ 
     \tcp{See~\cite{koh_understanding_2017} for implementation}
 }
\end{algorithm}

\begin{algorithm}[h]
 \DontPrintSemicolon 
 \SetKwInOut{KwIn}{Input}
 \SetKwInOut{KwOut}{Output}
 \SetKwData{Ma}{M}
 \SetKwData{Sa}{S}
 \SetKwData{Da}{D}
 \SetKwData{Data}{Data}
 \SetKwData{SD}{SD}
 \SetKwData{Size}{Size}
 \SetKwData{Na}{N}
 \SetKwData{PMa}{M'}
 \SetKwFunction{Train}{Train}
 \KwIn {Training dataset \Da,
   Sensitive attribute \Sa,
   Input space threshold $\threshold$,
   \Train function}
 \KwOut{Debiased version of \Da, which is a subset of \Da}
 \caption{\label{alg:debiaseddata}Produce a debiased dataset}
 \SetKwFunction{GenerateSimilarPairs}{GenerateSimilarPairs}
 \SetKwFunction{themis}{\Themis}
 \SetKwData{PSI}{PSI}
 \SetKwData{DP}{DP}
 \SetKwData{Discm}{Discrim}
 \SetKwData{LeastDiscm}{LeastDiscrim}
 \SetKwData{InfluenceSet}{InfluenceSet}
 \SetKwFunction{InfluenceRanker}{InfluenceRanker}
 \SetKwFunction{FairModelPrediction}{FairModelPrediction}
 \SetKwFunction{SortDataset}{SortDataset}
 \SetKwFunction{DropFirst}{DropFirst}
 
 \SetKwFunction{Loss}{Loss}
 \SetKw{break}{break}
 \SetKw{by}{by}
 \SetKwFunction{DebiasedData}{DebiasData}
 \SetKwFunction{DiscmTest}{EstimateDiscrim}
 \SetKwFunction{LeastDiscriminationModel}{LeastDiscriminationModel}

 \Fn{$\mathit{\DebiasedData(\Da, \Sa, \threshold, \Train)}$}{
 \tcp{Model trained on full dataset}
 $\mathit{\Ma_\mathrm{full} \leftarrow \Train(\Da)}$\\
 \tcp{Sorted biased datapoints (see \cref{alg:rankpoints})} 
 $\mathit{\SD \leftarrow \SortDataset(\Da, \Sa, \Ma_\mathrm{full}, \threshold)}$ \label{line:callsortdataset}\\
 $\LeastDiscm \leftarrow \infty$ \\
 \For{$\mathit{i \leftarrow}$ 0 \KwTo 100} {
 \tcp{Model trained on remaining data}
 $\mathit{\Ma \leftarrow \Train(\DropFirst(\SD, i))}$ \hfill \tcp{\DropFirst removes the first i\% from input dataset \SD} \label{line:drop}
 $\mathit{\Discm \leftarrow \DiscmTest(\Ma, \Da, \Sa, \threshold)}$ \label{line:remaindiscm} \\
 \uIf{$\mathit{\Discm \ge \LeastDiscm}$}
 {\Return $\DropFirst(\SD, i-1)$ \label{line:debiased}}
 
 $\mathit{\LeastDiscm \leftarrow \Discm }$
 }
 
 }

 
 \tcp{Estimates the individual discrimination of model \Ma}
 \Fn{$\mathit{\DiscmTest(\Ma, \Da, \Sa, \threshold)}$ }{ 
 $\PSI \leftarrow \GenerateSimilarPairs(\Da, \Sa, \threshold)$ \\
 \tcp{Discriminating pairs: similar individuals who received dissimilar outcomes}
 $\DP \leftarrow \{\langle i_1, i_2 \rangle \in \PSI : \Ma(i_1) \neq \Ma(i_2)\}$\\
 \Return $|\DP| \, / \, |\PSI|$
 }
\end{algorithm}


\emph{\Cref{alg:rankpoints}} sorts a dataset in the order of most biased to least biased
datapoint. It has four main parts. 

First, \cref{line:call-GenerateSimilarPairs} generates a pool \textsf{PSI} of pairs of similar individuals using \functionname{GenerateSimilarPairs} (\cref{line:generatesimilar}). 
We arbitrarily choose the size of the pool to be 100 times the size of the dataset.
A larger pool leads to a better estimate of individual discrimination.
The individuals and their similar counterparts are automatically generated,
sampling uniformly at random from the feature space that is defined by the original training dataset.
For a randomly sampled individual \var{A1}, \functionname{SimilarIndividual}
(\cref{line:findsimilar}) generates a similar individual \var{A2} whose
distance from \var{A1} is less than the specified threshold $\threshold$,
which is a user-provided parameter that determines whether two
individuals are similar. For the input space similarity condition
in~\Cref{sec:example}, $\threshold$ is 0. 
For example, random sampling for \feature{income}, \feature{wealth}, and \feature{race} in the hypothetical dataset could generate an individual \var{A1} with the feature vector $\langle 0.9, 0.5, White \rangle$.
According to the input space similarity condition, a similar individual for \var{A1} must have the same \feature{income} and \feature{wealth} (these are the non-sensitive features). 
While generating similar individuals, we enforce changing the value of the sensitive attribute (\feature{race} in this case) to generate similar individuals in different demographic groups. 
Therefore, \var{A2} has the feature vector $\langle 0.9, 0.5, Black \rangle$. 
These pairs of similar individuals allow us to estimate the individual discrimination of 
a model because we can compare their predictions, which should be similar as well.
Note that generating pairs of similar individuals based on a given $\threshold$,
as opposed to finding pairs of similar individuals in the training dataset,
allows us to reliably generate a large pool of pairs.

Second, \cref{line:discm1} determines the discriminatory pairs: the pairs of similar individuals 
who received dissimilar model predictions (a subset of the pool \textsf{PSI}). 
In the case of binary classification, different labels are considered dissimilar. 

Third, \crefrange{line:influenceset-start}{line:influenceset-end}
determine, for each discriminatory pair, which individual was misclassified due
to model bias.
Internally, a classifier computes, for each outcome class, the probability of an individual
belonging to that class, and the predicted outcome for that individual is the
class with the highest probability. Classification confidence refers to this
highest probability.  Our heuristic (\cref{line:influ1}) is that the individual
with the lower classification confidence is the one who was treated unfairly.


Fourth, \functionname{RankByInfluence} (\cref{line:influencer}) identifies the datapoints in the original (biased) training 
dataset \textsf{D} responsible for discrimination against these individuals. 
Given a trained model and a set of datapoints along with
their predictions from the model, \functionname{RankByInfluence} ranks
the training data of the model from most influential to least influential training datapoint responsible for those predictions. 
If the most influential datapoints are removed from the training data, and the model is 
retrained with the same model architecture, the probability of change in the 
prediction of the discriminatory pairs is highest.
We hypothesize that \functionname{RankByInfluence} returns the training
datapoints sorted in order of most to least bias, and our experiments
support this. 
We show all the four steps in the left portion of \Cref{fig:algo-plot}.

None of the above steps is dependent on the number
of output categories, so the algorithm is applicable to multi-class classifiers. 
The notion of similar predictions would need to be adjusted accordingly.\looseness=-1


\emph{\Cref{alg:debiaseddata}} first calls \functionname{SortDataset} to rank the datapoints in decreasing order of bias (\cref{line:callsortdataset}). 
It then iteratively removes a chunk of the most biased datapoints from the
sorted original training dataset (\cref{line:drop}), retrains the same model architecture on the remaining
datapoints, and estimates the individual discrimination of the retrained model (\cref{line:remaindiscm}). 
When the remaining discrimination in a retrained model
reaches a local minimum, it returns a debiased dataset by dropping the most biased datapoints from the original training data (\cref{line:debiased}). 
The size of each chunk can be adjusted as desired (it is 1/100 of the original training data in \cref{alg:debiaseddata}).

\section{Evaluation}
\label{sec:eval}


\begin{table*}[h!]
    \caption{Datasets used in the evaluation}
    \label{tab:datasets}
    \vspace{-10pt}
     \centering
    \resizebox{\textwidth}{!}{
     \begin{tabular}{ccccccc}
     \toprule
     Id & Dataset & Size & \# Numerical Attrs. & \# Categorical Attrs. & Sensitive Attr. (\var{S}) & Training label (binary) \\ 
     \midrule 
     D1 & Adult income~\cite{UCI-repo-adult}      & 45222  & 1  & 11  & Sex  & Income $\ge$ \$50K             \\ 
     D2 & Adult income~\cite{UCI-repo-adult}      & 43131  & 1  & 11  & Race & Income $\ge$ \$50K             \\ 
     D3 & German credit~\cite{UCI-repo-german}    & 1000   & 3  & 17  & Sex  & Credit worthiness              \\ 
     D4 & Student~\cite{UCI-repo-student}         & 649    & 4  & 28  & Sex  & Exam score $\ge$ 11            \\ 
     D5 & Recidivism~\cite{compas-data}           & 6150   & 7  & 3   & Race & Ground-truth recidivism        \\ 
     D6 & Recidivism~\cite{compas-data}           & 6150   & 7  & 3   & Race & Prediction of recidivism       \\ 
     D7 & Credit default~\cite{UCI-repo-default}  & 30000  & 14 & 9   & Sex  & Credit worthiness              \\ 
     D8 & Salary ~\cite{sensitive_removal_data}   & 52     & 2  & 3   & Sex  & Salary $\ge$ \$23719           \\ 
     \bottomrule
     \end{tabular}
    }
\end{table*}    


We conducted experiments to answer the following research questions:
\begin{description}[noitemsep,topsep=-1pt] 
 \item[\RQ{1}] Does our technique reduce individual discrimination?
 \item[\RQ{2}] Do previous techniques reduce individual discrimination?
 \item[\RQ{3}] How does our technique impact test accuracy? 
 \item[\RQ{4}] How do previous techniques impact test accuracy? 
 \item[\RQ{5}] How do the techniques compare in terms of statistical disparity?
 \item[\RQ{6}] How sensitive are the techniques to hyperparameter choices?

\end{description}


We compared our pre-processing technique against 7 other techniques:
a baseline model trained on the full training dataset (\technique{Full}),
five pre-processing techniques --- simple removal of the sensitive attribute (\technique{SR}),
Disparate Impact Removal (\technique{DIR})~\cite{feldman_certifying_2015} (used at the
highest fairness enforcing level, repair=1), Preferential Sampling
(\technique{PS})~\cite{kamiran_data_2012}, Massaging (\technique{MA})~\cite{kamiran_data_2012}, and
Learning Fair Representations (\technique{LFR})~\cite{zemel_learning} --- and one
in-processing technique --- Adversarial Debiasing
(\technique{AD})~\cite{Zhang2018MitigatingUB}.
The implementations for some of these techniques were taken from IBM AIF360~\cite{Bellamy2018AIF3}. 

We evaluated the techniques using six real-world datasets that are commonly
used in the fairness literature (see
\cref{tab:datasets}). 

For all experiments, the machine learning model architecture we used was a neural network with 2 hidden layers. 
We trained models with 240 different hyperparameter settings:
\begin{itemize}[noitemsep,topsep=-1pt,leftmargin=15pt]
 \item In the first hidden layer, the number of neurons is 16, 24, or 32. 
 \item In the second hidden layer, the number of neurons is 8 or 12. 
 \item Each model had two choices for batch sizes:
 the closest powers of 2 to the numbers obtained by dividing the dataset size by 10 and 20, respectively. For example, if the dataset size is 1000, the batchsizes are 64 and 128. 
 \item Each experiment had 20 choices for random permutations for the full dataset. The choice of random permutation affects the datapoints that form the training and test datasets. 
\end{itemize}

\subsection{Experimental methodology}

To create the models for our approach (to answer \RQ{1} and \RQ{3}), we executed the following methodology:

\begin{enumerate}[noitemsep,topsep=-1pt] 
 \item For each of the 240 choices of hyperparameters:
 \begin{enumerate}
 \item Split the dataset into the first 80\% training and last 20\%
 testing (without randomness, but depending on the data permutation, which is
 one of the hyperparameters). The dataset is normalized before usage. 
 \item Debias the training dataset using \Cref{alg:debiaseddata}; that is,
 remove some points from it. 
 \item Compute a ``debiased model'', which is trained on the debiased training dataset. 
 \end{enumerate}
 \item Let the ``unfair datapoints'' for a dataset be the union of the datapoints removed by all the 240 models: that is, any datapoint removed by any debiasing step.
\end{enumerate}

To create the models for other approaches (to answer \RQ{2} and \RQ{4}), we
ran each approach 240 times, once for each choice of
hyperparameters. 
The in-processing technique \technique{AD}~\cite{Zhang2018MitigatingUB} does not take hyperparameters other than data permutations, 
therefore we only repeated the process 20 times, once for each data permutation. 

To measure the performance of the models, we executed the following methodology
for each trained model:

\begin{enumerate}[noitemsep,topsep=-1pt,leftmargin=10pt]
 \item Measure the model's individual discrimination using the function
 \functionname{DiscmTest} of \Cref{alg:debiaseddata} (\RQ{1} and \RQ{2}).
 \item Measure the model's test accuracy on a debiased test set (\RQ{3} and \RQ{4}). 
 The debiased test set is computed by removing the unfair points from the test
 set, which is the last 20\% of the dataset. 
\end{enumerate}
\smallskip
 Note that when evaluating our technique, the points removed from the test set of 
 a model is not affected by that model itself, 
 but only by other models that have those test points in their training set.
 Thus, there is no leak between the training and test dataset.
 The fourth hyperparameter, random perturbation, affects the datapoints
 that form the training and test datapoints for a model.

We conducted two sets of experiments
(\cref{sec:experiment-threshhold-zero,sec:experiment-threshhold-nonzero}) with
different similarity conditions in the input space.

\subsubsection{Debiasing the test set}

Our evaluation methodology uses a debiased test set from which unfair
points have been removed.
The reason is that a user's goal is not to
obtain a model that performs well on the entire dataset that includes biased decisions, but a model that
performs well on \emph{fair decisions}. Our experimental results indicate that
our debiasing technique identifies fair decisions, so we use it for this
purpose. 

This experimental methodology addresses an observation made by~\citet{trade_off_2019}
that in most previous discrimination
mitigation approaches, there is a discrepancy between the algorithm and its evaluation. 
Previous authors have agreed on the existence of bias in the data and 
have devised algorithms to mitigate bias in the resulting data or classifier, but they
evaluated the accuracy of their approach on the original test set, which is potentially biased. 
Due to this discrepancy, most previous work suggests that a technique must
trade off fairness and
accuracy~\cite{berk2017convex,Chouldechova2016FairPW,CorbettDavies2017AlgorithmicDM,feldman_certifying_2015,Fish2016ACA,kamiran_data_2012,Kleinberg2016InherentTI,Menon-CF:2018,Zafar2017FairnessBD,calmon_optimized,Calders-Naive-2010,Kamishima:2012},
which \citet{trade_off_2019} refute.

Using our experimental methodology, the classifier debiased using our
approach usually has higher accuracy on the debiased test set than the
classifier trained on the full dataset (see \cref{sec:results1,sec:results2}).
We think of this phenomenon as a classifier with improved generalization, an intuition also shared by~\citet{berk2017convex}, who remark that fairness constraints might act as regularizers and improve generalization. 

Another advantage is that using the same test set for a particular set of hyperparameters provides an
apples-to-apples comparison of our technique with all the \baselines baselines. 
For the \technique{Full} baseline, the full training set is used, while the test set is still debiased.

\subsection{$\!$Experiments with input space threshold $\threshold$=0}
\label{sec:experiment-threshhold-zero}

In the first set of experiments we used the following distance functions with the definitions of individual discrimination.

\textbf{Input space similarity condition:}
We consider two individuals to be similar if they
are the same in all non-sensitive features.

\textbf{Output space similarity condition:}
We consider two outcomes to be similar if they are the same outcome.
Note that for all our experiments, the outcomes are binary.

\textbf{Generating similar individuals:}
For the given similarity condition in the input space, \functionname{Generate\-Similar\-Pairs} randomly generates
the first individual, and then flips its sensitive feature to generate an individual similar to it (similar procedure as used for the hypothetical dataset in \Cref{sec:example}). 
For example, the \var{Salary} dataset has features
\feature{sex}, \feature{rank}, \feature{age}, \feature{degree}, and \feature{experience},
out of which \feature{age} and \feature{experience} are numerical while the
others are categorical features. If the first randomly generated individual had
feature values $\langle$Male, Full, 35, Doctorate, 5$\rangle$, then the similar individual would
have feature values $\langle$Female, Full, 35, Doctorate, 5$\rangle$.
In this set of experiments, the number of pairs of similar individuals generated for each dataset was 
set to 100 times the size of the respective total dataset (e.g., 3,000,000
for the Credit default dataset). 

When generating individuals, there is no guarantee that the generated
individuals are representative of the actual population. For example,
consider a dataset whose features include gender and college alma mater.
It might generate a datapoint for a male who graduated from a women's
college. Such individuals are uncommon
(Timothy Boatwright of Wellesley College is one example). Therefore, a
dataset with many such individuals might or might not be useful in
determining whether there is discrimination against graduates of the
women's college. As another example, consider determining whether a
basketball coach has discriminated ethnically in selecting team members;
generated individuals might not be representative since Bolivian men have
an average height of 160cm (5'1"), and Bosnian men have an average height of 184cm (6').
Data selection (as opposed to generation) approaches can guarantee that the individuals are
characteristic, but they require accurate characterizations of, or large samples
from, the population (which we don't have access to). 
Even so, it may not be possible or easy to find
many similar pairs of individuals to compare. We acknowledge these
limitations. Future work should explore how to obtain similar pairs that
are characteristic of real-world populations.

\subsection{$\!\!$Experiments with input space threshold $\threshold$\relax$>$0}
\label{sec:experiment-threshhold-nonzero}
The second set of experiments used these distance functions.

\textbf{Input space similarity condition:}
We consider two individuals to be similar if, among the non-sensitive features, they have the same value for all categorical features and are within a 10\% range for all numerical features (after normalization). 

\textbf{Output space similarity condition:}
We consider two outcomes similar if they are the same outcome.

\textbf{Generating similar individuals:}
\functionname{Generate\-Similar\-Pairs} randomly generates the first individual \var{A1}. 
It then generates 2 
similar individuals for \var{A1} following the input space similarity condition. 
(Therefore, in this set of experiments, the size of the pool of similar pairs was equal to 200 times the size of the dataset.) 
A similar individual is generated by maintaining the same values for all categorical features, and random sampling within the $-10$\% to $+10$\% range for all numerical features. 







\begin{table*}
 \caption{Information about the model with the \emph{least remaining discrimination}, 
 among 240 hyperparameter settings when $\threshold=0$. All numbers are percentages.}
\vspace{-10pt}
 \label{tab:min-discm}
 \centering 
 \resizebox{\textwidth}{!}{%
 \setlength{\tabcolsep}{.24\tabcolsep}
 \begin{tabular}{l|llllllll|llllllll|llllllll}
 \toprule
 & \vbarmc{8}{Individual discrimination} & \vbarmc{8}{Test accuracy} & \vbarmc{8}{Statistical parity difference} \\
 Id 
 & \technique{Full} & \technique{SR} & \technique{DIR} & \technique{PS} & \technique{MA} & \technique{LFR} & \technique{AD} & \technique{Our}
 & \technique{Full} & \technique{SR} & \technique{DIR} & \technique{PS} & \technique{MA} & \technique{LFR} & \technique{AD} & \technique{Our} 
 & \technique{Full} & \technique{SR} & \technique{DIR} & \technique{PS} & \technique{MA} & \technique{LFR} & \technique{AD} & \technique{Our} \\
 \midrule
 D1 & 19.0 &  \textbf{0.0}  & 19.0 & 0.0064 & 5.7 & 0.096 & 12.0 &  \textbf{0.0}  
& 80 & 82 & 80 & 81 & 80 & 81 & 85 &  \textbf{92}  
& 29 & 13 & 28 & 8.7 & 3.3 & 3.6 &  \textbf{2.3}  & 7.8 \\
D2 & 11.0 &  \textbf{0.0}  & 11.0 & 0.38 & 6.0 & 0.0063 & 4.3 &  \textbf{0.0}  
& 83 & 85 & 84 & 84 & 83 & 84 & 87 &  \textbf{92}  
& 21 & 13 & 20 & 12 & 7 &  \textbf{0.33}  & 7.8 & 10 \\
D3 & 6.2 &  \textbf{0.0}  & 3.8 & 0.014 & 0.083 &  \textbf{0.0}  & 8.1 &  \textbf{0.0}  
& 75 &  \textbf{82}  & 74 & 73 & 75 & 62 & 76 & 81 
& 1.6 & 3.1 & 2.1 & 14 & 6.2 &  \textbf{0.085}  & 1.5 & 9.8 \\
D4 & 0.0015 &  \textbf{0.0}  & 0.02 &  \textbf{0.0}  & 3.5 & 0.037 & 2.3 &  \textbf{0.0}  
& 96 &  \textbf{98}  & 95 & 92 & 92 & 69 & 92 & 96 
& 15 & 5.7 & 20 & 12 & 13 &  \textbf{3.8}  & 11 & 25 \\
D5 & 0.013 &  \textbf{0.0}  & 0.0046 & 0.11 & 0.87 &  \textbf{0.0}  & 0.34 &  \textbf{0.0}  
& 73 &  \textbf{77}  & 62 & 48 & 76 & 73 & 76 & 74 
& 21 & 26 & 33 & 2.3 & 3.7 &  \textbf{0.0}  & 0.96 & 25 \\
D6 & 0.045 &  \textbf{0.0}  & 0.0046 & 0.02 & 0.16 & 9.8e-4 & 0.01 &  \textbf{0.0}  
& 67 & 81 & 49 & 65 & 78 & 78 & 79 &  \textbf{84}  
& 39 & 25 & 49 & 14 &  \textbf{1.1}  & 22 & 26 & 23 \\
D7 & 1.2 &  \textbf{0.0}  & 0.04 & 0.65 & 1.3 &  \textbf{0.0}  & 0.046 &  \textbf{0.0}  
& 76 & 78 & 71 & 77 & 75 & 83 &  \textbf{85}  & 80 
& 12 & 6.1 & 11 & 1.6 & 4.6 &  \textbf{0.0}  & 3.4 & 0.12 \\
D8 & 0.019 &  \textbf{0.0}  &  \textbf{0.0}  & 0.019 & 19.0 &  \textbf{0.0}  & 33.0 &  \textbf{0.0}  
&  \textbf{100}  &  \textbf{100}  &  \textbf{100}  &  \textbf{100}  &  \textbf{100}  & 50 & 75 &  \textbf{100}  
& 33 & 11 & 12 & 22 & 50 &  \textbf{0.0}  &  \textbf{0.0}  &  \textbf{0.0}  \\
\midrule
Avg. & 4.7 &  \textbf{0.0}  & 4.2 & 0.15 & 4.6 & 0.018 & 7.5 &  \textbf{0.0}  
& 81 & 85 & 76 & 77 & 82 & 72 & 81 &  \textbf{87}  
& 21 & 13 & 22 & 11 & 11 &  \textbf{3.7}  & 6.6 & 13 \\

 \bottomrule
 \end{tabular}
}
\end{table*}

\begin{table*}
 \caption{Information about the model with \emph{highest test accuracy}, among 240
 hyperparameter settings when $\threshold=0$. All numbers are percentages.}
\vspace{-10pt}
 \label{tab:max-accuracy}
 \centering 
 \resizebox{\textwidth}{!}{%
 \setlength{\tabcolsep}{.26\tabcolsep}
 \begin{tabular}{l|llllllll|llllllll|llllllll}
 \toprule
 & \vbarmc{8}{Individual discrimination} & \vbarmc{8}{Test accuracy} & \vbarmc{8}{Statistical parity difference} \\
 Id 
 & \technique{Full} & \technique{SR} & \technique{DIR} & \technique{PS} & \technique{MA} & \technique{LFR} & \technique{AD} & \technique{Our}
 & \technique{Full} & \technique{SR} & \technique{DIR} & \technique{PS} & \technique{MA} & \technique{LFR} & \technique{AD} & \technique{Our} 
 & \technique{Full} & \technique{SR} & \technique{DIR} & \technique{PS} & \technique{MA} & \technique{LFR} & \technique{AD} & \technique{Our} \\
 \midrule
 D1 & 21.0 &  \textbf{0.0}  & 21.0 & 0.73 & 6.3 & 4.5 & 17.0 & 6.6e-5 
& 82 & 82 & 82 & 82 & 82 & 83 & 85 &  \textbf{93}  
& 29 & 13 & 29 & 11 & 1.5 & 3.3 &  \textbf{0.18}  & 9.9 \\
D2 & 12.0 &  \textbf{0.0}  & 12.0 & 1.8 & 8.0 & 0.42 & 5.2 & 9.3e-5 
& 85 & 85 & 85 & 85 & 85 & 85 & 88 &  \textbf{92}  
& 22 & 13 & 21 & 11 & 5.4 &  \textbf{0.18}  & 0.37 & 13 \\
D3 & 11.0 &  \textbf{0.0}  & 8.7 & 1.1 & 1.7 & 0.56 & 11.0 & 0.007 
& 82 & 82 & 81 & 83 &  \textbf{85}  & 78 & 82 & 82 
& 11 & 3.1 & 6.1 & 3.2 &  \textbf{1.2}  & 2.6 & 5 & 7.9 \\
D4 & 0.94 &  \textbf{0.0}  & 0.44 & 0.85 & 6.2 & 1.7 & 3.3 & 0.0031 
& 98 & 98 & 98 &  \textbf{99}  & 98 & 89 & 96 & 98 
& 18 & 5.7 & 20 &  \textbf{2.4}  & 8 & 16 & 12 & 17 \\
D5 & 0.094 &  \textbf{0.0}  & 0.044 & 0.64 & 2.4 & 0.044 & 0.5 & 0.0016 
& 77 & 77 & 72 & 61 & 85 & 87 & 80 &  \textbf{100}  
& 32 & 26 & 38 & 3.9 &  \textbf{1.6}  & 21 & 22 & 26 \\
D6 & 0.05 &  \textbf{0.0}  & 0.011 & 0.028 & 0.5 & 3.0 & 0.037 & 0.0013 
& 74 & 81 & 50 & 76 & 87 & 83 & 81 &  \textbf{100}  
& 38 & 25 & 48 & 15 &  \textbf{0.078}  & 22 & 26 & 26 \\
D7 & 1.4 &  \textbf{0.0}  & 0.12 & 0.77 & 1.5 & 2.8 & 0.24 &  \textbf{0.0}  
& 78 & 78 & 75 & 78 & 78 &  \textbf{85}  &  \textbf{85}  & 80 
& 13 & 6.1 & 9 & 2.1 &  \textbf{1.2}  & 2 & 2.3 & 3.4 \\
D8 & 0.019 &  \textbf{0.0}  &  \textbf{0.0}  & 0.019 & 19.0 & 6.8 & 33.0 &  \textbf{0.0}  
&  \textbf{100}  &  \textbf{100}  &  \textbf{100}  &  \textbf{100}  &  \textbf{100}  &  \textbf{100}  & 75 &  \textbf{100}  
& 11 & 11 & 10 & 10 &  \textbf{0.0}  & 12 &  \textbf{0.0}  & 10 \\
\midrule
Avg. & 5.8 &  \textbf{0.0}  & 5.3 & 0.74 & 5.7 & 2.5 & 8.8 & 0.0016 
& 84 & 85 & 80 & 83 & 87 & 86 & 84 &  \textbf{93}  
& 22 & 13 & 23 & 7.3 &  \textbf{2.4}  & 9.9 & 8.5 & 14 \\

 \bottomrule
 \end{tabular}
 }
\end{table*}

\begin{table*}
 \caption{Information about the model with \emph{least statistical parity difference}, among 240
 hyperparameter settings when $\threshold=0$. All numbers are percentages.}
 \vspace{-10pt}
 \label{tab:min-parity}
 \centering
 \resizebox{\textwidth}{!}{%
 \setlength{\tabcolsep}{.25\tabcolsep}
 \begin{tabular}{l|llllllll|llllllll|llllllll}
 \toprule
 & \vbarmc{8}{Individual discrimination} & \vbarmc{8}{Test accuracy} & \vbarmc{8}{Statistical parity difference} \\
 Id 
 & \technique{Full} & \technique{SR} & \technique{DIR} & \technique{PS} & \technique{MA} & \technique{LFR} & \technique{AD} & \technique{Our}
 & \technique{Full} & \technique{SR} & \technique{DIR} & \technique{PS} & \technique{MA} & \technique{LFR} & \technique{AD} & \technique{Our} 
 & \technique{Full} & \technique{SR} & \technique{DIR} & \technique{PS} & \technique{MA} & \technique{LFR} & \technique{AD} & \technique{Our} \\
 \midrule
 D1 & 20.0 &  \textbf{0.0}  & 20.0 & 0.1 & 6.7 & 5.2 & 17.0 & 2.2e-5 
& 81 & 81 & 81 & 81 & 80 & 82 & 85 &  \textbf{92}  
& 28 & 12 & 27 & 8.7 &  \textbf{0.013}  &  \textbf{0.013}  & 0.18 & 7.1 \\
D2 & 13.0 &  \textbf{0.0}  & 13.0 & 1.8 & 7.5 & 0.025 & 5.3 & 4.2e-4 
& 84 & 84 & 85 & 85 & 84 & 84 & 88 &  \textbf{91}  
& 18 & 7.9 & 18 & 5 & 0.28 &  \textbf{0.0024}  & 0.37 & 7.6 \\
D3 & 7.6 &  \textbf{0.0}  & 9.1 & 2.0 & 1.2 & 0.39 & 9.1 & 0.24 
& 78 &  \textbf{82}  & 76 & 76 & 71 & 72 & 80 & 74 
& 0.32 & 0.22 & 0.026 & 0.25 &  \textbf{0.0}  &  \textbf{0.0}  & 0.19 & 0.016 \\
D4 & 3.9 &  \textbf{0.0}  & 3.6 & 1.2 & 7.1 & 5.5 & 4.6 & 0.0046 
& 93 &  \textbf{96}  & 92 & 93 & 90 & 77 & 89 & 94 
& 0.27 &  \textbf{0.075}  & 0.39 &  \textbf{0.075}  & 0.077 & 0.15 & 1.1 &  \textbf{0.075}  \\
D5 & 0.085 &  \textbf{0.0}  & 0.014 & 0.24 & 1.5 &  \textbf{0.0}  & 0.34 & 0.0059 
& 67 & 75 & 64 & 44 & 81 & 73 & 76 &  \textbf{100}  
& 20 & 17 & 26 & 0.019 & 0.011 &  \textbf{0.0}  & 0.96 & 12 \\
D6 & 0.049 &  \textbf{0.0}  & 0.0083 & 0.032 & 0.25 & 0.041 & 0.013 & 0.0021 
& 72 & 78 & 48 & 68 & 82 & 71 & 77 &  \textbf{100}  
& 33 & 19 & 42 & 9.6 &  \textbf{0.0013}  & 18 & 5.5 & 16 \\
D7 & 1.2 &  \textbf{0.0}  & 0.071 & 0.98 & 1.4 &  \textbf{0.0}  & 0.49 &  \textbf{0.0}  
& 76 & 77 & 73 & 77 & 78 & 83 &  \textbf{84}  & 77 
& 8.9 & 3 & 8.3 & 0.022 & 0.14 &  \textbf{0.0}  & 0.62 & 0.12 \\
D8 & 1.0 &  \textbf{0.0}  & 2.2 & 0.038 & 25.0 &  \textbf{0.0}  & 33.0 &  \textbf{0.0}  
&  \textbf{100}  &  \textbf{100}  &  \textbf{100}  &  \textbf{100}  &  \textbf{100}  &  \textbf{100}  & 71 &  \textbf{100}  
&  \textbf{0.0}  &  \textbf{0.0}  &  \textbf{0.0}  &  \textbf{0.0}  &  \textbf{0.0}  &  \textbf{0.0}  &  \textbf{0.0}  &  \textbf{0.0}  \\
\midrule
Avg. & 5.9 &  \textbf{0.0}  & 6.0 & 0.8 & 6.3 & 1.4 & 8.7 & 0.032 
& 81 & 84 & 77 & 78 & 83 & 80 & 81 &  \textbf{91}  
& 14 & 7.4 & 15 & 3 &  \textbf{0.065}  & 2.3 & 1.1 & 5.4 \\

 \bottomrule
 \end{tabular}
 }
\end{table*}

\subsection{Results for input space threshold $\threshold$ = 0}
\label{sec:results1}

\begin{figure*}
 \includegraphics[width=\textwidth]{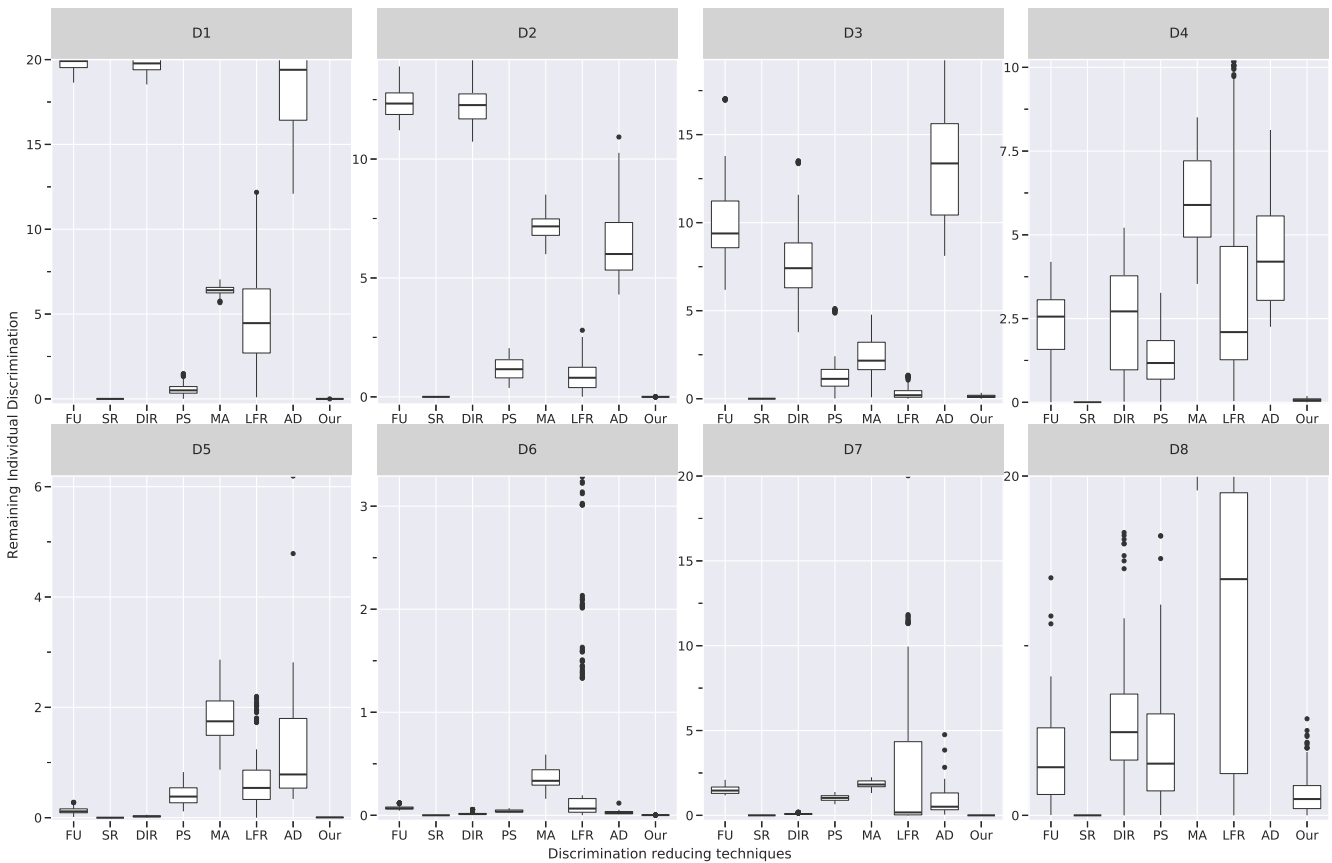}
 \vspace{-20pt}
 \caption{The individual discrimination for all 240 hyperparameter choices (lower
 is better).
 Our approach (rightmost in each boxplot) always achieves 0\% discrimination for some (often many)
 hyperparameter choices, and it has a little variance across choices.
 }
 \label{fig:facet_boxplot_discm}
\end{figure*}

The \underline{left}
portion of \cref{tab:min-discm} reports, for each technique, the \emph{best (lowest)
individual discrimination} among its 240 models.
(\Cref{fig:facet_boxplot_discm} plots, for each technique, the individual
discrimination of all 240 models.)
When multiple models have the lowest discrimination, we
choose the model with the highest accuracy. 
\textbf{Answer to \RQ{1}:}
For all datasets, our technique achieves \textbf{0\%} remaining discrimination.
\textbf{Answer to \RQ{2}:}
\technique{SR} also achieves 0\% individual discrimination. This follows from our choice of input space similarity condition. 
When the sensitive feature is removed, the remaining features are the same for all pairs of similar individuals. 
And therefore, \technique{SR} gives the same prediction for
two individuals with the same features.
\technique{LFR} and \technique{PS} leave little remaining discrimination. 
The other techniques (\technique{DIR}, \technique{MA}, and \technique{AD}) are not
effective in eliminating individual discrimination.

The \underline{center} portion of \cref{tab:min-discm} reports, for each technique, the \emph{accuracy} of the
lowest-discrimination model.
\textbf{Answer to \RQ{3}:} Our technique produces models with better
accuracy than models trained on the entire training dataset. 
This agrees with the observation made by~\citet{trade_off_2019}: ``it
requires no stretch of credulity to imagine that various personal
attributes (e.g., race, gender, religion; sometimes termed `protected
attributes') have no bearing on a person's intelligence, capability,
potential, qualifications, etc., and consequently no bearing on the ground
truth classification labels — such as job qualification status — that might
be functions of these qualities.
It then follows that enforcing fairness across these attributes should on
average increase accuracy.''
\textbf{Answer to \RQ{4}:}
\technique{DIR}, \technique{PS}, and \technique{LFR} degrade the accuracy; \technique{AD} and \technique{MA} affect it little; and \technique{SR}
improves it, though not as much as our technique does.
Our approach is always either best or within 3 percentage points of best,
and is best on average.

A user may train multiple models (e.g., using different hyperparameter
choices) and choose the best one for their application.
\Cref{tab:min-discm} assumes that the user chooses the least-discriminating model.
\Cref{tab:max-accuracy}, by contrast, assumes that the user chooses the most accurate model.
And \cref{tab:min-parity} assumes that the user chooses the model with the least
statistical disparity.
These are the
extremal points on the Pareto frontier; a user might also choose a point
between them~\cite{pareto_1}.

The left and center portions of
\cref{tab:max-accuracy} report, for each technique, the
individual discrimination and accuracy of the \emph{most accurate model} among the 240 models. 
We can answer \RQ{1}--\RQ{4} about these models.
\textbf{Answer to \RQ{1} and \RQ{2}:}
Our approach achieves, on average, nearly 0\% (0.0016\%) remaining
discrimination; under the definition of our input space similarity condition,
\technique{SR} achieves 0\% discrimination; other techniques are
much higher.
\textbf{Answer to \RQ{3} and \RQ{4}:}
Our approach achieves by far the highest average accuracy: 93\% compared
to the \technique{Full} model's 84\% accuracy and \technique{MA}'s 87\% accuracy.

\begin{table*}
 \caption{Information about the model with the \emph{least remaining discrimination}, 
 among 240 hyperparameter settings when $\threshold>0$. All numbers are percentages.}
 \label{tab:min-discm2}
 \vspace{-10pt}
 \centering 
 \resizebox{\textwidth}{!}{%
 \setlength{\tabcolsep}{.24\tabcolsep}
 \begin{tabular}{l|llllllll|llllllll|llllllll}
 \toprule
 & \vbarmc{8}{Individual discrimination} & \vbarmc{8}{Test accuracy} & \vbarmc{8}{Statistical parity difference} \\
 Id 
 & \technique{Full} & \technique{SR} & \technique{DIR} & \technique{PS} & \technique{MA} & \technique{LFR} & \technique{AD} & \technique{Our}
 & \technique{Full} & \technique{SR} & \technique{DIR} & \technique{PS} & \technique{MA} & \technique{LFR} & \technique{AD} & \technique{Our} 
 & \technique{Full} & \technique{SR} & \technique{DIR} & \technique{PS} & \technique{MA} & \technique{LFR} & \technique{AD} & \technique{Our} \\
 \midrule
 D1 & 19.0 &  \textbf{0.0026}  & 19.0 & 0.042 & 5.6 & 0.41 & 12.0 & 0.62 
& 81 & 83 & 81 & 85 & 81 & 84 & 85 &  \textbf{86}  
& 29 & 15 & 28 & 8.9 & 3.3 & 3.6 &  \textbf{2.3}  & 9.8 \\
D2 & 11.0 &  \textbf{0.00044}  & 11.0 & 0.45 & 6.0 & 0.37 & 4.3 & 0.67 
& 79 & 80 & 80 & 80 & 79 & 80 &  \textbf{85}  & 81 
& 21 & 8.1 & 20 & 12 & 7 &  \textbf{0.59}  & 7.8 & 12 \\
D3 & 6.4 & 1.9 & 4.2 & 2.2 & 2.0 &  \textbf{0.066}  & 8.7 & 1.6 
& 75 & 81 & 74 & 73 & 76 & 62 & 76 &  \textbf{82}  
& 1.6 &  \textbf{1.1}  & 2.1 & 2.1 & 3.3 & 2.7 & 1.5 & 7.6 \\
D4 & 4.7 & 4.6 & 4.8 & 4.7 & 4.2 &  \textbf{0.87}  & 4.5 & 2.3 
& 99 & 98 &  \textbf{100}  & 98 & 96 & 90 & 98 & 97 
& 13 & 12 & 6.6 & 13 & 12 & 5.4 & 11 &  \textbf{4.4}  \\
D5 & 0.19 & 0.2 & 0.0052 & 0.14 & 0.97 &  \textbf{0.0}  & 0.36 & 8.1e$-$5 
& 63 & 62 & 60 & 44 & 63 & 70 & 64 &  \textbf{100}  
& 26 & 24 & 33 & 2.4 & 3.7 &  \textbf{0.0}  & 8.4 & 0.72 \\
D6 & 0.048 & 0.046 & 0.0033 & 0.033 & 0.21 & 0.18 & 0.0088 &  \textbf{0.0002}  
& 60 & 67 & 49 & 59 & 72 & 72 & 67 &  \textbf{99}  
& 38 & 23 & 47 & 14 &  \textbf{1.1}  & 21 & 26 & 26 \\
D7 & 1.8 & 1.6 & 0.19 & 1.5 & 1.9 &  \textbf{0.0}  & 0.079 &  \textbf{0.0}  
& 76 & 76 & 71 & 76 & 75 &  \textbf{86}  & 85 & 83 
& 12 & 4.7 & 11 & 1.4 & 4.6 &  \textbf{0.0}  & 3.4 & 6.5 \\
D8 & 1.7 & 1.3 & 0.033 & 1.5 & 19.0 &  \textbf{0.0}  & 33.0 &  \textbf{0.0}  
&  \textbf{100}  &  \textbf{100}  &  \textbf{100}  &  \textbf{100}  &  \textbf{100}  & 57 & 75 &  \textbf{100}  
&  \textbf{0.0}  &  \textbf{0.0}  & 30 &  \textbf{0.0}  & 50 &  \textbf{0.0}  &  \textbf{0.0}  & 30 \\
\midrule
Avg. & 5.6 & 1.2 & 4.9 & 1.3 & 5.0 &  \textbf{0.24}  & 7.9 & 0.65 
& 79 & 80 & 76 & 76 & 80 & 75 & 79 &  \textbf{91}  
& 18 & 11 & 22 & 6.7 & 11 &  \textbf{4.2}  & 7.5 & 12 \\

 \bottomrule
 \end{tabular}
}
\end{table*}

\begin{table*}
 \caption{Information about the model with \emph{highest test accuracy}, among 240
 hyperparameter settings when $\threshold>0$. All numbers are percentages.}
 \label{tab:max-accuracy2}
 \vspace{-10pt}
 \centering 
 \resizebox{\textwidth}{!}{%
 \setlength{\tabcolsep}{.26\tabcolsep}
 \begin{tabular}{l|llllllll|llllllll|llllllll}
 \toprule
 & \vbarmc{8}{Individual discrimination} & \vbarmc{8}{Test accuracy} & \vbarmc{8}{Statistical parity difference} \\
 Id 
 & \technique{Full} & \technique{SR} & \technique{DIR} & \technique{PS} & \technique{MA} & \technique{LFR} & \technique{AD} & \technique{Our}
 & \technique{Full} & \technique{SR} & \technique{DIR} & \technique{PS} & \technique{MA} & \technique{LFR} & \technique{AD} & \technique{Our} 
 & \technique{Full} & \technique{SR} & \technique{DIR} & \technique{PS} & \technique{MA} & \technique{LFR} & \technique{AD} & \technique{Our} \\
 \midrule
 D1 & 20.0 &  \textbf{0.11}  & 21.0 & 0.5 & 6.2 & 4.5 & 17.0 & 0.62 
& 82 & 84 & 82 & 85 & 82 & 85 & 85 &  \textbf{86}  
& 29 & 13 & 29 & 11 & 1.5 & 3.3 &  \textbf{0.18}  & 12 \\
D2 & 12.0 &  \textbf{0.22}  & 12.0 & 1.8 & 7.7 & 0.81 & 5.3 & 0.89 
& 80 & 81 & 81 & 81 & 81 & 80 &  \textbf{87}  & 82 
& 22 & 13 & 21 & 11 & 5.4 &  \textbf{0.18}  & 0.37 & 14 \\
D3 & 11.0 & 2.1 & 8.7 & 2.6 & 2.7 &  \textbf{1.1}  & 11.0 & 1.6 
& 82 & 82 & 81 & 83 &  \textbf{85}  & 78 & 83 & 82 
& 11 & 3.1 & 6.1 & 3.2 &  \textbf{1.2}  & 2.6 & 5 & 5.8 \\
D4 & 5.1 & 4.8 & 4.8 & 4.9 & 5.0 & 2.8 & 5.1 &  \textbf{2.7}  
&  \textbf{100}  &  \textbf{100}  &  \textbf{100}  &  \textbf{100}  &  \textbf{100}  & 96 &  \textbf{100}  &  \textbf{100}  
& 18 & 5.7 & 20 &  \textbf{2.4}  & 8 & 16 & 12 & 16 \\
D5 & 0.58 & 0.57 & 0.065 & 0.71 & 2.6 & 4.8 & 0.61 &  \textbf{8.1e$-$5}  
& 75 & 74 & 74 & 55 & 75 & 87 & 67 &  \textbf{100}  
& 32 & 26 & 38 & 3.9 &  \textbf{1.6}  & 21 & 22 & 28 \\
D6 & 0.054 & 0.06 & 0.01 & 0.11 & 0.59 & 5.4 & 0.034 &  \textbf{0.00021}  
& 66 & 74 & 50 & 70 & 82 & 81 & 75 &  \textbf{100}  
& 38 & 25 & 48 & 15 &  \textbf{0.078}  & 22 & 26 & 19 \\
D7 & 2.0 & 1.7 & 0.62 & 1.7 & 2.1 & 3.7 & 0.26 &  \textbf{0.0}  
& 78 & 78 & 75 & 79 & 78 &  \textbf{87}  & 85 & 83 
& 13 & 6.1 & 9 & 2.1 &  \textbf{1.2}  & 2 & 2.3 & 12 \\
D8 & 1.7 & 1.3 & 0.033 & 1.5 & 19.0 & 7.3 & 33.0 &  \textbf{0.0}  
&  \textbf{100}  &  \textbf{100}  &  \textbf{100}  &  \textbf{100}  &  \textbf{100}  &  \textbf{100}  & 75 &  \textbf{100}  
& 11 & 11 & 10 & 10 &  \textbf{0.0}  & 12 &  \textbf{0.0}  & 10 \\
\midrule
Avg. & 6.6 & 1.4 & 5.9 & 1.7 & 5.7 & 3.8 & 9.0 &  \textbf{0.73}  
& 82 & 84 & 80 & 81 & 85 & 86 & 82 &  \textbf{91}  
& 22 & 13 & 23 & 7.3 &  \textbf{2.4}  & 9.9 & 8.5 & 15 \\

 \bottomrule
 \end{tabular}
 }
\end{table*}

\begin{table*}
 \caption{Information about the model with \emph{least statistical parity difference}, among 240
 hyperparameter settings when $\threshold>0$. All numbers are percentages.}
 \label{tab:min-parity2}
 \vspace{-10pt}
 \centering
 \resizebox{\textwidth}{!}{%
 \setlength{\tabcolsep}{.25\tabcolsep}
 \begin{tabular}{l|llllllll|llllllll|llllllll}
 \toprule
 & \vbarmc{8}{Individual discrimination} & \vbarmc{8}{Test accuracy} & \vbarmc{8}{Statistical parity difference} \\
 Id 
 & \technique{Full} & \technique{SR} & \technique{DIR} & \technique{PS} & \technique{MA} & \technique{LFR} & \technique{AD} & \technique{Our}
 & \technique{Full} & \technique{SR} & \technique{DIR} & \technique{PS} & \technique{MA} & \technique{LFR} & \technique{AD} & \technique{Our} 
 & \technique{Full} & \technique{SR} & \technique{DIR} & \technique{PS} & \technique{MA} & \technique{LFR} & \technique{AD} & \technique{Our} \\
 \midrule
 D1 & 20.0 & 0.11 & 20.0 &  \textbf{0.1}  & 6.7 & 5.2 & 17.0 & 0.75 
& 81 & 84 & 82 &  \textbf{85}  & 81 & 82 &  \textbf{85}  &  \textbf{85}  
& 28 & 12 & 27 & 8.7 &  \textbf{0.013}  &  \textbf{0.013}  & 0.18 & 8.8 \\
D2 & 13.0 &  \textbf{0.22}  & 13.0 & 1.8 & 7.5 & 1.1 & 5.3 & 1.2 
& 79 & 79 & 80 & 80 & 79 & 79 &  \textbf{87}  & 81 
& 18 & 7.9 & 18 & 5 & 0.28 &  \textbf{0.0024}  & 0.37 & 9 \\
D3 & 7.8 & 2.6 & 9.1 & 3.0 & 3.3 &  \textbf{1.5}  & 9.5 & 2.8 
& 79 &  \textbf{82}  & 75 & 77 & 71 & 73 & 80 & 78 
& 0.32 & 0.22 & 0.026 & 0.25 &  \textbf{0.0}  &  \textbf{0.0}  & 0.19 & 0.095 \\
D4 & 6.0 & 5.3 & 5.9 & 5.4 & 7.2 & 5.9 & 6.2 &  \textbf{4.2}  
&  \textbf{99}  &  \textbf{99}  &  \textbf{99}  &  \textbf{99}  & 97 & 79 & 98 &  \textbf{99}  
& 0.27 &  \textbf{0.075}  & 0.39 &  \textbf{0.075}  & 0.077 & 0.15 & 1.1 &  \textbf{0.075}  \\
D5 & 0.32 & 0.41 & 0.021 & 0.28 & 1.7 &  \textbf{0.0}  & 0.38 & 9.8e$-$5 
& 63 & 64 & 62 & 44 & 72 & 70 & 65 &  \textbf{100}  
& 20 & 17 & 26 & 0.019 & 0.011 &  \textbf{0.0}  & 0.96 & 0.044 \\
D6 & 0.052 & 0.061 & 0.0088 & 0.058 & 0.31 & 0.46 & 0.014 &  \textbf{0.00023}  
& 66 & 70 & 48 & 64 & 78 & 70 & 74 &  \textbf{99}  
& 33 & 19 & 42 & 9.6 &  \textbf{0.0013}  & 18 & 5.5 & 17 \\
D7 & 2.1 & 1.9 & 0.46 & 1.9 & 2.2 &  \textbf{0.0}  & 0.65 &  \textbf{0.0}  
& 76 & 75 & 73 & 78 & 78 &  \textbf{86}  & 84 & 81 
& 8.9 & 3 & 8.3 & 0.022 & 0.14 &  \textbf{0.0}  & 0.62 & 6.5 \\
D8 & 1.7 & 1.3 & 2.2 & 1.5 & 25.0 &  \textbf{0.0}  & 33.0 & 0.86 
&  \textbf{100}  &  \textbf{100}  &  \textbf{100}  &  \textbf{100}  &  \textbf{100}  &  \textbf{100}  & 75 &  \textbf{100}  
&  \textbf{0.0}  &  \textbf{0.0}  &  \textbf{0.0}  &  \textbf{0.0}  &  \textbf{0.0}  &  \textbf{0.0}  &  \textbf{0.0}  &  \textbf{0.0}  \\
\midrule
Avg. & 6.4 & 1.5 & 6.3 & 1.8 & 6.7 & 1.8 & 9.0 &  \textbf{1.2}  
& 80 & 81 & 77 & 78 & 82 & 79 & 81 &  \textbf{90}  
& 14 & 7.4 & 15 & 3 &  \textbf{0.065}  & 2.3 & 1.1 & 5.2 \\

 \bottomrule
 \end{tabular}
 }
\end{table*}

\begin{figure*}[h!]
 \includegraphics[width=\textwidth]{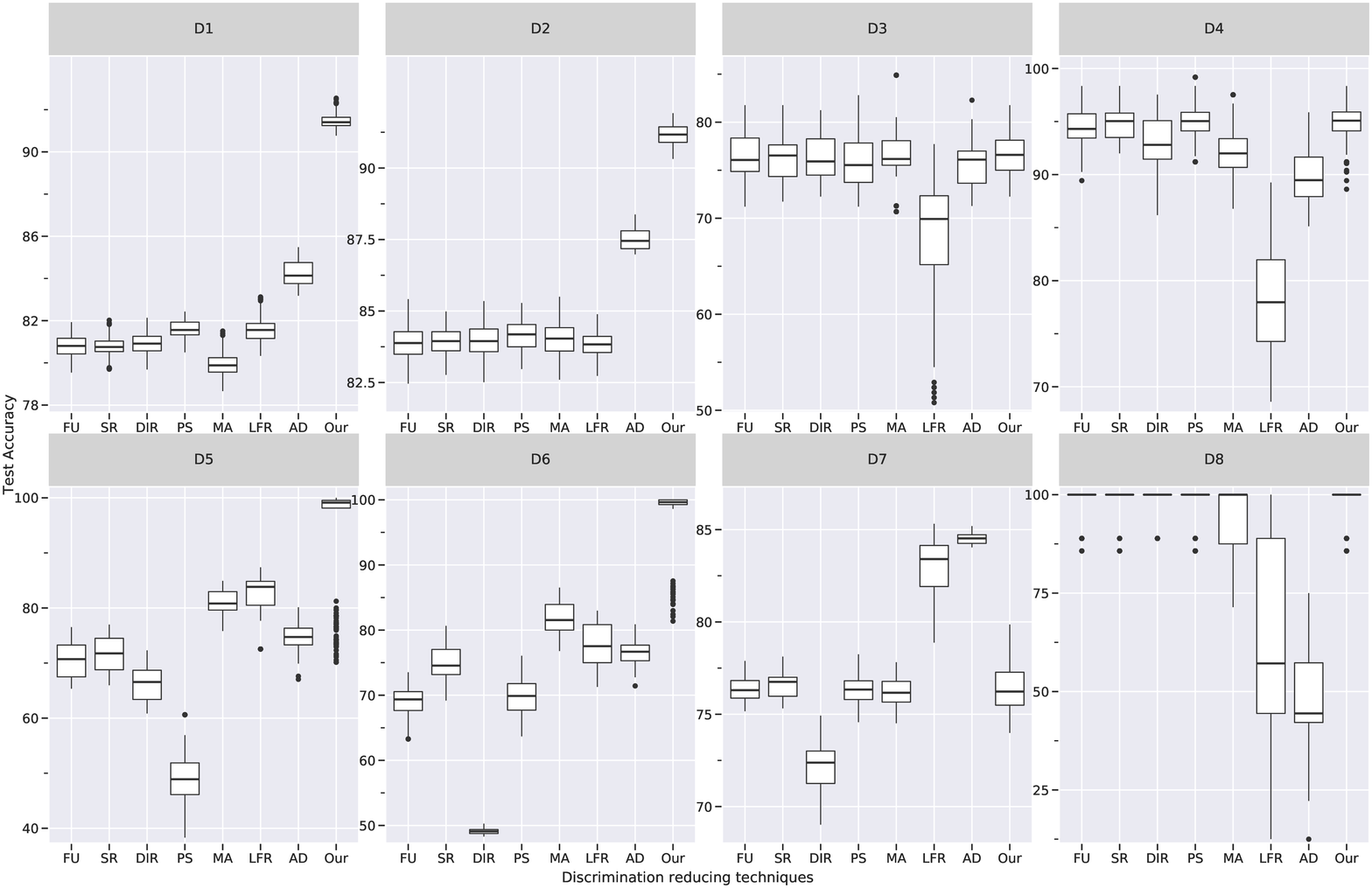}
 \vspace{-20pt}
 \caption{The test accuracy for all 240 hyperparameter choices (higher is
 better).
 Our approach is best or comparable to the best in terms of both accuracy
 and its variance, for all experiments except D7.\looseness=-1}
\label{fig:facet_boxplot_accuracy}
\end{figure*}

We also measured the statistical disparity: the absolute
difference between the success rate for individuals with one value for the
sensitive attribute and the success rate for individuals with the other
value for the sensitive attribute. 
The \underline{right} portions of \cref{tab:min-discm,tab:max-accuracy,tab:min-parity}
report the results.
\textbf{Answer to \RQ{5}:} 
Our technique achieves a lower statistical disparity than the
baseline model trained on the full dataset. Compared to the other
techniques (which are designed to optimize for statistical parity
difference), our technique is in the middle of the pack. For each technique
that has a lower statistical disparity than ours, our technique
achieves considerably lower individual discrimination \emph{and} higher accuracy.


If a bias mitigation technique is highly sensitive to hyperparameter
choices, then users might need to run it many times to achieve desirable performance,
and they may have less confidence in its generalizability. 

\Cref{fig:facet_boxplot_discm} shows the remaining individual
discrimination for all 240 models we trained for each experiment and each technique, and 
\cref{fig:facet_boxplot_accuracy} shows the test accuracy for all the 240 models for each experiment and each technique. 

\textbf{Answer to \RQ{6}:}
Our technique not only usually results in lower remaining
discrimination and higher accuracy than previous techniques, it is also much
less sensitive to hyperparameter choices (narrower range). 

\subsection{Results for input space threshold $\threshold$ $>$ 0}
\label{sec:results2}
\Cref{tab:min-discm2,tab:max-accuracy2,tab:min-parity2} report the same
statistics for the experiment with $\threshold > 0$.
We answer \RQ{1}-\RQ{4} using \Cref{tab:min-discm2} and \Cref{tab:max-accuracy2}. 

\textbf{Answer to \RQ{1}:} For both the models with the least discrimination and highest accuracy, our approach achieves very low individual discrimination (0.65\% and 0.73\% respectively). 

\textbf{Answer to \RQ{2}:} For the most accurate model, our approach gets the lowest discrimination among all approaches (0.73\%). 
For the models with the least discrimination, our approach is second and very close to the best performing approach (\technique{LFR}). 
Notably, \technique{SR} has much higher discrimination for this choice of similarity metric. 
In the previous set of experiments, \technique{SR} was able to get 0\% individual discrimination just by choice of input space similarity. 

\textbf{Answer to \RQ{3}:} Our approach gets the best accuracy by far compared to all the baselines for both the models with the least discrimination and highest accuracy. 

\textbf{Answer to \RQ{4}:} Similar to the conclusions from the experiments
with $\threshold=0$, for the least discriminative model: \technique{DIR}, \technique{PS}, and \technique{LFR} degrade the accuracy; \technique{AD}, \technique{MA}, and \technique{SR} affect it little. 
For the most accurate model: \technique{SR}, \technique{MA}, and \technique{LFR} improve the accuracy but not as much as our approach. 

\textbf{Answer to \RQ{5}:} For statistical disparity (shown in the
right portions of \cref{tab:min-discm2,tab:max-accuracy2,tab:min-parity2})
the conclusions are the same as when $\threshold=0$.

\textbf{Answer to \RQ{6}:} Due to lack of space, we have added the plots
showing the remaining discrimination and test accuracy for the experiments
with $\threshold>0$ in the appendix (\Cref{sec:extraplots}). 
The conclusions are the same as when $\threshold=0$.


\section{Related Work}
\label{sec:related}

\subsection{Fairness Metrics}

More than 20 metrics for fairness have been proposed~\cite{verma_fairness_2018}. 
They can be broadly divided into three categories: group fairness, causal fairness, and individual fairness. 

\paragraph{Group fairness}

Most group fairness metrics declare a model to be fair if it is satisfies a
specific constraint from the confusion
matrix~\cite{Confusion-Matrix} e.g.,
predictive parity~\cite{Simoiu2016,Chouldechova2016FairPW} (equal probability of being correctly classified into the favorable class for all demographic groups), 
predictive equality~\cite{CorbettDavies2017AlgorithmicDM,Chouldechova2016FairPW} (equal true negative rate for all groups),
equality of opportunity~\cite{Chouldechova2016FairPW,Hardt2016EqualityOO,kusner_counterfactual} (equal true positive rate for all groups),
equalized odds~\cite{Hardt2016EqualityOO,Zafar2017FairnessBD} (equal true positive rate and equal true negative rate for all groups), 
accuracy equality~\cite{Simoiu2016} (equal predictive accuracy for all groups), 
treatment equality~\cite{Simoiu2016} (equal ratio of false negatives and false positives for all groups),
calibration~\cite{Chouldechova2016FairPW,Hardt2016EqualityOO},
well-calibration~\cite{Kleinberg2016InherentTI}, 
and balance for positive and negative classes~\cite{Kleinberg2016InherentTI}. 
Most definitions in the group fairness category require the ground truth,
which is unavailable
 before deploying a model~\cite{group_fairness:2020}. 

The most popular definition from the group fairness category is statistical
parity, which states that a decision-producing system is fair if the
probability of getting a favorable outcome is equal for members of all
demographic groups formed by sensitive features. 


Most discrimination-reducing approaches evaluate their methodology on
statistical parity or measures close to it~\cite{feldman_certifying_2015,
 kamiran_data_2012, calmon_optimized, zemel_learning,
 Calders-Naive-2010,Kamishima:2012, friedler_comparative_2018,
 wadsworth_achieving_2018}.
 Dwork et al.~\cite{dwork_fairness_2011} criticize statistical parity by
 showing how three evils (reduced utility, self-fulfilling prophecy, and
 subset targeting) can occur even while statistical parity is maintained.
Statistical parity is also not applicable in scenarios where the base rates of ground-truth occurrence are different, 
e.g., criminal justice~\cite{Chouldechova2016FairPW,dwork_fairness_2011,Hardt2016EqualityOO,Kleinberg2016InherentTI}.

\paragraph{Causal fairness}

Causal fairness metrics require a causal model that is used to reason about the effects of certain features on other features and the outcome~\cite{causal_1:2019,causal_2:2017}. 
Causal models are represented by a graph with features as nodes and directed edges showing the effects of one feature on another. 
Learning causal models from the data is not always possible and therefore requires domain knowledge~\citet{pearl2009causality}. 
A causal model is consequently an untestable assumption about the data generation process. 
Papers proposing causal fairness definitions assume a given causal model and evaluate their methodology based on this assumption~\cite{kusner_counterfactual,kilbertus_avoiding,Nabi2017FairIO}.

\paragraph{Individual fairness}

Individual fairness~\cite{dwork_fairness_2011,avg_IF,Jung2019ElicitingAE,Lahoti2019_ops_IF}
(defined in \cref{sec:intro}) states that similar individuals should
be treated similarly: they should be given similar predictions.
The similarity metric for individuals should only
consider the non-sensitive features as that ensures adherence to
anti-discrimination laws.
This matches a common intuition, does not
make assumptions about data generation~\cite{sexgottodo:2020} or base
rates, and does not require the presence of ground truth labels.
\paragraph{Trading off group and individual fairness}

Both group fairness and individual fairness are desirable. It is a policy
and political decision which one to prioritize. (A related
policy/political question is what forms of affirmative action, if any, are
just.)
Previous work has largely ignored individual fairness, which we argue is an
oversight.

Our experiments show that group and individual fairness must be traded off
in a relative sense: maximizing one leads to the other taking on a
non-maximal value. However, they do not need to be traded off in an
absolute sense: while maximizing one, it is still possible to improve the
other. Our technique is a bright spot: in
\cref{tab:min-discm,tab:max-accuracy,tab:min-parity}, ours is the only
technique (out of \baselinesPlusOurs) that always improves test accuracy,
individual discrimination, \emph{and} statistical parity. Its
interventions may be acceptable across the political spectrum. This is an
exciting new direction for research in fairness in machine learning.

\subsection{Fairness Literature}
Most previous work in the fairness literature~\cite{dunkelau_fairness-aware,mehrabi_survey_2019,friedler_comparative_2018,survey_accountability:2020} can be categorized into
discrimination detection and interventional approaches.

Discrimination detection~\cite{galhotra_fairness_2017,Udeshi-testing:2018,Aggarwal-testing:2019,fliptest:2020}
measures whether a learned model is biased.
Our experiments take inspiration from Themis~\cite{galhotra_fairness_2017}, which measures 
individual discrimination by generating similar individuals, who only differ in 
a sensitive attribute, sampling individuals uniformly at random from the feature space, which is captured from training data. 

Interventional approaches aim to improve the fairness of a learned model. They
can be further categorized into three groups based on the stage of machine
learning pipeline they intervene in:

\begin{enumerate}[nosep,leftmargin=0pt]
\item
\emph{Pre-processing}: Intervention at the stage of training data. Previous work modifies the training data to reduce discrimination and measures it using their preferred fairness metric~\cite{Edwards2015CensoringRW,Calders-building-2009,Kamiran2009ClassifyingWD,Kamiran-no-discm-2010,kamiran_data_2012,calmon_optimized,feldman_certifying_2015,Li2014LearningUF,Louizos2015TheVF,Hacker-continuous-2017,Hajian-method-2013,Hajian-rule-2011,Johndrow2017AnAF,kilbertus-blind-18,Lum2016ASF,Luong-knn-2011,McNamara2017ProvablyFR,Thomas2019,Metevier2019,salimi_capuchin:_2019}. 
 Merely removing the sensitive feature (the technique called SR in this paper)
 does not necessarily yield a fair model~\cite{sensitive_removal1,sensitive_removal2,sensitive_removal3,sensitive_removal4,sensitive_removal5,dwork_fairness_2011} and is important to assess disparities~\cite{Bogen:2020,Kallus2020AssessingAF,Lipton2018DoesMM}. 
 A model can learn to make decisions based on proxy features that are
 correlated with sensitive attributes, e.g., zip code can encode racial groups. 
 Removing all the proxies would cause a large dip in accuracy~\cite{sensitive_removal3}.

\item
\emph{In-processing}: Intervention at the stage of training. Previous work
 modifies the learning
 algorithm~\cite{Calders-Naive-2010,Kamiran-tree-2010,Russell-WWC-2017,Dwork-DC:2018},
 modifies the loss
 function~\cite{Kamishima:2012,Kamishima-RA:2011,zemel_learning,Bechavod2017LearningFC},
 uses an adversarial
 approach~\cite{Beutel2017DataDA,Zhang2018MitigatingUB,wadsworth_achieving_2018},
 or adds fairness
 constraints~\cite{Zafar2017FairnessBD,Zafar2017FairnessCM,Agarwal-RA:2018,Fish2016ACA,Donini-RM:2018,Celis2018ClassificationWF,Ruoss2020LearningCI}.
 
\item
 \emph{Post-processing}: Intervention at the stage of deployment of a
 trained model. Previous work modifies the input before passing it to the
 model~\cite{Adler-AI:2018,Hardt2016EqualityOO,Pleiss-FC:2017,kusner_counterfactual,Card2018DeepWA,Woodworth2017LearningNP},
 or modifies the model's prediction
 depending on the input's sensitive
 attribute~\cite{Kamiran-tree-2010,Kamiran-DT-2012,Pedreshi-DA-2008,Pedreschi-SS:2009,Menon-CF:2018}.
 Since
 the sensitive attribute is used to affect model predictions directly,
 this class of approach might be illegal due to disparate treatment laws~\cite{jolly-ryan-have,Winrow2010TheDB}.

\end{enumerate}



\section{Conclusion}
\label{sec:conclusion}

Building fair machine learning models is required for adherence to anti-discrimination laws; it
leads to more desirable outcomes (e.g., higher profits); and it is the morally right thing to do. Training
a machine learning model on biased historical decisions
would perpetuate injustice.

We have proposed a novel approach to improve fairness. Our approach
heuristically identifies unfair decisions made by a model, uses influence
functions to identify the training data (e.g., biased historical decisions)
that are most responsible for the unfair decisions, and then removes the
biased training points.

Compared to a baseline model that is trained on historical data without
removing any datapoints, our technique improves test accuracy, individual
discrimination, \emph{and} statistical disparity. Ours is the only
technique (out of \baselinesPlusOurs tested) that improves all three
measures, no matter which is chosen as the optimization goal. By
contrast, much previous work increases fairness only at the expense of
accuracy.



\bibliographystyle{ACM-Reference-Format}
\bibliography{refs.bib}


\begin{thebibliography}{124}


\ifx \showCODEN    \undefined \def \showCODEN     #1{\unskip}     \fi
\ifx \showDOI      \undefined \def \showDOI       #1{#1}\fi
\ifx \showISBNx    \undefined \def \showISBNx     #1{\unskip}     \fi
\ifx \showISBNxiii \undefined \def \showISBNxiii  #1{\unskip}     \fi
\ifx \showISSN     \undefined \def \showISSN      #1{\unskip}     \fi
\ifx \showLCCN     \undefined \def \showLCCN      #1{\unskip}     \fi
\ifx \shownote     \undefined \def \shownote      #1{#1}          \fi
\ifx \showarticletitle \undefined \def \showarticletitle #1{#1}   \fi
\ifx \showURL      \undefined \def \showURL       {\relax}        \fi
\providecommand\bibfield[2]{#2}
\providecommand\bibinfo[2]{#2}
\providecommand\natexlab[1]{#1}
\providecommand\showeprint[2][]{arXiv:#2}

\bibitem[\protect\citeauthoryear{Adler, Falk, Friedler, Nix, Rybeck,
  Scheidegger, Smith, and Venkatasubramanian}{Adler et~al\mbox{.}}{2018}]%
        {Adler-AI:2018}
\bibfield{author}{\bibinfo{person}{Philip Adler}, \bibinfo{person}{Casey Falk},
  \bibinfo{person}{Sorelle~A. Friedler}, \bibinfo{person}{Tionney Nix},
  \bibinfo{person}{Gabriel Rybeck}, \bibinfo{person}{Carlos Scheidegger},
  \bibinfo{person}{Brandon Smith}, {and} \bibinfo{person}{Suresh
  Venkatasubramanian}.} \bibinfo{year}{2018}\natexlab{}.
\newblock \showarticletitle{Auditing Black-Box Models for Indirect Influence}.
\newblock \bibinfo{journal}{\emph{Knowl. Inf. Syst.}} \bibinfo{volume}{54},
  \bibinfo{number}{1} (\bibinfo{date}{Jan.} \bibinfo{year}{2018}),
  \bibinfo{pages}{95--122}.
\newblock
\showISSN{0219-1377}
\urldef\tempurl%
\url{https://doi.org/10.1007/s10115-017-1116-3}
\showDOI{\tempurl}


\bibitem[\protect\citeauthoryear{Agarwal, Beygelzimer, Dudik, Langford, and
  Wallach}{Agarwal et~al\mbox{.}}{2018}]%
        {Agarwal-RA:2018}
\bibfield{author}{\bibinfo{person}{Alekh Agarwal}, \bibinfo{person}{Alina
  Beygelzimer}, \bibinfo{person}{Miroslav Dudik}, \bibinfo{person}{John
  Langford}, {and} \bibinfo{person}{Hanna Wallach}.}
  \bibinfo{year}{2018}\natexlab{}.
\newblock \showarticletitle{A Reductions Approach to Fair Classification}. In
  \bibinfo{booktitle}{\emph{Proceedings of Machine Learning Research}},
  \bibfield{editor}{\bibinfo{person}{Jennifer Dy} {and}
  \bibinfo{person}{Andreas Krause}} (Eds.), Vol.~\bibinfo{volume}{80}.
  \bibinfo{publisher}{PMLR}, \bibinfo{address}{Stockholmsmässan, Stockholm
  Sweden}, \bibinfo{pages}{60--69}.
\newblock
\urldef\tempurl%
\url{http://proceedings.mlr.press/v80/agarwal18a.html}
\showURL{%
\tempurl}


\bibitem[\protect\citeauthoryear{Aggarwal, Lohia, Nagar, Dey, and
  Saha}{Aggarwal et~al\mbox{.}}{2019}]%
        {Aggarwal-testing:2019}
\bibfield{author}{\bibinfo{person}{Aniya Aggarwal}, \bibinfo{person}{Pranay
  Lohia}, \bibinfo{person}{Seema Nagar}, \bibinfo{person}{Kuntal Dey}, {and}
  \bibinfo{person}{Diptikalyan Saha}.} \bibinfo{year}{2019}\natexlab{}.
\newblock \showarticletitle{Black Box Fairness Testing of Machine Learning
  Models}. In \bibinfo{booktitle}{\emph{Proceedings of the 2019 27th ACM Joint
  Meeting on European Software Engineering Conference and Symposium on the
  Foundations of Software Engineering}} (Tallinn, Estonia)
  \emph{(\bibinfo{series}{ESEC/FSE 2019})}. \bibinfo{publisher}{Association for
  Computing Machinery}, \bibinfo{address}{New York, NY, USA},
  \bibinfo{pages}{625--635}.
\newblock
\showISBNx{9781450355728}
\urldef\tempurl%
\url{https://doi.org/10.1145/3338906.3338937}
\showDOI{\tempurl}


\bibitem[\protect\citeauthoryear{Alexandra}{Alexandra}{2018}]%
        {less_effective2}
\bibfield{author}{\bibinfo{person}{Alexandra}.}
  \bibinfo{year}{2018}\natexlab{}.
\newblock \bibinfo{title}{11 Ways to Reduce Hiring Bias}.
\newblock
  \bibinfo{howpublished}{\url{https://harver.com/blog/reduce-hiring-bias/}}.
\newblock
\newblock
\shownote{[Online; accessed 13-Septmeber-2020].}


\bibitem[\protect\citeauthoryear{{Australia}}{{Australia}}{2014}]%
        {australia_act}
\bibfield{author}{\bibinfo{person}{{Australia}}.}
  \bibinfo{year}{2014}\natexlab{}.
\newblock \bibinfo{title}{Sex Discrimination Act 1984}.
\newblock
  \bibinfo{howpublished}{\url{https://www.legislation.gov.au/Details/C2014C00002}}.
\newblock
\newblock
\shownote{[Online; accessed 10-September-2020].}


\bibitem[\protect\citeauthoryear{Barocas and Selbst}{Barocas and
  Selbst}{2016}]%
        {Barocas2016BigDD}
\bibfield{author}{\bibinfo{person}{Solon Barocas} {and}
  \bibinfo{person}{Andrew~D. Selbst}.} \bibinfo{year}{2016}\natexlab{}.
\newblock \showarticletitle{Big Data's Disparate Impact}.
\newblock \bibinfo{journal}{\emph{California Law Review}}
  \bibinfo{volume}{104} (\bibinfo{year}{2016}), \bibinfo{pages}{671}.
\newblock


\bibitem[\protect\citeauthoryear{Bechavod and Ligett}{Bechavod and
  Ligett}{2017}]%
        {Bechavod2017LearningFC}
\bibfield{author}{\bibinfo{person}{Yahav Bechavod} {and}
  \bibinfo{person}{Katrina Ligett}.} \bibinfo{year}{2017}\natexlab{}.
\newblock \bibinfo{title}{Learning Fair Classifiers: A Regularization-Inspired
  Approach}.
\newblock \bibinfo{howpublished}{ArXiv}.
\newblock
\newblock
\shownote{abs/1707.00044.}


\bibitem[\protect\citeauthoryear{Bellamy, Dey, Hind, Hoffman, Houde, Kannan,
  Lohia, Martino, Mehta, Mojsilovic, Nagar, Ramamurthy, Richards, Saha,
  Sattigeri, Singh, Varshney, and Zhang}{Bellamy et~al\mbox{.}}{2018}]%
        {Bellamy2018AIF3}
\bibfield{author}{\bibinfo{person}{Rachel K.~E. Bellamy},
  \bibinfo{person}{Kuntal Dey}, \bibinfo{person}{Michael Hind},
  \bibinfo{person}{Samuel~C. Hoffman}, \bibinfo{person}{Stephanie Houde},
  \bibinfo{person}{Kalapriya Kannan}, \bibinfo{person}{Pranay Lohia},
  \bibinfo{person}{Jacquelyn Martino}, \bibinfo{person}{Sameep Mehta},
  \bibinfo{person}{Aleksandra Mojsilovic}, \bibinfo{person}{Seema Nagar},
  \bibinfo{person}{Karthikeyan~Natesan Ramamurthy}, \bibinfo{person}{John~T.
  Richards}, \bibinfo{person}{Diptikalyan Saha}, \bibinfo{person}{Prasanna
  Sattigeri}, \bibinfo{person}{Moninder Singh}, \bibinfo{person}{Kush~R.
  Varshney}, {and} \bibinfo{person}{Yunfeng Zhang}.}
  \bibinfo{year}{2018}\natexlab{}.
\newblock \bibinfo{title}{AI Fairness 360: An Extensible Toolkit for Detecting,
  Understanding, and Mitigating Unwanted Algorithmic Bias}.
\newblock \bibinfo{howpublished}{ArXiv}.
\newblock
\newblock
\shownote{abs/1810.01943.}


\bibitem[\protect\citeauthoryear{Berk, Heidari, Jabbari, Joseph, Kearns,
  Morgenstern, Neel, and Roth}{Berk et~al\mbox{.}}{2017}]%
        {berk2017convex}
\bibfield{author}{\bibinfo{person}{R. Berk}, \bibinfo{person}{H. Heidari},
  \bibinfo{person}{S. Jabbari}, \bibinfo{person}{Matthew Joseph},
  \bibinfo{person}{M. Kearns}, \bibinfo{person}{J. Morgenstern},
  \bibinfo{person}{Seth Neel}, {and} \bibinfo{person}{A. Roth}.}
  \bibinfo{year}{2017}\natexlab{}.
\newblock \bibinfo{title}{A Convex Framework for Fair Regression}.
\newblock \bibinfo{howpublished}{ArXiv}.
\newblock
\newblock
\shownote{abs/1706.02409.}


\bibitem[\protect\citeauthoryear{Beutel, Chen, Zhao, and hsin Chi}{Beutel
  et~al\mbox{.}}{2017}]%
        {Beutel2017DataDA}
\bibfield{author}{\bibinfo{person}{Alex Beutel}, \bibinfo{person}{Jilin Chen},
  \bibinfo{person}{Zhe Zhao}, {and} \bibinfo{person}{Ed~Huai hsin Chi}.}
  \bibinfo{year}{2017}\natexlab{}.
\newblock \bibinfo{title}{Data Decisions and Theoretical Implications when
  Adversarially Learning Fair Representations}.
\newblock \bibinfo{howpublished}{ArXiv}.
\newblock
\newblock
\shownote{abs/1707.00075.}


\bibitem[\protect\citeauthoryear{Black, Yeom, and Fredrikson}{Black
  et~al\mbox{.}}{2020}]%
        {fliptest:2020}
\bibfield{author}{\bibinfo{person}{Emily Black}, \bibinfo{person}{Samuel Yeom},
  {and} \bibinfo{person}{Matt Fredrikson}.} \bibinfo{year}{2020}\natexlab{}.
\newblock \showarticletitle{FlipTest: Fairness Testing via Optimal Transport}.
  In \bibinfo{booktitle}{\emph{Proceedings of the 2020 Conference on Fairness,
  Accountability, and Transparency}} (Barcelona, Spain)
  \emph{(\bibinfo{series}{FAT* '20})}. \bibinfo{publisher}{Association for
  Computing Machinery}, \bibinfo{address}{New York, NY, USA},
  \bibinfo{pages}{111--121}.
\newblock
\showISBNx{9781450369367}
\urldef\tempurl%
\url{https://doi.org/10.1145/3351095.3372845}
\showDOI{\tempurl}


\bibitem[\protect\citeauthoryear{Blum and Stangl}{Blum and Stangl}{2020}]%
        {Blum2019RecoveringFB}
\bibfield{author}{\bibinfo{person}{Avrim Blum} {and} \bibinfo{person}{Kevin
  Stangl}.} \bibinfo{year}{2020}\natexlab{}.
\newblock \showarticletitle{{Recovering from Biased Data: Can Fairness
  Constraints Improve Accuracy?}}. In \bibinfo{booktitle}{\emph{1st Symposium
  on Foundations of Responsible Computing (FORC 2020)}}
  \emph{(\bibinfo{series}{Leibniz International Proceedings in Informatics
  (LIPIcs)}, Vol.~\bibinfo{volume}{156})}. \bibinfo{publisher}{Schloss
  Dagstuhl--Leibniz-Zentrum f{\"u}r Informatik}, \bibinfo{address}{Dagstuhl,
  Germany}, \bibinfo{pages}{3:1--3:20}.
\newblock
\urldef\tempurl%
\url{https://doi.org/10.4230/LIPIcs.FORC.2020.3}
\showDOI{\tempurl}


\bibitem[\protect\citeauthoryear{Bogen, Rieke, and Ahmed}{Bogen
  et~al\mbox{.}}{2020}]%
        {Bogen:2020}
\bibfield{author}{\bibinfo{person}{Miranda Bogen}, \bibinfo{person}{Aaron
  Rieke}, {and} \bibinfo{person}{Shazeda Ahmed}.}
  \bibinfo{year}{2020}\natexlab{}.
\newblock \showarticletitle{Awareness in Practice: Tensions in Access to
  Sensitive Attribute Data for Antidiscrimination}. In
  \bibinfo{booktitle}{\emph{Proceedings of the 2020 Conference on Fairness,
  Accountability, and Transparency}} (Barcelona, Spain)
  \emph{(\bibinfo{series}{FAT* '20})}. \bibinfo{publisher}{Association for
  Computing Machinery}, \bibinfo{address}{New York, NY, USA},
  \bibinfo{pages}{492--500}.
\newblock
\showISBNx{9781450369367}
\urldef\tempurl%
\url{https://doi.org/10.1145/3351095.3372877}
\showDOI{\tempurl}


\bibitem[\protect\citeauthoryear{Calders, Kamiran, and Pechenizkiy}{Calders
  et~al\mbox{.}}{2009}]%
        {Calders-building-2009}
\bibfield{author}{\bibinfo{person}{Toon Calders}, \bibinfo{person}{Faisal
  Kamiran}, {and} \bibinfo{person}{Mykola Pechenizkiy}.}
  \bibinfo{year}{2009}\natexlab{}.
\newblock \showarticletitle{Building Classifiers with Independency
  Constraints}. In \bibinfo{booktitle}{\emph{Proceedings of the 2009 IEEE
  International Conference on Data Mining Workshops}}
  \emph{(\bibinfo{series}{ICDMW '09})}. \bibinfo{publisher}{IEEE Computer
  Society}, \bibinfo{address}{USA}, \bibinfo{pages}{13--18}.
\newblock
\showISBNx{9780769539027}
\urldef\tempurl%
\url{https://doi.org/10.1109/ICDMW.2009.83}
\showDOI{\tempurl}


\bibitem[\protect\citeauthoryear{Calders and Verwer}{Calders and
  Verwer}{2010}]%
        {Calders-Naive-2010}
\bibfield{author}{\bibinfo{person}{Toon Calders} {and} \bibinfo{person}{Sicco
  Verwer}.} \bibinfo{year}{2010}\natexlab{}.
\newblock \showarticletitle{Three naive {Bayes} approaches for
  discrimination-free classification}.
\newblock \bibinfo{journal}{\emph{Data Mining and Knowledge Discovery}}
  \bibinfo{volume}{21}, \bibinfo{number}{2} (\bibinfo{date}{01 Sep}
  \bibinfo{year}{2010}), \bibinfo{pages}{277--292}.
\newblock
\showISSN{1573-756X}
\urldef\tempurl%
\url{https://doi.org/10.1007/s10618-010-0190-x}
\showDOI{\tempurl}


\bibitem[\protect\citeauthoryear{Calmon, Wei, Vinzamuri, Ramamurthy, and
  Varshney}{Calmon et~al\mbox{.}}{2017}]%
        {calmon_optimized}
\bibfield{author}{\bibinfo{person}{Flavio~P. Calmon}, \bibinfo{person}{Dennis
  Wei}, \bibinfo{person}{Bhanukiran Vinzamuri},
  \bibinfo{person}{Karthikeyan~Natesan Ramamurthy}, {and}
  \bibinfo{person}{Kush~R. Varshney}.} \bibinfo{year}{2017}\natexlab{}.
\newblock \showarticletitle{Optimized Pre-Processing for Discrimination
  Prevention}. In \bibinfo{booktitle}{\emph{Proceedings of the 31st
  International Conference on Neural Information Processing Systems}} (Long
  Beach, California, USA) \emph{(\bibinfo{series}{NIPS'17})}.
  \bibinfo{publisher}{Curran Associates Inc.}, \bibinfo{address}{Red Hook, NY,
  USA}, \bibinfo{pages}{3995--4004}.
\newblock
\showISBNx{9781510860964}


\bibitem[\protect\citeauthoryear{Card}{Card}{2017}]%
        {blackbox1}
\bibfield{author}{\bibinfo{person}{Dallas Card}.}
  \bibinfo{year}{2017}\natexlab{}.
\newblock \bibinfo{title}{The ``black box'' metaphor in machine learning}.
\newblock
  \bibinfo{howpublished}{\url{https://towardsdatascience.com/the-black-box-metaphor-in-machine-learning-4e57a3a1d2b0}}.
\newblock
\newblock
\shownote{[Online; accessed 13-April-2020].}


\bibitem[\protect\citeauthoryear{Card, Zhang, and Smith}{Card
  et~al\mbox{.}}{2019}]%
        {Card2018DeepWA}
\bibfield{author}{\bibinfo{person}{Dallas Card}, \bibinfo{person}{Michael
  Zhang}, {and} \bibinfo{person}{Noah~A. Smith}.}
  \bibinfo{year}{2019}\natexlab{}.
\newblock \showarticletitle{Deep Weighted Averaging Classifiers}. In
  \bibinfo{booktitle}{\emph{Proceedings of the Conference on Fairness,
  Accountability, and Transparency}} (Atlanta, GA, USA)
  \emph{(\bibinfo{series}{FAT* '19})}. \bibinfo{publisher}{Association for
  Computing Machinery}, \bibinfo{address}{New York, NY, USA},
  \bibinfo{pages}{369--378}.
\newblock
\showISBNx{9781450361255}
\urldef\tempurl%
\url{https://doi.org/10.1145/3287560.3287595}
\showDOI{\tempurl}


\bibitem[\protect\citeauthoryear{Celis, Huang, Keswani, and Vishnoi}{Celis
  et~al\mbox{.}}{2019}]%
        {Celis2018ClassificationWF}
\bibfield{author}{\bibinfo{person}{L.~Elisa Celis}, \bibinfo{person}{Lingxiao
  Huang}, \bibinfo{person}{Vijay Keswani}, {and} \bibinfo{person}{Nisheeth~K.
  Vishnoi}.} \bibinfo{year}{2019}\natexlab{}.
\newblock \showarticletitle{Classification with Fairness Constraints: A
  Meta-Algorithm with Provable Guarantees}. In
  \bibinfo{booktitle}{\emph{Proceedings of the Conference on Fairness,
  Accountability, and Transparency}} (Atlanta, GA, USA)
  \emph{(\bibinfo{series}{FAT* '19})}. \bibinfo{publisher}{Association for
  Computing Machinery}, \bibinfo{address}{New York, NY, USA},
  \bibinfo{pages}{319--328}.
\newblock
\showISBNx{9781450361255}
\urldef\tempurl%
\url{https://doi.org/10.1145/3287560.3287586}
\showDOI{\tempurl}


\bibitem[\protect\citeauthoryear{Chouldechova}{Chouldechova}{2017}]%
        {Chouldechova2016FairPW}
\bibfield{author}{\bibinfo{person}{Alexandra Chouldechova}.}
  \bibinfo{year}{2017}\natexlab{}.
\newblock \showarticletitle{Fair prediction with disparate impact: A study of
  bias in recidivism prediction instruments}.
\newblock \bibinfo{journal}{\emph{Big data}}  \bibinfo{volume}{5}
  (\bibinfo{date}{June} \bibinfo{year}{2017}), \bibinfo{pages}{153--163}.
\newblock


\bibitem[\protect\citeauthoryear{Corbett-Davies, Pierson, Feller, Goel, and
  Huq}{Corbett-Davies et~al\mbox{.}}{2017}]%
        {CorbettDavies2017AlgorithmicDM}
\bibfield{author}{\bibinfo{person}{Sam Corbett-Davies}, \bibinfo{person}{Emma
  Pierson}, \bibinfo{person}{Avi Feller}, \bibinfo{person}{Sharad Goel}, {and}
  \bibinfo{person}{Aziz Huq}.} \bibinfo{year}{2017}\natexlab{}.
\newblock \showarticletitle{Algorithmic Decision Making and the Cost of
  Fairness}. In \bibinfo{booktitle}{\emph{Proceedings of the 23rd ACM SIGKDD
  International Conference on Knowledge Discovery and Data Mining}} (Halifax,
  NS, Canada) \emph{(\bibinfo{series}{KDD '17})}.
  \bibinfo{publisher}{Association for Computing Machinery},
  \bibinfo{address}{New York, NY, USA}, \bibinfo{pages}{797--806}.
\newblock
\showISBNx{9781450348874}
\urldef\tempurl%
\url{https://doi.org/10.1145/3097983.3098095}
\showDOI{\tempurl}


\bibitem[\protect\citeauthoryear{Coston, Mishler, Kennedy, and
  Chouldechova}{Coston et~al\mbox{.}}{2020}]%
        {biased_data_1}
\bibfield{author}{\bibinfo{person}{Amanda Coston}, \bibinfo{person}{Alan
  Mishler}, \bibinfo{person}{Edward~H. Kennedy}, {and}
  \bibinfo{person}{Alexandra Chouldechova}.} \bibinfo{year}{2020}\natexlab{}.
\newblock \showarticletitle{Counterfactual Risk Assessments, Evaluation, and
  Fairness}. In \bibinfo{booktitle}{\emph{Proceedings of the 2020 Conference on
  Fairness, Accountability, and Transparency}} (Barcelona, Spain)
  \emph{(\bibinfo{series}{FAT* '20})}. \bibinfo{publisher}{Association for
  Computing Machinery}, \bibinfo{address}{New York, NY, USA},
  \bibinfo{pages}{582--593}.
\newblock
\showISBNx{9781450369367}
\urldef\tempurl%
\url{https://doi.org/10.1145/3351095.3372851}
\showDOI{\tempurl}


\bibitem[\protect\citeauthoryear{Crosman}{Crosman}{2018}]%
        {biased-loan3}
\bibfield{author}{\bibinfo{person}{Penny Crosman}.}
  \bibinfo{year}{2018}\natexlab{}.
\newblock \bibinfo{title}{Weren't algorithms supposed to make digital mortgages
  colorblind?}
\newblock
  \bibinfo{howpublished}{\url{https://www.americanbanker.com/news/werent-algorithms-supposed-to-make-digital-mortgages-colorblind}}.
\newblock
\newblock
\shownote{[Online; accessed 13-April-2020].}


\bibitem[\protect\citeauthoryear{Dastin}{Dastin}{2018}]%
        {biased-hiring1}
\bibfield{author}{\bibinfo{person}{Jeffrey Dastin}.}
  \bibinfo{year}{2018}\natexlab{}.
\newblock \bibinfo{title}{Amazon scraps secret {AI} recruiting tool that showed
  bias against women}.
\newblock
  \bibinfo{howpublished}{\url{https://www.reuters.com/misc/us-amazon-com-jobs-automation-insight/amazon-scraps-secret-ai-recruiting-tool-that-showed-bias-against-women-idUSKCN1MK08G}}.
\newblock
\newblock
\shownote{[Online; accessed 13-April-2020].}


\bibitem[\protect\citeauthoryear{Dent}{Dent}{2019}]%
        {weakened_appeal}
\bibfield{author}{\bibinfo{person}{Kyle Dent}.}
  \bibinfo{year}{2019}\natexlab{}.
\newblock \bibinfo{title}{The risks of amoral AI}.
\newblock
  \bibinfo{howpublished}{\url{https://techcrunch.com/2019/08/25/the-risks-of-amoral-a-i/}}.
\newblock
\newblock
\shownote{[Online; accessed 13-Septmeber-2020].}


\bibitem[\protect\citeauthoryear{Didur}{Didur}{2018}]%
        {finance_2}
\bibfield{author}{\bibinfo{person}{Konstantin Didur}.}
  \bibinfo{year}{2018}\natexlab{}.
\newblock \bibinfo{title}{12 Use Cases of AI and Machine Learning In Finance}.
\newblock
  \bibinfo{howpublished}{\url{https://towardsdatascience.com/machine-learning-in-finance-why-what-how-d524a2357b56}}.
\newblock
\newblock
\shownote{[Online; accessed 13-Septmeber-2020].}


\bibitem[\protect\citeauthoryear{Donini, Oneto, Ben-David, Shawe-Taylor, and
  Pontil}{Donini et~al\mbox{.}}{2018}]%
        {Donini-RM:2018}
\bibfield{author}{\bibinfo{person}{Michele Donini}, \bibinfo{person}{Luca
  Oneto}, \bibinfo{person}{Shai Ben-David}, \bibinfo{person}{John
  Shawe-Taylor}, {and} \bibinfo{person}{Massimiliano Pontil}.}
  \bibinfo{year}{2018}\natexlab{}.
\newblock \showarticletitle{Empirical Risk Minimization under Fairness
  Constraints}. In \bibinfo{booktitle}{\emph{Proceedings of the 32nd
  International Conference on Neural Information Processing Systems}}
  (Montr\'{e}al, Canada) \emph{(\bibinfo{series}{NIPS'18})}.
  \bibinfo{publisher}{Curran Associates Inc.}, \bibinfo{address}{Red Hook, NY,
  USA}, \bibinfo{pages}{2796--2806}.
\newblock


\bibitem[\protect\citeauthoryear{Dua and Graff}{Dua and Graff}{2017a}]%
        {UCI-repo-adult}
\bibfield{author}{\bibinfo{person}{Dheeru Dua} {and} \bibinfo{person}{Casey
  Graff}.} \bibinfo{year}{2017}\natexlab{a}.
\newblock \bibinfo{title}{{UCI} Machine Learning Repository - Adult Income}.
\newblock
\newblock
\urldef\tempurl%
\url{http://archive.ics.uci.edu/ml/datasets/Adult}
\showURL{%
\tempurl}


\bibitem[\protect\citeauthoryear{Dua and Graff}{Dua and Graff}{2017b}]%
        {UCI-repo-default}
\bibfield{author}{\bibinfo{person}{Dheeru Dua} {and} \bibinfo{person}{Casey
  Graff}.} \bibinfo{year}{2017}\natexlab{b}.
\newblock \bibinfo{title}{{UCI} Machine Learning Repository - Default
  Prediction}.
\newblock
\newblock
\urldef\tempurl%
\url{http://archive.ics.uci.edu/ml/datasets/default+of+credit+card+clients}
\showURL{%
\tempurl}


\bibitem[\protect\citeauthoryear{Dua and Graff}{Dua and Graff}{2017c}]%
        {UCI-repo-german}
\bibfield{author}{\bibinfo{person}{Dheeru Dua} {and} \bibinfo{person}{Casey
  Graff}.} \bibinfo{year}{2017}\natexlab{c}.
\newblock \bibinfo{title}{{UCI} Machine Learning Repository - German Credit}.
\newblock
\newblock
\urldef\tempurl%
\url{http://archive.ics.uci.edu/ml/datasets/statlog+(german+credit+data)}
\showURL{%
\tempurl}


\bibitem[\protect\citeauthoryear{Dua and Graff}{Dua and Graff}{2017d}]%
        {UCI-repo-student}
\bibfield{author}{\bibinfo{person}{Dheeru Dua} {and} \bibinfo{person}{Casey
  Graff}.} \bibinfo{year}{2017}\natexlab{d}.
\newblock \bibinfo{title}{{UCI} Machine Learning Repository - Student
  Performance}.
\newblock
\newblock
\urldef\tempurl%
\url{http://archive.ics.uci.edu/ml/datasets/Student%2BPerformance}
\showURL{%
\tempurl}


\bibitem[\protect\citeauthoryear{Dunkelau and Leuschel}{Dunkelau and
  Leuschel}{2019}]%
        {dunkelau_fairness-aware}
\bibfield{author}{\bibinfo{person}{Jannik Dunkelau} {and}
  \bibinfo{person}{Michael Leuschel}.} \bibinfo{year}{2019}\natexlab{}.
\newblock \bibinfo{title}{Fairness-{Aware} {Machine} {Learning}}.
\newblock , \bibinfo{numpages}{60}~pages.
\newblock


\bibitem[\protect\citeauthoryear{Dwork, Hardt, Pitassi, Reingold, and
  Zemel}{Dwork et~al\mbox{.}}{2012}]%
        {dwork_fairness_2011}
\bibfield{author}{\bibinfo{person}{Cynthia Dwork}, \bibinfo{person}{Moritz
  Hardt}, \bibinfo{person}{Toniann Pitassi}, \bibinfo{person}{Omer Reingold},
  {and} \bibinfo{person}{Richard Zemel}.} \bibinfo{year}{2012}\natexlab{}.
\newblock \showarticletitle{Fairness through Awareness}. In
  \bibinfo{booktitle}{\emph{Proceedings of the 3rd Innovations in Theoretical
  Computer Science Conference}} (Cambridge, Massachusetts)
  \emph{(\bibinfo{series}{ITCS '12})}. \bibinfo{publisher}{Association for
  Computing Machinery}, \bibinfo{address}{New York, NY, USA},
  \bibinfo{pages}{214--226}.
\newblock
\showISBNx{9781450311151}
\urldef\tempurl%
\url{https://doi.org/10.1145/2090236.2090255}
\showDOI{\tempurl}


\bibitem[\protect\citeauthoryear{Dwork, Immorlica, Kalai, and Leiserson}{Dwork
  et~al\mbox{.}}{2018}]%
        {Dwork-DC:2018}
\bibfield{author}{\bibinfo{person}{Cynthia Dwork}, \bibinfo{person}{Nicole
  Immorlica}, \bibinfo{person}{Adam~Tauman Kalai}, {and} \bibinfo{person}{Max
  Leiserson}.} \bibinfo{year}{2018}\natexlab{}.
\newblock \showarticletitle{Decoupled Classifiers for Group-Fair and Efficient
  Machine Learning}. In \bibinfo{booktitle}{\emph{Proceedings of Machine
  Learning Research}}, \bibfield{editor}{\bibinfo{person}{Sorelle~A. Friedler}
  {and} \bibinfo{person}{Christo Wilson}} (Eds.), Vol.~\bibinfo{volume}{81}.
  \bibinfo{publisher}{PMLR}, \bibinfo{address}{New York, NY, USA},
  \bibinfo{pages}{119--133}.
\newblock
\urldef\tempurl%
\url{http://proceedings.mlr.press/v81/dwork18a.html}
\showURL{%
\tempurl}


\bibitem[\protect\citeauthoryear{Edwards and Storkey}{Edwards and
  Storkey}{2015}]%
        {Edwards2015CensoringRW}
\bibfield{author}{\bibinfo{person}{Harrison~A Edwards} {and}
  \bibinfo{person}{Amos~J. Storkey}.} \bibinfo{year}{2015}\natexlab{}.
\newblock \bibinfo{title}{Censoring Representations with an Adversary}.
\newblock \bibinfo{howpublished}{CoRR}.
\newblock
\newblock
\shownote{abs/1511.05897.}


\bibitem[\protect\citeauthoryear{Feldman, Friedler, Moeller, Scheidegger, and
  Venkatasubramanian}{Feldman et~al\mbox{.}}{2015}]%
        {feldman_certifying_2015}
\bibfield{author}{\bibinfo{person}{Michael Feldman},
  \bibinfo{person}{Sorelle~A. Friedler}, \bibinfo{person}{John Moeller},
  \bibinfo{person}{Carlos Scheidegger}, {and} \bibinfo{person}{Suresh
  Venkatasubramanian}.} \bibinfo{year}{2015}\natexlab{}.
\newblock \showarticletitle{Certifying and Removing Disparate Impact}. In
  \bibinfo{booktitle}{\emph{Proceedings of the 21th ACM SIGKDD International
  Conference on Knowledge Discovery and Data Mining}} (Sydney, NSW, Australia)
  \emph{(\bibinfo{series}{KDD '15})}. \bibinfo{publisher}{Association for
  Computing Machinery}, \bibinfo{address}{New York, NY, USA},
  \bibinfo{pages}{259--268}.
\newblock
\showISBNx{9781450336642}
\urldef\tempurl%
\url{https://doi.org/10.1145/2783258.2783311}
\showDOI{\tempurl}


\bibitem[\protect\citeauthoryear{Fish, Kun, and Lelkes}{Fish
  et~al\mbox{.}}{2016}]%
        {Fish2016ACA}
\bibfield{author}{\bibinfo{person}{Benjamin Fish}, \bibinfo{person}{Jeremy
  Kun}, {and} \bibinfo{person}{{\'A}d{\'a}m~D{\'a}niel Lelkes}.}
  \bibinfo{year}{2016}\natexlab{}.
\newblock \showarticletitle{A Confidence-Based Approach for Balancing Fairness
  and Accuracy}. In \bibinfo{booktitle}{\emph{Proceedings of the 2016 SIAM
  International Conference on Data Mining}}. \bibinfo{publisher}{Society for
  Industrial and Applied Mathematics}, \bibinfo{address}{3600 University City
  Science Center Philadelphia, PA, United States}, \bibinfo{pages}{144--152}.
\newblock
\urldef\tempurl%
\url{https://doi.org/10.1137/1.9781611974348.17}
\showDOI{\tempurl}


\bibitem[\protect\citeauthoryear{Friedler, Scheidegger, Venkatasubramanian,
  Choudhary, Hamilton, and Roth}{Friedler et~al\mbox{.}}{2019}]%
        {friedler_comparative_2018}
\bibfield{author}{\bibinfo{person}{Sorelle~A. Friedler},
  \bibinfo{person}{Carlos Scheidegger}, \bibinfo{person}{Suresh
  Venkatasubramanian}, \bibinfo{person}{Sonam Choudhary},
  \bibinfo{person}{Evan~P. Hamilton}, {and} \bibinfo{person}{Derek Roth}.}
  \bibinfo{year}{2019}\natexlab{}.
\newblock \showarticletitle{A Comparative Study of Fairness-Enhancing
  Interventions in Machine Learning}. In \bibinfo{booktitle}{\emph{Proceedings
  of the Conference on Fairness, Accountability, and Transparency}} (Atlanta,
  GA, USA) \emph{(\bibinfo{series}{FAT* '19})}. \bibinfo{publisher}{Association
  for Computing Machinery}, \bibinfo{address}{New York, NY, USA},
  \bibinfo{pages}{329--338}.
\newblock
\showISBNx{9781450361255}
\urldef\tempurl%
\url{https://doi.org/10.1145/3287560.3287589}
\showDOI{\tempurl}


\bibitem[\protect\citeauthoryear{Galhotra, Brun, and Meliou}{Galhotra
  et~al\mbox{.}}{2017}]%
        {galhotra_fairness_2017}
\bibfield{author}{\bibinfo{person}{Sainyam Galhotra}, \bibinfo{person}{Yuriy
  Brun}, {and} \bibinfo{person}{Alexandra Meliou}.}
  \bibinfo{year}{2017}\natexlab{}.
\newblock \showarticletitle{Fairness Testing: Testing Software for
  Discrimination}. In \bibinfo{booktitle}{\emph{Proceedings of the 2017 11th
  Joint Meeting on Foundations of Software Engineering}} (Paderborn, Germany)
  \emph{(\bibinfo{series}{ESEC/FSE 2017})}. \bibinfo{publisher}{Association for
  Computing Machinery}, \bibinfo{address}{New York, NY, USA},
  \bibinfo{pages}{498--510}.
\newblock
\showISBNx{9781450351058}
\urldef\tempurl%
\url{https://doi.org/10.1145/3106237.3106277}
\showDOI{\tempurl}


\bibitem[\protect\citeauthoryear{Geiger, Yu, Yang, Dai, Qiu, Tang, and
  Huang}{Geiger et~al\mbox{.}}{2020}]%
        {garbageinout}
\bibfield{author}{\bibinfo{person}{R.~Stuart Geiger}, \bibinfo{person}{Kevin
  Yu}, \bibinfo{person}{Yanlai Yang}, \bibinfo{person}{Mindy Dai},
  \bibinfo{person}{Jie Qiu}, \bibinfo{person}{Rebekah Tang}, {and}
  \bibinfo{person}{Jenny Huang}.} \bibinfo{year}{2020}\natexlab{}.
\newblock \showarticletitle{Garbage in, Garbage out? Do Machine Learning
  Application Papers in Social Computing Report Where Human-Labeled Training
  Data Comes From?}. In \bibinfo{booktitle}{\emph{Proceedings of the 2020
  Conference on Fairness, Accountability, and Transparency}} (Barcelona, Spain)
  \emph{(\bibinfo{series}{FAT* '20})}. \bibinfo{publisher}{Association for
  Computing Machinery}, \bibinfo{address}{New York, NY, USA},
  \bibinfo{pages}{325--336}.
\newblock
\showISBNx{9781450369367}
\urldef\tempurl%
\url{https://doi.org/10.1145/3351095.3372862}
\showDOI{\tempurl}


\bibitem[\protect\citeauthoryear{Goel, Rao, and Shroff}{Goel
  et~al\mbox{.}}{2016}]%
        {NYC_policing}
\bibfield{author}{\bibinfo{person}{Sharad Goel}, \bibinfo{person}{Justin~M.
  Rao}, {and} \bibinfo{person}{Ravi Shroff}.} \bibinfo{year}{2016}\natexlab{}.
\newblock \showarticletitle{Precinct or prejudice? {Understanding} racial
  disparities in {New} {York} {City}'s stop-and-frisk policy}.
\newblock \bibinfo{journal}{\emph{The Annals of Applied Statistics}}
  \bibinfo{volume}{10}, \bibinfo{number}{1} (\bibinfo{year}{2016}),
  \bibinfo{pages}{365--394}.
\newblock
\showISSN{19326157}
\urldef\tempurl%
\url{http://www.jstor.org/stable/43826483}
\showURL{%
\tempurl}


\bibitem[\protect\citeauthoryear{Greenwood}{Greenwood}{2020}]%
        {redlining-thinkprogress}
\bibfield{author}{\bibinfo{person}{Shannon Greenwood}.}
  \bibinfo{year}{2020}\natexlab{}.
\newblock \bibinfo{title}{It'll Be Hard To Get Approved For A Mortgage If
  You're Black}.
\newblock
  \bibinfo{howpublished}{\url{https://thinkprogress.org/itll-be-hard-to-get-approved-for-a-mortgage-if-you-re-black-43634b642bdf/}}.
\newblock
\newblock
\shownote{[Online; accessed 13-April-2020].}


\bibitem[\protect\citeauthoryear{Hacker and Wiedemann}{Hacker and
  Wiedemann}{2017}]%
        {Hacker-continuous-2017}
\bibfield{author}{\bibinfo{person}{Philipp Hacker} {and} \bibinfo{person}{Emil
  Wiedemann}.} \bibinfo{year}{2017}\natexlab{}.
\newblock \showarticletitle{A continuous framework for fairness}.
\newblock \bibinfo{journal}{\emph{ArXiv}} (\bibinfo{year}{2017}).
\newblock
\newblock
\shownote{abs/1712.07924.}


\bibitem[\protect\citeauthoryear{Hajian and Domingo-Ferrer}{Hajian and
  Domingo-Ferrer}{2013}]%
        {Hajian-method-2013}
\bibfield{author}{\bibinfo{person}{Sara Hajian} {and} \bibinfo{person}{Josep
  Domingo-Ferrer}.} \bibinfo{year}{2013}\natexlab{}.
\newblock \showarticletitle{Direct and Indirect Discrimination Prevention
  Methods}. In \bibinfo{booktitle}{\emph{Discrimination and Privacy in the
  Information Society: Data Mining and Profiling in Large Databases}}.
  \bibinfo{publisher}{Springer Berlin Heidelberg}, \bibinfo{address}{Berlin,
  Heidelberg}, \bibinfo{pages}{241--254}.
\newblock
\showISBNx{978-3-642-30487-3}
\urldef\tempurl%
\url{https://doi.org/10.1007/978-3-642-30487-3_13}
\showDOI{\tempurl}


\bibitem[\protect\citeauthoryear{Hajian, Domingo-Ferrer, and
  Mart\'{\i}nez-Ballest\'{e}}{Hajian et~al\mbox{.}}{2011}]%
        {Hajian-rule-2011}
\bibfield{author}{\bibinfo{person}{Sara Hajian}, \bibinfo{person}{Josep
  Domingo-Ferrer}, {and} \bibinfo{person}{Antoni Mart\'{\i}nez-Ballest\'{e}}.}
  \bibinfo{year}{2011}\natexlab{}.
\newblock \showarticletitle{Rule Protection for Indirect Discrimination
  Prevention in Data Mining}. In \bibinfo{booktitle}{\emph{Proceedings of the
  8th International Conference on Modeling Decisions for Artificial
  Intelligence}} (Changsha, China) \emph{(\bibinfo{series}{MDAI'11})}.
  \bibinfo{publisher}{Springer-Verlag}, \bibinfo{address}{Berlin, Heidelberg},
  \bibinfo{pages}{211--222}.
\newblock
\showISBNx{9783642225888}


\bibitem[\protect\citeauthoryear{Hamilton}{Hamilton}{2019}]%
        {biased-loan1}
\bibfield{author}{\bibinfo{person}{Ernest Hamilton}.}
  \bibinfo{year}{2019}\natexlab{}.
\newblock \bibinfo{title}{AI Perpetuating Human Bias In The Lending Space}.
\newblock
  \bibinfo{howpublished}{\url{https://www.techtimes.com/articles/240769/20190402/ai-perpetuating-human-bias-in-the-lending-space}}.
\newblock
\newblock
\shownote{[Online; accessed 13-April-2020].}


\bibitem[\protect\citeauthoryear{Hardt, Price, and Srebro}{Hardt
  et~al\mbox{.}}{2016}]%
        {Hardt2016EqualityOO}
\bibfield{author}{\bibinfo{person}{Moritz Hardt}, \bibinfo{person}{Eric Price},
  {and} \bibinfo{person}{Nathan Srebro}.} \bibinfo{year}{2016}\natexlab{}.
\newblock \showarticletitle{Equality of Opportunity in Supervised Learning}. In
  \bibinfo{booktitle}{\emph{Proceedings of the 30th International Conference on
  Neural Information Processing Systems}} (Barcelona, Spain)
  \emph{(\bibinfo{series}{NIPS'16})}. \bibinfo{publisher}{Curran Associates
  Inc.}, \bibinfo{address}{Red Hook, NY, USA}, \bibinfo{pages}{3323--3331}.
\newblock
\showISBNx{9781510838819}


\bibitem[\protect\citeauthoryear{Hu and Chen}{Hu and Chen}{2020}]%
        {pareto_1}
\bibfield{author}{\bibinfo{person}{Lily Hu} {and} \bibinfo{person}{Yiling
  Chen}.} \bibinfo{year}{2020}\natexlab{}.
\newblock \showarticletitle{Fair Classification and Social Welfare}. In
  \bibinfo{booktitle}{\emph{Proceedings of the 2020 Conference on Fairness,
  Accountability, and Transparency}} (Barcelona, Spain)
  \emph{(\bibinfo{series}{FAT* '20})}. \bibinfo{publisher}{Association for
  Computing Machinery}, \bibinfo{address}{New York, NY, USA},
  \bibinfo{pages}{535--545}.
\newblock
\showISBNx{9781450369367}
\urldef\tempurl%
\url{https://doi.org/10.1145/3351095.3372857}
\showDOI{\tempurl}


\bibitem[\protect\citeauthoryear{Hu and Kohler-Hausmann}{Hu and
  Kohler-Hausmann}{2020}]%
        {sexgottodo:2020}
\bibfield{author}{\bibinfo{person}{Lily Hu} {and} \bibinfo{person}{Issa
  Kohler-Hausmann}.} \bibinfo{year}{2020}\natexlab{}.
\newblock \showarticletitle{What's Sex Got to Do with Machine Learning?}. In
  \bibinfo{booktitle}{\emph{Proceedings of the 2020 Conference on Fairness,
  Accountability, and Transparency}} (Barcelona, Spain)
  \emph{(\bibinfo{series}{FAT* '20})}. \bibinfo{publisher}{Association for
  Computing Machinery}, \bibinfo{address}{New York, NY, USA},
  \bibinfo{pages}{513}.
\newblock
\showISBNx{9781450369367}
\urldef\tempurl%
\url{https://doi.org/10.1145/3351095.3375674}
\showDOI{\tempurl}


\bibitem[\protect\citeauthoryear{Jeff~Larson and Angwin}{Jeff~Larson and
  Angwin}{2016}]%
        {compas-data}
\bibfield{author}{\bibinfo{person}{Lauren~Kirchner Jeff~Larson, Surya~Mattu}
  {and} \bibinfo{person}{Julia Angwin}.} \bibinfo{year}{2016}\natexlab{}.
\newblock \bibinfo{title}{UCI Machine Learning Repository}.
\newblock
\newblock
\urldef\tempurl%
\url{https://www.propublica.org/datastore/dataset/compas-recidivism-risk-score-data-and-analysis}
\showURL{%
\tempurl}


\bibitem[\protect\citeauthoryear{Johndrow and Lum}{Johndrow and Lum}{2019}]%
        {Johndrow2017AnAF}
\bibfield{author}{\bibinfo{person}{James~E. Johndrow} {and}
  \bibinfo{person}{Kristian Lum}.} \bibinfo{year}{2019}\natexlab{}.
\newblock \showarticletitle{An algorithm for removing sensitive information:
  Application to race-independent recidivism prediction}.
\newblock \bibinfo{journal}{\emph{Ann. Appl. Stat.}} \bibinfo{volume}{13},
  \bibinfo{number}{1} (\bibinfo{date}{03} \bibinfo{year}{2019}),
  \bibinfo{pages}{189--220}.
\newblock
\urldef\tempurl%
\url{https://doi.org/10.1214/18-AOAS1201}
\showDOI{\tempurl}


\bibitem[\protect\citeauthoryear{Jolly-Ryan}{Jolly-Ryan}{2012}]%
        {jolly-ryan-have}
\bibfield{author}{\bibinfo{person}{Jennifer Jolly-Ryan}.}
  \bibinfo{year}{2012}\natexlab{}.
\newblock \showarticletitle{Have a Job to Get a Job: Disparate Treatment and
  Disparate Impact of the 'Currently Employed' Requirement}.
\newblock \bibinfo{journal}{\emph{Michigan Journal of Race and Law}}
  \bibinfo{volume}{18} (\bibinfo{year}{2012}), 25.
\newblock
\urldef\tempurl%
\url{https://repository.law.umich.edu/mjrl/vol18/iss1/4}
\showURL{%
\tempurl}


\bibitem[\protect\citeauthoryear{Julia~Angwin and Kirchner}{Julia~Angwin and
  Kirchner}{2016}]%
        {propublica-main}
\bibfield{author}{\bibinfo{person}{Surya~Mattu Julia~Angwin, Jeff~Larson} {and}
  \bibinfo{person}{Lauren Kirchner}.} \bibinfo{year}{2016}\natexlab{}.
\newblock \bibinfo{title}{Machine Bias}.
\newblock
  \bibinfo{howpublished}{\url{https://www.propublica.org/article/machine-bias-risk-assessments-in-criminal-sentencing}}.
\newblock
\newblock
\shownote{[Online; accessed 13-April-2020].}


\bibitem[\protect\citeauthoryear{Jung, Kearns, Neel, Roth, Stapleton, and
  Wu}{Jung et~al\mbox{.}}{2019}]%
        {Jung2019ElicitingAE}
\bibfield{author}{\bibinfo{person}{Christopher Jung}, \bibinfo{person}{Michael
  Kearns}, \bibinfo{person}{Seth Neel}, \bibinfo{person}{Aaron Roth},
  \bibinfo{person}{Logan Stapleton}, {and} \bibinfo{person}{Zhiwei~Steven Wu}.}
  \bibinfo{year}{2019}\natexlab{}.
\newblock \bibinfo{title}{Eliciting and Enforcing Subjective Individual
  Fairness}.
\newblock \bibinfo{howpublished}{ArXiv}.
\newblock
\newblock
\shownote{abs/1905.10660.}


\bibitem[\protect\citeauthoryear{Kallus, Mao, and Zhou}{Kallus
  et~al\mbox{.}}{2020}]%
        {Kallus2020AssessingAF}
\bibfield{author}{\bibinfo{person}{Nathan Kallus}, \bibinfo{person}{Xiaojie
  Mao}, {and} \bibinfo{person}{Angela Zhou}.} \bibinfo{year}{2020}\natexlab{}.
\newblock \showarticletitle{Assessing Algorithmic Fairness with Unobserved
  Protected Class Using Data Combination}. In
  \bibinfo{booktitle}{\emph{Proceedings of the 2020 Conference on Fairness,
  Accountability, and Transparency}} (Barcelona, Spain)
  \emph{(\bibinfo{series}{FAT* '20})}. \bibinfo{publisher}{Association for
  Computing Machinery}, \bibinfo{address}{New York, NY, USA},
  \bibinfo{pages}{110}.
\newblock
\showISBNx{9781450369367}
\urldef\tempurl%
\url{https://doi.org/10.1145/3351095.3373154}
\showDOI{\tempurl}


\bibitem[\protect\citeauthoryear{Kamiran and Calders}{Kamiran and
  Calders}{2009}]%
        {Kamiran2009ClassifyingWD}
\bibfield{author}{\bibinfo{person}{Faisal Kamiran} {and} \bibinfo{person}{Toon
  Calders}.} \bibinfo{year}{2009}\natexlab{}.
\newblock \showarticletitle{Classifying without discriminating}. In
  \bibinfo{booktitle}{\emph{2009 2nd International Conference on Computer,
  Control and Communication}}. \bibinfo{publisher}{IEEE Computer Society},
  \bibinfo{address}{USA}, 6.
\newblock


\bibitem[\protect\citeauthoryear{Kamiran and Calders}{Kamiran and
  Calders}{2010}]%
        {Kamiran-no-discm-2010}
\bibfield{author}{\bibinfo{person}{Faisal Kamiran} {and} \bibinfo{person}{Toon
  Calders}.} \bibinfo{year}{2010}\natexlab{}.
\newblock \showarticletitle{Classification with no discrimination by
  preferential sampling}. In \bibinfo{booktitle}{\emph{Informal proceedings of
  the 19th Annual Machine Learning Conference of Belgium and The Netherlands
  (Benelearn'10, Leuven, Belgium, May 27-28, 2010)}}. 6.
\newblock


\bibitem[\protect\citeauthoryear{Kamiran and Calders}{Kamiran and
  Calders}{2012}]%
        {kamiran_data_2012}
\bibfield{author}{\bibinfo{person}{Faisal Kamiran} {and} \bibinfo{person}{Toon
  Calders}.} \bibinfo{year}{2012}\natexlab{}.
\newblock \showarticletitle{Data preprocessing techniques for classification
  without discrimination}.
\newblock \bibinfo{journal}{\emph{Knowledge and Information Systems}}
  \bibinfo{volume}{33}, \bibinfo{number}{1} (\bibinfo{date}{01 Oct}
  \bibinfo{year}{2012}), \bibinfo{pages}{1--33}.
\newblock
\showISSN{0219-3116}
\urldef\tempurl%
\url{https://doi.org/10.1007/s10115-011-0463-8}
\showDOI{\tempurl}


\bibitem[\protect\citeauthoryear{Kamiran, Calders, and Pechenizkiy}{Kamiran
  et~al\mbox{.}}{2010}]%
        {Kamiran-tree-2010}
\bibfield{author}{\bibinfo{person}{Faisal Kamiran}, \bibinfo{person}{Toon
  Calders}, {and} \bibinfo{person}{Mykola Pechenizkiy}.}
  \bibinfo{year}{2010}\natexlab{}.
\newblock \showarticletitle{Discrimination Aware Decision Tree Learning}. In
  \bibinfo{booktitle}{\emph{Proceedings of the 2010 IEEE International
  Conference on Data Mining}} \emph{(\bibinfo{series}{ICDM '10})}.
  \bibinfo{publisher}{IEEE Computer Society}, \bibinfo{address}{USA},
  \bibinfo{pages}{869--874}.
\newblock
\showISBNx{9780769542560}
\urldef\tempurl%
\url{https://doi.org/10.1109/ICDM.2010.50}
\showDOI{\tempurl}


\bibitem[\protect\citeauthoryear{Kamiran, Karim, and Zhang}{Kamiran
  et~al\mbox{.}}{2012}]%
        {Kamiran-DT-2012}
\bibfield{author}{\bibinfo{person}{Faisal Kamiran}, \bibinfo{person}{Asim
  Karim}, {and} \bibinfo{person}{Xiangliang Zhang}.}
  \bibinfo{year}{2012}\natexlab{}.
\newblock \showarticletitle{Decision Theory for Discrimination-Aware
  Classification}. In \bibinfo{booktitle}{\emph{Proceedings of the 2012 IEEE
  12th International Conference on Data Mining}} \emph{(\bibinfo{series}{ICDM
  '12})}. \bibinfo{publisher}{IEEE Computer Society}, \bibinfo{address}{USA},
  \bibinfo{pages}{924--929}.
\newblock
\showISBNx{9780769549057}
\urldef\tempurl%
\url{https://doi.org/10.1109/ICDM.2012.45}
\showDOI{\tempurl}


\bibitem[\protect\citeauthoryear{Kamishima, Akaho, Asoh, and Sakuma}{Kamishima
  et~al\mbox{.}}{2012}]%
        {Kamishima:2012}
\bibfield{author}{\bibinfo{person}{Toshihiro Kamishima},
  \bibinfo{person}{Shotaro Akaho}, \bibinfo{person}{Hideki Asoh}, {and}
  \bibinfo{person}{Jun Sakuma}.} \bibinfo{year}{2012}\natexlab{}.
\newblock \showarticletitle{Fairness-Aware Classifier with Prejudice Remover
  Regularizer}. In \bibinfo{booktitle}{\emph{Proceedings of the 2012th European
  Conference on Machine Learning and Knowledge Discovery in Databases - Volume
  Part II}} (Bristol, UK) \emph{(\bibinfo{series}{ECMLPKDD'12})}.
  \bibinfo{publisher}{Springer-Verlag}, \bibinfo{address}{Berlin, Heidelberg},
  \bibinfo{pages}{35--50}.
\newblock
\showISBNx{9783642334856}


\bibitem[\protect\citeauthoryear{Kamishima, Akaho, and Sakuma}{Kamishima
  et~al\mbox{.}}{2011}]%
        {Kamishima-RA:2011}
\bibfield{author}{\bibinfo{person}{Toshihiro Kamishima},
  \bibinfo{person}{Shotaro Akaho}, {and} \bibinfo{person}{Jun Sakuma}.}
  \bibinfo{year}{2011}\natexlab{}.
\newblock \showarticletitle{Fairness-Aware Learning through Regularization
  Approach}. In \bibinfo{booktitle}{\emph{Proceedings of the 2011 IEEE 11th
  International Conference on Data Mining Workshops}}
  \emph{(\bibinfo{series}{ICDMW '11})}. \bibinfo{publisher}{IEEE Computer
  Society}, \bibinfo{address}{USA}, \bibinfo{pages}{643--650}.
\newblock
\showISBNx{9780769544090}
\urldef\tempurl%
\url{https://doi.org/10.1109/ICDMW.2011.83}
\showDOI{\tempurl}


\bibitem[\protect\citeauthoryear{Kaplan and Hardy}{Kaplan and Hardy}{2020}]%
        {unfair_arrest_2}
\bibfield{author}{\bibinfo{person}{Joshua Kaplan} {and}
  \bibinfo{person}{Benjamin Hardy}.} \bibinfo{year}{2020}\natexlab{}.
\newblock \bibinfo{title}{Early Data Shows Black People Are Being
  Disproportionally Arrested for Social Distancing Violations}.
\newblock
  \bibinfo{howpublished}{\url{https://www.propublica.org/article/in-some-of-ohios-most-populous-areas-black-people-were-at-least-4-times-as-likely-to-be-charged-with-stay-at-home-violations-as-whites}}.
\newblock
\newblock
\shownote{[Online; accessed 13-April-2020].}


\bibitem[\protect\citeauthoryear{Kilbertus, Gascon, Kusner, Veale, Gummadi, and
  Weller}{Kilbertus et~al\mbox{.}}{2018}]%
        {kilbertus-blind-18}
\bibfield{author}{\bibinfo{person}{Niki Kilbertus}, \bibinfo{person}{Adria
  Gascon}, \bibinfo{person}{Matt Kusner}, \bibinfo{person}{Michael Veale},
  \bibinfo{person}{Krishna Gummadi}, {and} \bibinfo{person}{Adrian Weller}.}
  \bibinfo{year}{2018}\natexlab{}.
\newblock \showarticletitle{Blind Justice: Fairness with Encrypted Sensitive
  Attributes}. In \bibinfo{booktitle}{\emph{Proceedings of Machine Learning
  Research}}, \bibfield{editor}{\bibinfo{person}{Jennifer Dy} {and}
  \bibinfo{person}{Andreas Krause}} (Eds.), Vol.~\bibinfo{volume}{80}.
  \bibinfo{publisher}{PMLR}, \bibinfo{address}{Stockholmsmässan, Stockholm
  Sweden}, \bibinfo{pages}{2630--2639}.
\newblock
\urldef\tempurl%
\url{http://proceedings.mlr.press/v80/kilbertus18a.html}
\showURL{%
\tempurl}


\bibitem[\protect\citeauthoryear{Kilbertus, Rojas-Carulla, Parascandolo, Hardt,
  Janzing, and Sch\"{o}lkopf}{Kilbertus et~al\mbox{.}}{2017}]%
        {kilbertus_avoiding}
\bibfield{author}{\bibinfo{person}{Niki Kilbertus}, \bibinfo{person}{Mateo
  Rojas-Carulla}, \bibinfo{person}{Giambattista Parascandolo},
  \bibinfo{person}{Moritz Hardt}, \bibinfo{person}{Dominik Janzing}, {and}
  \bibinfo{person}{Bernhard Sch\"{o}lkopf}.} \bibinfo{year}{2017}\natexlab{}.
\newblock \showarticletitle{Avoiding Discrimination through Causal Reasoning}.
  In \bibinfo{booktitle}{\emph{Proceedings of the 31st International Conference
  on Neural Information Processing Systems}}
  \emph{(\bibinfo{series}{NIPS'17})}. \bibinfo{publisher}{Curran Associates
  Inc.}, \bibinfo{address}{Long Beach, California, USA},
  \bibinfo{pages}{656--666}.
\newblock
\showISBNx{9781510860964}


\bibitem[\protect\citeauthoryear{Kleinberg, Mullainathan, and
  Raghavan}{Kleinberg et~al\mbox{.}}{2016}]%
        {Kleinberg2016InherentTI}
\bibfield{author}{\bibinfo{person}{Jon~M. Kleinberg}, \bibinfo{person}{Sendhil
  Mullainathan}, {and} \bibinfo{person}{Manish Raghavan}.}
  \bibinfo{year}{2016}\natexlab{}.
\newblock \bibinfo{title}{Inherent Trade-Offs in the Fair Determination of Risk
  Scores}.
\newblock \bibinfo{howpublished}{ArXiv}.
\newblock


\bibitem[\protect\citeauthoryear{Kleinman}{Kleinman}{2017}]%
        {less_effective1}
\bibfield{author}{\bibinfo{person}{Gabe Kleinman}.}
  \bibinfo{year}{2017}\natexlab{}.
\newblock \bibinfo{title}{Down With Bias, Up With Profitability}.
\newblock
  \bibinfo{howpublished}{\url{https://worldpositive.com/down-with-bias-up-with-profitability-b221e7fee0ad}}.
\newblock
\newblock
\shownote{[Online; accessed 13-Septmeber-2020].}


\bibitem[\protect\citeauthoryear{Koh and Liang}{Koh and Liang}{2017}]%
        {koh_understanding_2017}
\bibfield{author}{\bibinfo{person}{Pang~Wei Koh} {and} \bibinfo{person}{Percy
  Liang}.} \bibinfo{year}{2017}\natexlab{}.
\newblock \showarticletitle{Understanding Black-Box Predictions via Influence
  Functions}. In \bibinfo{booktitle}{\emph{Proceedings of the 34th
  International Conference on Machine Learning - Volume 70}}
  \emph{(\bibinfo{series}{ICML'17})}. \bibinfo{publisher}{JMLR.org},
  \bibinfo{address}{Sydney, NSW, Australia}, \bibinfo{pages}{1885--1894}.
\newblock


\bibitem[\protect\citeauthoryear{Kopf}{Kopf}{2019}]%
        {criticism_1}
\bibfield{author}{\bibinfo{person}{Dan Kopf}.} \bibinfo{year}{2019}\natexlab{}.
\newblock \bibinfo{title}{{Goldman} {Sachs'} misguided {World} Cup
  {predictions} could provide clues to the {Apple} {Card} controversy}.
\newblock
  \bibinfo{howpublished}{\url{https://qz.com/1748321/the-role-of-goldman-sachs-algorithms-in-the-apple-credit-card-scandal/}}.
\newblock
\newblock
\shownote{[Online; accessed 13-Septmeber-2020].}


\bibitem[\protect\citeauthoryear{Kusner, Loftus, Russell, and Silva}{Kusner
  et~al\mbox{.}}{2017}]%
        {kusner_counterfactual}
\bibfield{author}{\bibinfo{person}{Matt Kusner}, \bibinfo{person}{Joshua
  Loftus}, \bibinfo{person}{Chris Russell}, {and} \bibinfo{person}{Ricardo
  Silva}.} \bibinfo{year}{2017}\natexlab{}.
\newblock \showarticletitle{Counterfactual Fairness}. In
  \bibinfo{booktitle}{\emph{Proceedings of the 31st International Conference on
  Neural Information Processing Systems}} (Long Beach, California, USA)
  \emph{(\bibinfo{series}{NIPS'17})}. \bibinfo{publisher}{Curran Associates
  Inc.}, \bibinfo{address}{Red Hook, NY, USA}, \bibinfo{pages}{4069--4079}.
\newblock
\showISBNx{9781510860964}


\bibitem[\protect\citeauthoryear{Lahoti, Gummadi, and Weikum}{Lahoti
  et~al\mbox{.}}{2019}]%
        {Lahoti2019_ops_IF}
\bibfield{author}{\bibinfo{person}{Preethi Lahoti}, \bibinfo{person}{Krishna~P.
  Gummadi}, {and} \bibinfo{person}{Gerhard Weikum}.}
  \bibinfo{year}{2019}\natexlab{}.
\newblock \showarticletitle{Operationalizing Individual Fairness with Pairwise
  Fair Representations}.
\newblock \bibinfo{journal}{\emph{Proc. VLDB Endow.}} \bibinfo{volume}{13},
  \bibinfo{number}{4} (\bibinfo{date}{Dec.} \bibinfo{year}{2019}),
  \bibinfo{pages}{506--518}.
\newblock
\showISSN{2150-8097}
\urldef\tempurl%
\url{https://doi.org/10.14778/3372716.3372723}
\showDOI{\tempurl}


\bibitem[\protect\citeauthoryear{Leben}{Leben}{2020}]%
        {group_fairness:2020}
\bibfield{author}{\bibinfo{person}{Derek Leben}.}
  \bibinfo{year}{2020}\natexlab{}.
\newblock \showarticletitle{Normative Principles for Evaluating Fairness in
  Machine Learning}. In \bibinfo{booktitle}{\emph{Proceedings of the AAAI/ACM
  Conference on AI, Ethics, and Society}} (New York, NY, USA)
  \emph{(\bibinfo{series}{AIES '20})}. \bibinfo{publisher}{Association for
  Computing Machinery}, \bibinfo{address}{New York, NY, USA},
  \bibinfo{pages}{86--92}.
\newblock
\showISBNx{9781450371100}
\urldef\tempurl%
\url{https://doi.org/10.1145/3375627.3375808}
\showDOI{\tempurl}


\bibitem[\protect\citeauthoryear{Li, Swersky, and Zemel}{Li
  et~al\mbox{.}}{2014}]%
        {Li2014LearningUF}
\bibfield{author}{\bibinfo{person}{Yujia Li}, \bibinfo{person}{Kevin Swersky},
  {and} \bibinfo{person}{Richard~S. Zemel}.} \bibinfo{year}{2014}\natexlab{}.
\newblock \bibinfo{title}{Learning unbiased features}.
\newblock \bibinfo{howpublished}{ArXiv}.
\newblock
\newblock
\shownote{abs/1412.5244.}


\bibitem[\protect\citeauthoryear{Lipton, Chouldechova, and McAuley}{Lipton
  et~al\mbox{.}}{2018}]%
        {Lipton2018DoesMM}
\bibfield{author}{\bibinfo{person}{Zachary~C. Lipton},
  \bibinfo{person}{Alexandra Chouldechova}, {and} \bibinfo{person}{Julian
  McAuley}.} \bibinfo{year}{2018}\natexlab{}.
\newblock \showarticletitle{Does Mitigating ML's Impact Disparity Require
  Treatment Disparity?}. In \bibinfo{booktitle}{\emph{Proceedings of the 32nd
  International Conference on Neural Information Processing Systems}}
  \emph{(\bibinfo{series}{NIPS'18})}. \bibinfo{publisher}{Curran Associates
  Inc.}, \bibinfo{address}{Montr\'{e}al, Canada}, \bibinfo{pages}{8136--8146}.
\newblock


\bibitem[\protect\citeauthoryear{Louizos, Swersky, Li, Welling, and
  Zemel}{Louizos et~al\mbox{.}}{2016}]%
        {Louizos2015TheVF}
\bibfield{author}{\bibinfo{person}{Christos Louizos}, \bibinfo{person}{Kevin
  Swersky}, \bibinfo{person}{Yujia Li}, \bibinfo{person}{Max Welling}, {and}
  \bibinfo{person}{Richard~S. Zemel}.} \bibinfo{year}{2016}\natexlab{}.
\newblock \showarticletitle{The Variational Fair Autoencoder}, In
  \bibinfo{booktitle}{ICLR}.
\newblock \bibinfo{journal}{\emph{CoRR}}, 11.
\newblock
\urldef\tempurl%
\url{http://arxiv.org/abs/1511.00830}
\showURL{%
\tempurl}
\newblock
\shownote{abs/1511.00830.}


\bibitem[\protect\citeauthoryear{Lum and Johndrow}{Lum and Johndrow}{2016}]%
        {Lum2016ASF}
\bibfield{author}{\bibinfo{person}{Kristian Lum} {and}
  \bibinfo{person}{James~E. Johndrow}.} \bibinfo{year}{2016}\natexlab{}.
\newblock \bibinfo{title}{A statistical framework for fair predictive
  algorithms}.
\newblock \bibinfo{howpublished}{ArXiv}.
\newblock
\newblock
\shownote{abs/1610.08077.}


\bibitem[\protect\citeauthoryear{Luong, Ruggieri, and Turini}{Luong
  et~al\mbox{.}}{2011}]%
        {Luong-knn-2011}
\bibfield{author}{\bibinfo{person}{Binh~Thanh Luong},
  \bibinfo{person}{Salvatore Ruggieri}, {and} \bibinfo{person}{Franco Turini}.}
  \bibinfo{year}{2011}\natexlab{}.
\newblock \showarticletitle{K-NN as an Implementation of Situation Testing for
  Discrimination Discovery and Prevention}. In
  \bibinfo{booktitle}{\emph{Proceedings of the 17th ACM SIGKDD International
  Conference on Knowledge Discovery and Data Mining}} (San Diego, California,
  USA) \emph{(\bibinfo{series}{KDD '11})}. \bibinfo{publisher}{Association for
  Computing Machinery}, \bibinfo{address}{New York, NY, USA},
  \bibinfo{pages}{502--510}.
\newblock
\showISBNx{9781450308137}
\urldef\tempurl%
\url{https://doi.org/10.1145/2020408.2020488}
\showDOI{\tempurl}


\bibitem[\protect\citeauthoryear{Madhavan and Wadhwa}{Madhavan and
  Wadhwa}{2020}]%
        {sensitive_removal4}
\bibfield{author}{\bibinfo{person}{Ramanujam Madhavan} {and}
  \bibinfo{person}{Mohit Wadhwa}.} \bibinfo{year}{2020}\natexlab{}.
\newblock \bibinfo{title}{Fairness-Aware Learning with Prejudice Free
  Representations}.
\newblock \bibinfo{howpublished}{ArXiv}.
\newblock
\newblock
\shownote{abs/2002.12143.}


\bibitem[\protect\citeauthoryear{Madras, Creager, Pitassi, and Zemel}{Madras
  et~al\mbox{.}}{2019}]%
        {causal_1:2019}
\bibfield{author}{\bibinfo{person}{David Madras}, \bibinfo{person}{Elliot
  Creager}, \bibinfo{person}{Toniann Pitassi}, {and} \bibinfo{person}{Richard
  Zemel}.} \bibinfo{year}{2019}\natexlab{}.
\newblock \showarticletitle{Fairness through Causal Awareness: Learning Causal
  Latent-Variable Models for Biased Data}. In
  \bibinfo{booktitle}{\emph{Proceedings of the Conference on Fairness,
  Accountability, and Transparency}} (Atlanta, GA, USA)
  \emph{(\bibinfo{series}{FAT* '19})}. \bibinfo{publisher}{Association for
  Computing Machinery}, \bibinfo{address}{New York, NY, USA},
  \bibinfo{pages}{349--358}.
\newblock
\showISBNx{9781450361255}
\urldef\tempurl%
\url{https://doi.org/10.1145/3287560.3287564}
\showDOI{\tempurl}


\bibitem[\protect\citeauthoryear{Martin}{Martin}{2018}]%
        {biased-hiring2}
\bibfield{author}{\bibinfo{person}{Nicole Martin}.}
  \bibinfo{year}{2018}\natexlab{}.
\newblock \bibinfo{title}{Are AI Hiring Programs Eliminating Bias Or Making It
  Worse?}
\newblock
  \bibinfo{howpublished}{\url{https://www.forbes.com/sites/nicolemartin1/2018/12/13/are-ai-hiring-programs-eliminating-bias-or-making-it-worse/}}.
\newblock
\newblock
\shownote{[Online; accessed 13-April-2020].}


\bibitem[\protect\citeauthoryear{McNamara, Ong, and Williamson}{McNamara
  et~al\mbox{.}}{2017}]%
        {McNamara2017ProvablyFR}
\bibfield{author}{\bibinfo{person}{Daniel McNamara},
  \bibinfo{person}{Cheng~Soon Ong}, {and} \bibinfo{person}{Robert~C.
  Williamson}.} \bibinfo{year}{2017}\natexlab{}.
\newblock \bibinfo{title}{Provably Fair Representations}.
\newblock \bibinfo{howpublished}{ArXiv}.
\newblock
\newblock
\shownote{abs/1710.04394.}


\bibitem[\protect\citeauthoryear{Mehrabi, Morstatter, Saxena, Lerman, and
  Galstyan}{Mehrabi et~al\mbox{.}}{2019}]%
        {mehrabi_survey_2019}
\bibfield{author}{\bibinfo{person}{Ninareh Mehrabi}, \bibinfo{person}{Fred
  Morstatter}, \bibinfo{person}{Nripsuta Saxena}, \bibinfo{person}{Kristina
  Lerman}, {and} \bibinfo{person}{Aram Galstyan}.}
  \bibinfo{year}{2019}\natexlab{}.
\newblock \bibinfo{title}{A {Survey} on {Bias} and {Fairness} in {Machine}
  {Learning}}.
\newblock \bibinfo{howpublished}{arXiv:1908.09635 [cs]}.
\newblock
\urldef\tempurl%
\url{http://arxiv.org/abs/1908.09635}
\showURL{%
\tempurl}


\bibitem[\protect\citeauthoryear{Menon and Williamson}{Menon and
  Williamson}{2018}]%
        {Menon-CF:2018}
\bibfield{author}{\bibinfo{person}{Aditya~Krishna Menon} {and}
  \bibinfo{person}{Robert~C Williamson}.} \bibinfo{year}{2018}\natexlab{}.
\newblock \showarticletitle{The cost of fairness in binary classification}. In
  \bibinfo{booktitle}{\emph{Proceedings of Machine Learning Research}},
  Vol.~\bibinfo{volume}{81}. \bibinfo{publisher}{PMLR}, \bibinfo{address}{New
  York, NY, USA}, \bibinfo{pages}{107--118}.
\newblock
\urldef\tempurl%
\url{http://proceedings.mlr.press/v81/menon18a.html}
\showURL{%
\tempurl}


\bibitem[\protect\citeauthoryear{Metevier, Giguere, Brockman, Kobren, Brun,
  Brunskill, and Thomas}{Metevier et~al\mbox{.}}{2019}]%
        {Metevier2019}
\bibfield{author}{\bibinfo{person}{Blossom Metevier}, \bibinfo{person}{Stephen
  Giguere}, \bibinfo{person}{Sarah Brockman}, \bibinfo{person}{Ari Kobren},
  \bibinfo{person}{Yuriy Brun}, \bibinfo{person}{Emma Brunskill}, {and}
  \bibinfo{person}{Philip~S. Thomas}.} \bibinfo{year}{2019}\natexlab{}.
\newblock \showarticletitle{Offline Contextual Bandits with High Probability
  Fairness Guarantees}.
\newblock In \bibinfo{booktitle}{\emph{Advances in Neural Information
  Processing Systems 32}}. \bibinfo{publisher}{Curran Associates, Inc.},
  \bibinfo{address}{Red Hook, NY, USA}, \bibinfo{pages}{14922--14933}.
\newblock
\urldef\tempurl%
\url{http://papers.nips.cc/paper/9630-offline-contextual-bandits-with-high-probability-fairness-guarantees.pdf}
\showURL{%
\tempurl}


\bibitem[\protect\citeauthoryear{Nabi and Shpitser}{Nabi and Shpitser}{2018}]%
        {Nabi2017FairIO}
\bibfield{author}{\bibinfo{person}{Razieh Nabi} {and} \bibinfo{person}{Ilya
  Shpitser}.} \bibinfo{year}{2018}\natexlab{}.
\newblock \showarticletitle{Fair Inference on Outcomes}. In
  \bibinfo{booktitle}{\emph{Proceedings of the Thirty-Second {AAAI} Conference
  on Artificial Intelligence}}. \bibinfo{publisher}{{AAAI} Press},
  \bibinfo{address}{New Orleans, Louisiana, USA}, \bibinfo{pages}{1931--1940}.
\newblock
\urldef\tempurl%
\url{https://www.aaai.org/ocs/index.php/AAAI/AAAI18/paper/view/16683}
\showURL{%
\tempurl}


\bibitem[\protect\citeauthoryear{Ncasas}{Ncasas}{2017}]%
        {blackbox2}
\bibfield{author}{\bibinfo{person}{Ncasas}.} \bibinfo{year}{2017}\natexlab{}.
\newblock \bibinfo{title}{Why are Machine Learning models called black boxes?}
\newblock
  \bibinfo{howpublished}{\url{https://datascience.stackexchange.com/questions/22335/why-are-machine-learning-models-called-black-boxes}}.
\newblock
\newblock
\shownote{[Online; accessed 13-April-2020].}


\bibitem[\protect\citeauthoryear{Pearl}{Pearl}{2009}]%
        {pearl2009causality}
\bibfield{author}{\bibinfo{person}{Judea Pearl}.}
  \bibinfo{year}{2009}\natexlab{}.
\newblock \bibinfo{booktitle}{\emph{Causality: Models, Reasoning and Inference}
  (\bibinfo{edition}{2nd} ed.)}.
\newblock \bibinfo{publisher}{Cambridge University Press},
  \bibinfo{address}{USA}.
\newblock
\showISBNx{052189560X}


\bibitem[\protect\citeauthoryear{Pedreschi, Ruggieri, and Turini}{Pedreschi
  et~al\mbox{.}}{2009}]%
        {Pedreschi-SS:2009}
\bibfield{author}{\bibinfo{person}{Dino Pedreschi}, \bibinfo{person}{Salvatore
  Ruggieri}, {and} \bibinfo{person}{Franco Turini}.}
  \bibinfo{year}{2009}\natexlab{}.
\newblock \showarticletitle{Measuring Discrimination in Socially-Sensitive
  Decision Records}. In \bibinfo{booktitle}{\emph{Proceedings of the 2009 SIAM
  International Conference on Data Mining}}. \bibinfo{publisher}{Society for
  Industrial and Applied Mathematics}, \bibinfo{address}{3600 University City
  Science Center Philadelphia, PA, United States}, \bibinfo{pages}{581--592}.
\newblock
\urldef\tempurl%
\url{https://doi.org/10.1137/1.9781611972795.50}
\showDOI{\tempurl}


\bibitem[\protect\citeauthoryear{Pedreshi, Ruggieri, and Turini}{Pedreshi
  et~al\mbox{.}}{2008}]%
        {Pedreshi-DA-2008}
\bibfield{author}{\bibinfo{person}{Dino Pedreshi}, \bibinfo{person}{Salvatore
  Ruggieri}, {and} \bibinfo{person}{Franco Turini}.}
  \bibinfo{year}{2008}\natexlab{}.
\newblock \showarticletitle{Discrimination-Aware Data Mining}. In
  \bibinfo{booktitle}{\emph{Proceedings of the 14th ACM SIGKDD International
  Conference on Knowledge Discovery and Data Mining}} (Las Vegas, Nevada, USA)
  \emph{(\bibinfo{series}{KDD '08})}. \bibinfo{publisher}{Association for
  Computing Machinery}, \bibinfo{address}{New York, NY, USA},
  \bibinfo{pages}{560--568}.
\newblock
\showISBNx{9781605581934}
\urldef\tempurl%
\url{https://doi.org/10.1145/1401890.1401959}
\showDOI{\tempurl}


\bibitem[\protect\citeauthoryear{Pleiss, Raghavan, Wu, Kleinberg, and
  Weinberger}{Pleiss et~al\mbox{.}}{2017}]%
        {Pleiss-FC:2017}
\bibfield{author}{\bibinfo{person}{Geoff Pleiss}, \bibinfo{person}{Manish
  Raghavan}, \bibinfo{person}{Felix Wu}, \bibinfo{person}{Jon Kleinberg}, {and}
  \bibinfo{person}{Kilian~Q. Weinberger}.} \bibinfo{year}{2017}\natexlab{}.
\newblock \showarticletitle{On Fairness and Calibration}. In
  \bibinfo{booktitle}{\emph{Proceedings of the 31st International Conference on
  Neural Information Processing Systems}} (Long Beach, California, USA)
  \emph{(\bibinfo{series}{NIPS'17})}. \bibinfo{publisher}{Curran Associates
  Inc.}, \bibinfo{address}{Red Hook, NY, USA}, \bibinfo{pages}{5684--5693}.
\newblock
\showISBNx{9781510860964}


\bibitem[\protect\citeauthoryear{PODESTA}{PODESTA}{2014}]%
        {white_house}
\bibfield{author}{\bibinfo{person}{JOHN PODESTA}.}
  \bibinfo{year}{2014}\natexlab{}.
\newblock \bibinfo{title}{BIG DATA: SEIZING OPPORTUNITIES, PRESERVING VALUES}.
\newblock \bibinfo{howpublished}{\url{https://perma.cc/ZXB4-SDL9}}.
\newblock
\newblock
\shownote{[Online; accessed 13-Septmeber-2020].}


\bibitem[\protect\citeauthoryear{Pope and Sydnor}{Pope and Sydnor}{2011}]%
        {sensitive_removal2}
\bibfield{author}{\bibinfo{person}{Devin~G. Pope} {and}
  \bibinfo{person}{Justin~R. Sydnor}.} \bibinfo{year}{2011}\natexlab{}.
\newblock \showarticletitle{Implementing Anti-Discrimination Policies in
  Statistical Profiling Models}.
\newblock \bibinfo{journal}{\emph{American Economic Journal: Economic Policy}}
  \bibinfo{volume}{3}, \bibinfo{number}{3} (\bibinfo{year}{2011}),
  \bibinfo{pages}{206--231}.
\newblock
\showISSN{19457731, 1945774X}
\urldef\tempurl%
\url{http://www.jstor.org/stable/41238108}
\showURL{%
\tempurl}


\bibitem[\protect\citeauthoryear{Resnick}{Resnick}{2019}]%
        {criticism_2}
\bibfield{author}{\bibinfo{person}{Brian Resnick}.}
  \bibinfo{year}{2019}\natexlab{}.
\newblock \bibinfo{title}{Yes, artificial intelligence can be racist}.
\newblock
  \bibinfo{howpublished}{\url{https://www.vox.com/science-and-health/2019/1/23/18194717/alexandria-ocasio-cortez-ai-bias}}.
\newblock
\newblock
\shownote{[Online; accessed 13-Septmeber-2020].}


\bibitem[\protect\citeauthoryear{Rodriguez}{Rodriguez}{2020}]%
        {sensitive_removal_data}
\bibfield{author}{\bibinfo{person}{German Rodriguez}.}
  \bibinfo{year}{2020}\natexlab{}.
\newblock \bibinfo{title}{Discrimination in Salaries}.
\newblock
  \bibinfo{howpublished}{\url{https://data.princeton.edu/wws509/datasets/\#salary}}.
\newblock
\newblock
\shownote{[Online; accessed 13-April-2020].}


\bibitem[\protect\citeauthoryear{Ruoss, Balunovic, Fischer, and Vechev}{Ruoss
  et~al\mbox{.}}{2020}]%
        {Ruoss2020LearningCI}
\bibfield{author}{\bibinfo{person}{Anian Ruoss}, \bibinfo{person}{Mislav
  Balunovic}, \bibinfo{person}{Marc Fischer}, {and} \bibinfo{person}{Martin~T.
  Vechev}.} \bibinfo{year}{2020}\natexlab{}.
\newblock \bibinfo{title}{Learning Certified Individually Fair
  Representations}.
\newblock \bibinfo{howpublished}{ArXiv}.
\newblock
\newblock
\shownote{abs/2002.10312.}


\bibitem[\protect\citeauthoryear{Russell, Kusner, Loftus, and Silva}{Russell
  et~al\mbox{.}}{2017a}]%
        {causal_2:2017}
\bibfield{author}{\bibinfo{person}{Chris Russell}, \bibinfo{person}{Matt~J
  Kusner}, \bibinfo{person}{Joshua Loftus}, {and} \bibinfo{person}{Ricardo
  Silva}.} \bibinfo{year}{2017}\natexlab{a}.
\newblock \showarticletitle{When Worlds Collide: Integrating Different
  Counterfactual Assumptions in Fairness}.
\newblock In \bibinfo{booktitle}{\emph{Advances in Neural Information
  Processing Systems 30}}, \bibfield{editor}{\bibinfo{person}{I.~Guyon},
  \bibinfo{person}{U.~V. Luxburg}, \bibinfo{person}{S.~Bengio},
  \bibinfo{person}{H.~Wallach}, \bibinfo{person}{R.~Fergus},
  \bibinfo{person}{S.~Vishwanathan}, {and} \bibinfo{person}{R.~Garnett}}
  (Eds.). \bibinfo{publisher}{Curran Associates, Inc.}, \bibinfo{address}{Red
  Hook, NY, USA}, \bibinfo{pages}{6414--6423}.
\newblock
\urldef\tempurl%
\url{http://papers.nips.cc/paper/7220-when-worlds-collide-integrating-different-counterfactual-assumptions-in-fairness.pdf}
\showURL{%
\tempurl}


\bibitem[\protect\citeauthoryear{Russell, Kusner, Loftus, and Silva}{Russell
  et~al\mbox{.}}{2017b}]%
        {Russell-WWC-2017}
\bibfield{author}{\bibinfo{person}{Chris Russell}, \bibinfo{person}{Matt~J.
  Kusner}, \bibinfo{person}{Joshua~R. Loftus}, {and} \bibinfo{person}{Ricardo
  Silva}.} \bibinfo{year}{2017}\natexlab{b}.
\newblock \showarticletitle{When Worlds Collide: Integrating Different
  Counterfactual Assumptions in Fairness}. In
  \bibinfo{booktitle}{\emph{Proceedings of the 31st International Conference on
  Neural Information Processing Systems}} (Long Beach, California, USA)
  \emph{(\bibinfo{series}{NIPS'17})}. \bibinfo{publisher}{Curran Associates
  Inc.}, \bibinfo{address}{Red Hook, NY, USA}, \bibinfo{pages}{6417--6426}.
\newblock
\showISBNx{9781510860964}


\bibitem[\protect\citeauthoryear{Salimi, Rodriguez, Howe, and Suciu}{Salimi
  et~al\mbox{.}}{2019}]%
        {salimi_capuchin:_2019}
\bibfield{author}{\bibinfo{person}{Babak Salimi}, \bibinfo{person}{Luke
  Rodriguez}, \bibinfo{person}{Bill Howe}, {and} \bibinfo{person}{Dan Suciu}.}
  \bibinfo{year}{2019}\natexlab{}.
\newblock \showarticletitle{Interventional Fairness: Causal Database Repair for
  Algorithmic Fairness}. In \bibinfo{booktitle}{\emph{Proceedings of the 2019
  International Conference on Management of Data}} (Amsterdam, Netherlands)
  \emph{(\bibinfo{series}{SIGMOD '19})}. \bibinfo{publisher}{Association for
  Computing Machinery}, \bibinfo{address}{New York, NY, USA},
  \bibinfo{pages}{793--810}.
\newblock
\showISBNx{9781450356435}
\urldef\tempurl%
\url{https://doi.org/10.1145/3299869.3319901}
\showDOI{\tempurl}


\bibitem[\protect\citeauthoryear{Selbst}{Selbst}{2017}]%
        {Selbst2017DisparateII}
\bibfield{author}{\bibinfo{person}{Andrew~D. Selbst}.}
  \bibinfo{year}{2017}\natexlab{}.
\newblock \showarticletitle{Disparate Impact in Big Data Policing}.
\newblock \bibinfo{journal}{\emph{Georgia law review}}  \bibinfo{volume}{52}
  (\bibinfo{year}{2017}), \bibinfo{pages}{3373}.
\newblock


\bibitem[\protect\citeauthoryear{Sharifi-Malvajerdi, Kearns, and
  Roth}{Sharifi-Malvajerdi et~al\mbox{.}}{2019}]%
        {avg_IF}
\bibfield{author}{\bibinfo{person}{Saeed Sharifi-Malvajerdi},
  \bibinfo{person}{Michael Kearns}, {and} \bibinfo{person}{Aaron Roth}.}
  \bibinfo{year}{2019}\natexlab{}.
\newblock \showarticletitle{Average Individual Fairness: Algorithms,
  Generalization and Experiments}.
\newblock In \bibinfo{booktitle}{\emph{Advances in Neural Information
  Processing Systems 32}}. \bibinfo{publisher}{Curran Associates, Inc.},
  \bibinfo{address}{Red Hook, NY, USA}, \bibinfo{pages}{8242--8251}.
\newblock
\urldef\tempurl%
\url{http://papers.nips.cc/paper/9034-average-individual-fairness-algorithms-generalization-and-experiments.pdf}
\showURL{%
\tempurl}


\bibitem[\protect\citeauthoryear{Simoiu, Corbett-Davies, and Goel}{Simoiu
  et~al\mbox{.}}{2017}]%
        {Simoiu2016}
\bibfield{author}{\bibinfo{person}{Camelia Simoiu}, \bibinfo{person}{Sam
  Corbett-Davies}, {and} \bibinfo{person}{Sharad Goel}.}
  \bibinfo{year}{2017}\natexlab{}.
\newblock \showarticletitle{The problem of infra-marginality in outcome tests
  for discrimination}.
\newblock \bibinfo{journal}{\emph{Annals of Applied Statistics}}
  \bibinfo{volume}{11}, \bibinfo{number}{3} (\bibinfo{date}{09}
  \bibinfo{year}{2017}), \bibinfo{pages}{1193--1216}.
\newblock
\urldef\tempurl%
\url{https://doi.org/10.1214/17-AOAS1058}
\showDOI{\tempurl}


\bibitem[\protect\citeauthoryear{Skeem and Lowenkamp}{Skeem and
  Lowenkamp}{2016}]%
        {Skeem2016RISKRA}
\bibfield{author}{\bibinfo{person}{Jennifer~Lynne Skeem} {and}
  \bibinfo{person}{Christopher~T. Lowenkamp}.} \bibinfo{year}{2016}\natexlab{}.
\newblock \showarticletitle{RISK, RACE, AND RECIDIVISM: PREDICTIVE BIAS AND
  DISPARATE IMPACT*: RISK, RACE, AND RECIDIVISM}.
\newblock \bibinfo{journal}{\emph{Criminology}}  \bibinfo{volume}{54}
  (\bibinfo{year}{2016}), \bibinfo{pages}{680--712}.
\newblock


\bibitem[\protect\citeauthoryear{Steil, Albright, Rugh, and Massey}{Steil
  et~al\mbox{.}}{2017}]%
        {Steil:2017-redlining}
\bibfield{author}{\bibinfo{person}{Justin Steil}, \bibinfo{person}{Len
  Albright}, \bibinfo{person}{Jacob Rugh}, {and} \bibinfo{person}{Douglas
  Massey}.} \bibinfo{year}{2017}\natexlab{}.
\newblock \showarticletitle{The Social Structure of Mortgage Discrimination}.
\newblock \bibinfo{journal}{\emph{Housing Studies}}  \bibinfo{volume}{33}
  (\bibinfo{date}{11} \bibinfo{year}{2017}), \bibinfo{pages}{1--18}.
\newblock
\urldef\tempurl%
\url{https://doi.org/10.1080/02673037.2017.1390076}
\showDOI{\tempurl}


\bibitem[\protect\citeauthoryear{Techlabs}{Techlabs}{2020}]%
        {finance_1}
\bibfield{author}{\bibinfo{person}{Maruti Techlabs}.}
  \bibinfo{year}{2020}\natexlab{}.
\newblock \bibinfo{title}{12 Use Cases of AI and Machine Learning In Finance}.
\newblock
  \bibinfo{howpublished}{\url{https://marutitech.com/ai-and-ml-in-finance/}}.
\newblock
\newblock
\shownote{[Online; accessed 13-Septmeber-2020].}


\bibitem[\protect\citeauthoryear{Thomas, Castro~da Silva, Barto, Giguere, Brun,
  and Brunskill}{Thomas et~al\mbox{.}}{2019}]%
        {Thomas2019}
\bibfield{author}{\bibinfo{person}{Philip~S. Thomas}, \bibinfo{person}{Bruno
  Castro~da Silva}, \bibinfo{person}{Andrew~G. Barto}, \bibinfo{person}{Stephen
  Giguere}, \bibinfo{person}{Yuriy Brun}, {and} \bibinfo{person}{Emma
  Brunskill}.} \bibinfo{year}{2019}\natexlab{}.
\newblock \showarticletitle{Preventing undesirable behavior of intelligent
  machines}.
\newblock \bibinfo{journal}{\emph{Science}} \bibinfo{volume}{366},
  \bibinfo{number}{6468} (\bibinfo{year}{2019}), \bibinfo{pages}{999--1004}.
\newblock
\showISSN{0036-8075}
\urldef\tempurl%
\url{https://doi.org/10.1126/science.aag3311}
\showDOI{\tempurl}
\showeprint{https://science.sciencemag.org/content/366/6468/999.full.pdf}


\bibitem[\protect\citeauthoryear{Ting}{Ting}{2017}]%
        {Confusion-Matrix}
\bibfield{author}{\bibinfo{person}{Kai~Ming Ting}.}
  \bibinfo{year}{2017}\natexlab{}.
\newblock \bibinfo{booktitle}{\emph{Confusion Matrix}}.
\newblock \bibinfo{publisher}{Springer US}, \bibinfo{address}{Boston, MA},
  \bibinfo{pages}{260--260}.
\newblock
\showISBNx{978-1-4899-7687-1}
\urldef\tempurl%
\url{https://doi.org/10.1007/978-1-4899-7687-1_50}
\showDOI{\tempurl}


\bibitem[\protect\citeauthoryear{Udeshi, Arora, and Chattopadhyay}{Udeshi
  et~al\mbox{.}}{2018}]%
        {Udeshi-testing:2018}
\bibfield{author}{\bibinfo{person}{Sakshi Udeshi}, \bibinfo{person}{Pryanshu
  Arora}, {and} \bibinfo{person}{Sudipta Chattopadhyay}.}
  \bibinfo{year}{2018}\natexlab{}.
\newblock \showarticletitle{Automated Directed Fairness Testing}. In
  \bibinfo{booktitle}{\emph{Proceedings of the 33rd ACM/IEEE International
  Conference on Automated Software Engineering}} (Montpellier, France)
  \emph{(\bibinfo{series}{ASE 2018})}. \bibinfo{publisher}{Association for
  Computing Machinery}, \bibinfo{address}{New York, NY, USA},
  \bibinfo{pages}{98--108}.
\newblock
\showISBNx{9781450359375}
\urldef\tempurl%
\url{https://doi.org/10.1145/3238147.3238165}
\showDOI{\tempurl}


\bibitem[\protect\citeauthoryear{Veale and Binns}{Veale and Binns}{2017a}]%
        {removingSensitive2}
\bibfield{author}{\bibinfo{person}{Michael Veale} {and} \bibinfo{person}{Reuben
  Binns}.} \bibinfo{year}{2017}\natexlab{a}.
\newblock \showarticletitle{Fairer machine learning in the real world:
  Mitigating discrimination without collecting sensitive data}.
\newblock \bibinfo{journal}{\emph{Big Data \& Society}} \bibinfo{volume}{4},
  \bibinfo{number}{2} (\bibinfo{year}{2017}),
  \bibinfo{pages}{2053951717743530}.
\newblock
\urldef\tempurl%
\url{https://doi.org/10.1177/2053951717743530}
\showDOI{\tempurl}


\bibitem[\protect\citeauthoryear{Veale and Binns}{Veale and Binns}{2017b}]%
        {sensitive_removal5}
\bibfield{author}{\bibinfo{person}{Michael Veale} {and} \bibinfo{person}{Reuben
  Binns}.} \bibinfo{year}{2017}\natexlab{b}.
\newblock \showarticletitle{Fairer machine learning in the real world:
  Mitigating discrimination without collecting sensitive data}.
\newblock \bibinfo{journal}{\emph{Big Data \& Society}}  \bibinfo{volume}{4}
  (\bibinfo{year}{2017}), 17.
\newblock


\bibitem[\protect\citeauthoryear{Verma and Rubin}{Verma and Rubin}{2018}]%
        {verma_fairness_2018}
\bibfield{author}{\bibinfo{person}{Sahil Verma} {and} \bibinfo{person}{Julia
  Rubin}.} \bibinfo{year}{2018}\natexlab{}.
\newblock \showarticletitle{Fairness definitions explained}. In
  \bibinfo{booktitle}{\emph{Proceedings of the {International} {Workshop} on
  {Software} {Fairness} - {FairWare} '18}}. \bibinfo{publisher}{ACM Press},
  \bibinfo{address}{Gothenburg, Sweden}, \bibinfo{pages}{1--7}.
\newblock
\showISBNx{978-1-4503-5746-3}
\urldef\tempurl%
\url{https://doi.org/10.1145/3194770.3194776}
\showDOI{\tempurl}


\bibitem[\protect\citeauthoryear{\v{Z}liobaite and Custers}{\v{Z}liobaite and
  Custers}{2016}]%
        {sensitive_removal1}
\bibfield{author}{\bibinfo{person}{Indre; \v{Z}liobaite} {and}
  \bibinfo{person}{Bart Custers}.} \bibinfo{year}{2016}\natexlab{}.
\newblock \showarticletitle{Using Sensitive Personal Data May Be Necessary for
  Avoiding Discrimination in Data-Driven Decision Models}.
\newblock \bibinfo{journal}{\emph{Artif. Intell. Law}} \bibinfo{volume}{24},
  \bibinfo{number}{2} (\bibinfo{date}{June} \bibinfo{year}{2016}),
  \bibinfo{pages}{183--201}.
\newblock
\showISSN{0924-8463}
\urldef\tempurl%
\url{https://doi.org/10.1007/s10506-016-9182-5}
\showDOI{\tempurl}


\bibitem[\protect\citeauthoryear{Wadsworth, Vera, and Piech}{Wadsworth
  et~al\mbox{.}}{2018}]%
        {wadsworth_achieving_2018}
\bibfield{author}{\bibinfo{person}{Christina Wadsworth},
  \bibinfo{person}{Francesca Vera}, {and} \bibinfo{person}{Chris Piech}.}
  \bibinfo{year}{2018}\natexlab{}.
\newblock \bibinfo{title}{Achieving Fairness through Adversarial Learning: an
  Application to Recidivism Prediction}.
\newblock \bibinfo{howpublished}{arXiv:1807.00199 [cs, stat]}. ,
  \bibinfo{numpages}{5}~pages.
\newblock
\urldef\tempurl%
\url{http://arxiv.org/abs/1807.00199}
\showURL{%
\tempurl}
\newblock
\shownote{arXiv: 1807.00199.}


\bibitem[\protect\citeauthoryear{Wick, panda, and Tristan}{Wick
  et~al\mbox{.}}{2019}]%
        {trade_off_2019}
\bibfield{author}{\bibinfo{person}{Michael Wick}, \bibinfo{person}{swetasudha
  panda}, {and} \bibinfo{person}{Jean-Baptiste Tristan}.}
  \bibinfo{year}{2019}\natexlab{}.
\newblock \showarticletitle{Unlocking Fairness: a Trade-off Revisited}.
\newblock In \bibinfo{booktitle}{\emph{Advances in Neural Information
  Processing Systems 32}}. \bibinfo{publisher}{Curran Associates, Inc.},
  \bibinfo{address}{Red Hook, NY, USA}, \bibinfo{pages}{8783--8792}.
\newblock
\urldef\tempurl%
\url{http://papers.nips.cc/paper/9082-unlocking-fairness-a-trade-off-revisited.pdf}
\showURL{%
\tempurl}


\bibitem[\protect\citeauthoryear{Wieringa}{Wieringa}{2020}]%
        {survey_accountability:2020}
\bibfield{author}{\bibinfo{person}{Maranke Wieringa}.}
  \bibinfo{year}{2020}\natexlab{}.
\newblock \showarticletitle{What to Account for When Accounting for Algorithms:
  A Systematic Literature Review on Algorithmic Accountability}. In
  \bibinfo{booktitle}{\emph{Proceedings of the 2020 Conference on Fairness,
  Accountability, and Transparency}} (Barcelona, Spain)
  \emph{(\bibinfo{series}{FAT* '20})}. \bibinfo{publisher}{Association for
  Computing Machinery}, \bibinfo{address}{New York, NY, USA},
  \bibinfo{pages}{1--18}.
\newblock
\showISBNx{9781450369367}
\urldef\tempurl%
\url{https://doi.org/10.1145/3351095.3372833}
\showDOI{\tempurl}


\bibitem[\protect\citeauthoryear{{Wikipedia contributors}}{{Wikipedia
  contributors}}{2020}]%
        {redliningwiki}
\bibfield{author}{\bibinfo{person}{{Wikipedia contributors}}.}
  \bibinfo{year}{2020}\natexlab{}.
\newblock \bibinfo{title}{Redlining --- {Wikipedia}{,} The Free Encyclopedia}.
\newblock
  \bibinfo{howpublished}{\url{https://en.wikipedia.org/w/index.php?title=Redlining}}.
\newblock
\newblock
\shownote{[Online; accessed 13-April-2020].}


\bibitem[\protect\citeauthoryear{Williams, Brooks, and Shmargad}{Williams
  et~al\mbox{.}}{2018a}]%
        {removingSensitive1}
\bibfield{author}{\bibinfo{person}{Betsy~Anne Williams},
  \bibinfo{person}{Catherine~F. Brooks}, {and} \bibinfo{person}{Yotam
  Shmargad}.} \bibinfo{year}{2018}\natexlab{a}.
\newblock \showarticletitle{How Algorithms Discriminate Based on Data They
  Lack: Challenges, Solutions, and Policy Implications}.
\newblock \bibinfo{journal}{\emph{Journal of Information Policy}}
  \bibinfo{volume}{8} (\bibinfo{year}{2018}), \bibinfo{pages}{78--115}.
\newblock
\showISSN{23815892, 21583897}
\urldef\tempurl%
\url{https://www.jstor.org/stable/10.5325/jinfopoli.8.2018.0078}
\showURL{%
\tempurl}


\bibitem[\protect\citeauthoryear{Williams, Brooks, and Shmargad}{Williams
  et~al\mbox{.}}{2018b}]%
        {sensitive_removal3}
\bibfield{author}{\bibinfo{person}{Betsy~Anne Williams},
  \bibinfo{person}{Catherine~F. Brooks}, {and} \bibinfo{person}{Yotam
  Shmargad}.} \bibinfo{year}{2018}\natexlab{b}.
\newblock \showarticletitle{How Algorithms Discriminate Based on Data They
  Lack: Challenges, Solutions, and Policy Implications}.
\newblock \bibinfo{journal}{\emph{Journal of Information Policy}}
  \bibinfo{volume}{8} (\bibinfo{year}{2018}), \bibinfo{pages}{78--115}.
\newblock


\bibitem[\protect\citeauthoryear{Winrow and Schieber}{Winrow and
  Schieber}{2010}]%
        {Winrow2010TheDB}
\bibfield{author}{\bibinfo{person}{Brian~P. Winrow} {and}
  \bibinfo{person}{Christen Schieber}.} \bibinfo{year}{2010}\natexlab{}.
\newblock \showarticletitle{The Disparity between Disparate Treatment and
  Disparate Impact: An Analysis of the Ricci Case}, In
  \bibinfo{booktitle}{Journal of Legal, Ethical and Regulatory Issues}.
\newblock \bibinfo{journal}{\emph{The DreamCatchers Group, LLC}}
  \bibinfo{volume}{13}, 14.
\newblock


\bibitem[\protect\citeauthoryear{Woodworth, Gunasekar, Ohannessian, and
  Srebro}{Woodworth et~al\mbox{.}}{2017}]%
        {Woodworth2017LearningNP}
\bibfield{author}{\bibinfo{person}{Blake Woodworth}, \bibinfo{person}{Suriya
  Gunasekar}, \bibinfo{person}{Mesrob~I. Ohannessian}, {and}
  \bibinfo{person}{Nathan Srebro}.} \bibinfo{year}{2017}\natexlab{}.
\newblock \showarticletitle{Learning Non-Discriminatory Predictors}. In
  \bibinfo{booktitle}{\emph{Proceedings of Machine Learning Research}},
  Vol.~\bibinfo{volume}{65}. \bibinfo{publisher}{PMLR},
  \bibinfo{address}{Amsterdam, Netherlands}, \bibinfo{pages}{1920--1953}.
\newblock
\urldef\tempurl%
\url{http://proceedings.mlr.press/v65/woodworth17a.html}
\showURL{%
\tempurl}


\bibitem[\protect\citeauthoryear{Yang, Qinami, Fei-Fei, Deng, and
  Russakovsky}{Yang et~al\mbox{.}}{2020}]%
        {ImageNet:2020}
\bibfield{author}{\bibinfo{person}{Kaiyu Yang}, \bibinfo{person}{Klint Qinami},
  \bibinfo{person}{Li Fei-Fei}, \bibinfo{person}{Jia Deng}, {and}
  \bibinfo{person}{Olga Russakovsky}.} \bibinfo{year}{2020}\natexlab{}.
\newblock \showarticletitle{Towards Fairer Datasets: Filtering and Balancing
  the Distribution of the People Subtree in the ImageNet Hierarchy}. In
  \bibinfo{booktitle}{\emph{Proceedings of the 2020 Conference on Fairness,
  Accountability, and Transparency}} (Barcelona, Spain)
  \emph{(\bibinfo{series}{FAT* '20})}. \bibinfo{publisher}{Association for
  Computing Machinery}, \bibinfo{address}{New York, NY, USA},
  \bibinfo{pages}{547--558}.
\newblock
\showISBNx{9781450369367}
\urldef\tempurl%
\url{https://doi.org/10.1145/3351095.3375709}
\showDOI{\tempurl}


\bibitem[\protect\citeauthoryear{Zafar, Valera, Gomez~Rodriguez, and
  Gummadi}{Zafar et~al\mbox{.}}{2017a}]%
        {Zafar2017FairnessBD}
\bibfield{author}{\bibinfo{person}{Muhammad~Bilal Zafar},
  \bibinfo{person}{Isabel Valera}, \bibinfo{person}{Manuel Gomez~Rodriguez},
  {and} \bibinfo{person}{Krishna~P. Gummadi}.}
  \bibinfo{year}{2017}\natexlab{a}.
\newblock \showarticletitle{Fairness Beyond Disparate Treatment \& Disparate
  Impact: Learning Classification without Disparate Mistreatment}. In
  \bibinfo{booktitle}{\emph{Proceedings of the 26th International Conference on
  World Wide Web}} (Perth, Australia) \emph{(\bibinfo{series}{WWW '17})}.
  \bibinfo{publisher}{International World Wide Web Conferences Steering
  Committee}, \bibinfo{address}{Republic and Canton of Geneva, CHE},
  \bibinfo{pages}{1171--1180}.
\newblock
\showISBNx{9781450349130}
\urldef\tempurl%
\url{https://doi.org/10.1145/3038912.3052660}
\showDOI{\tempurl}


\bibitem[\protect\citeauthoryear{Zafar, Valera, Gomez{-}Rodriguez, and
  Gummadi}{Zafar et~al\mbox{.}}{2017b}]%
        {Zafar2017FairnessCM}
\bibfield{author}{\bibinfo{person}{Muhammad~Bilal Zafar},
  \bibinfo{person}{Isabel Valera}, \bibinfo{person}{Manuel Gomez{-}Rodriguez},
  {and} \bibinfo{person}{Krishna~P. Gummadi}.}
  \bibinfo{year}{2017}\natexlab{b}.
\newblock \showarticletitle{Fairness Constraints: Mechanisms for Fair
  Classification}. In \bibinfo{booktitle}{\emph{Proceedings of the 20th
  International Conference on Artificial Intelligence and Statistics,
  {AISTATS}}} \emph{(\bibinfo{series}{Proceedings of Machine Learning
  Research}, Vol.~\bibinfo{volume}{54})}. \bibinfo{publisher}{PMLR},
  \bibinfo{address}{Fort Lauderdale, FL, {USA}}, \bibinfo{pages}{962--970}.
\newblock
\urldef\tempurl%
\url{http://proceedings.mlr.press/v54/zafar17a.html}
\showURL{%
\tempurl}


\bibitem[\protect\citeauthoryear{Zemel, Wu, Swersky, Pitassi, and Dwork}{Zemel
  et~al\mbox{.}}{2013}]%
        {zemel_learning}
\bibfield{author}{\bibinfo{person}{Richard Zemel}, \bibinfo{person}{Yu Wu},
  \bibinfo{person}{Kevin Swersky}, \bibinfo{person}{Toniann Pitassi}, {and}
  \bibinfo{person}{Cynthia Dwork}.} \bibinfo{year}{2013}\natexlab{}.
\newblock \showarticletitle{Learning Fair Representations}. In
  \bibinfo{booktitle}{\emph{Proceedings of the 30th International Conference on
  International Conference on Machine Learning - Volume 28}}
  \emph{(\bibinfo{series}{ICML'13})}. \bibinfo{publisher}{JMLR.org},
  \bibinfo{address}{Atlanta, GA, USA}, \bibinfo{pages}{III--325--III--333}.
\newblock


\bibitem[\protect\citeauthoryear{Zhang, Lemoine, and Mitchell}{Zhang
  et~al\mbox{.}}{2018}]%
        {Zhang2018MitigatingUB}
\bibfield{author}{\bibinfo{person}{Brian~Hu Zhang}, \bibinfo{person}{Blake
  Lemoine}, {and} \bibinfo{person}{Margaret Mitchell}.}
  \bibinfo{year}{2018}\natexlab{}.
\newblock \showarticletitle{Mitigating Unwanted Biases with Adversarial
  Learning}. In \bibinfo{booktitle}{\emph{Proceedings of the 2018 AAAI/ACM
  Conference on AI, Ethics, and Society}} (New Orleans, LA, USA)
  \emph{(\bibinfo{series}{AIES '18})}. \bibinfo{publisher}{Association for
  Computing Machinery}, \bibinfo{address}{New York, NY, USA},
  \bibinfo{pages}{335--340}.
\newblock
\showISBNx{9781450360128}
\urldef\tempurl%
\url{https://doi.org/10.1145/3278721.3278779}
\showDOI{\tempurl}


\end{thebibliography}

\appendix
\section*{Appendix}
\label{sec:appendix}


\section{Experimental plots}
\label{sec:extraplots}

\begin{figure*}[h!]
    \includegraphics[width=\textwidth]{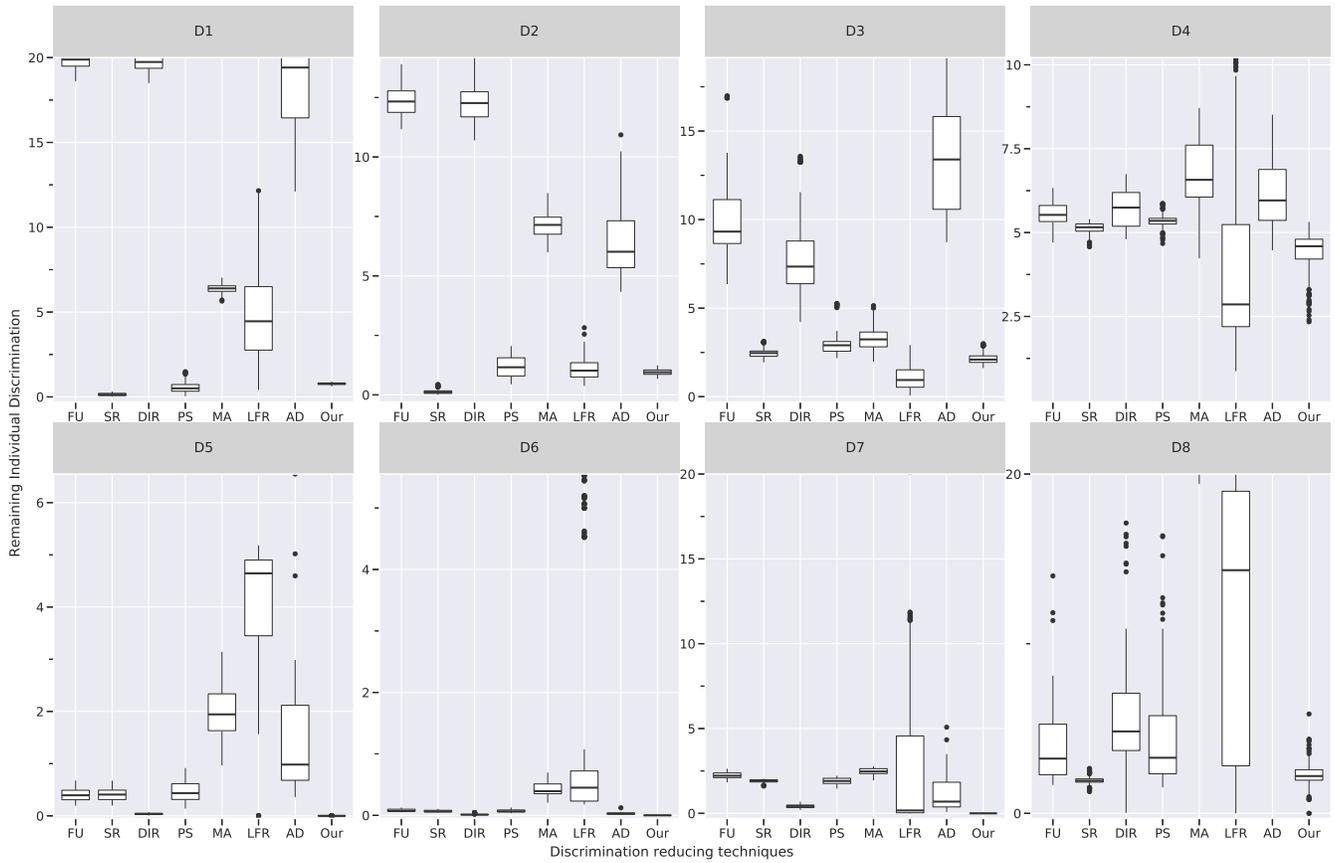}
    \vspace{-10pt}
    \caption{The individual discrimination for all 240 hyperparameter choices (lower
    is better).
    Our approach (rightmost in each boxplot) achieves low discrimination for many
    hyperparameter choices, and it has a little variance across choices for most datasets.
    }
    \label{fig:facet_boxplot_discm2}
\end{figure*}

\begin{figure*}[h!]
    \includegraphics[width=\textwidth]{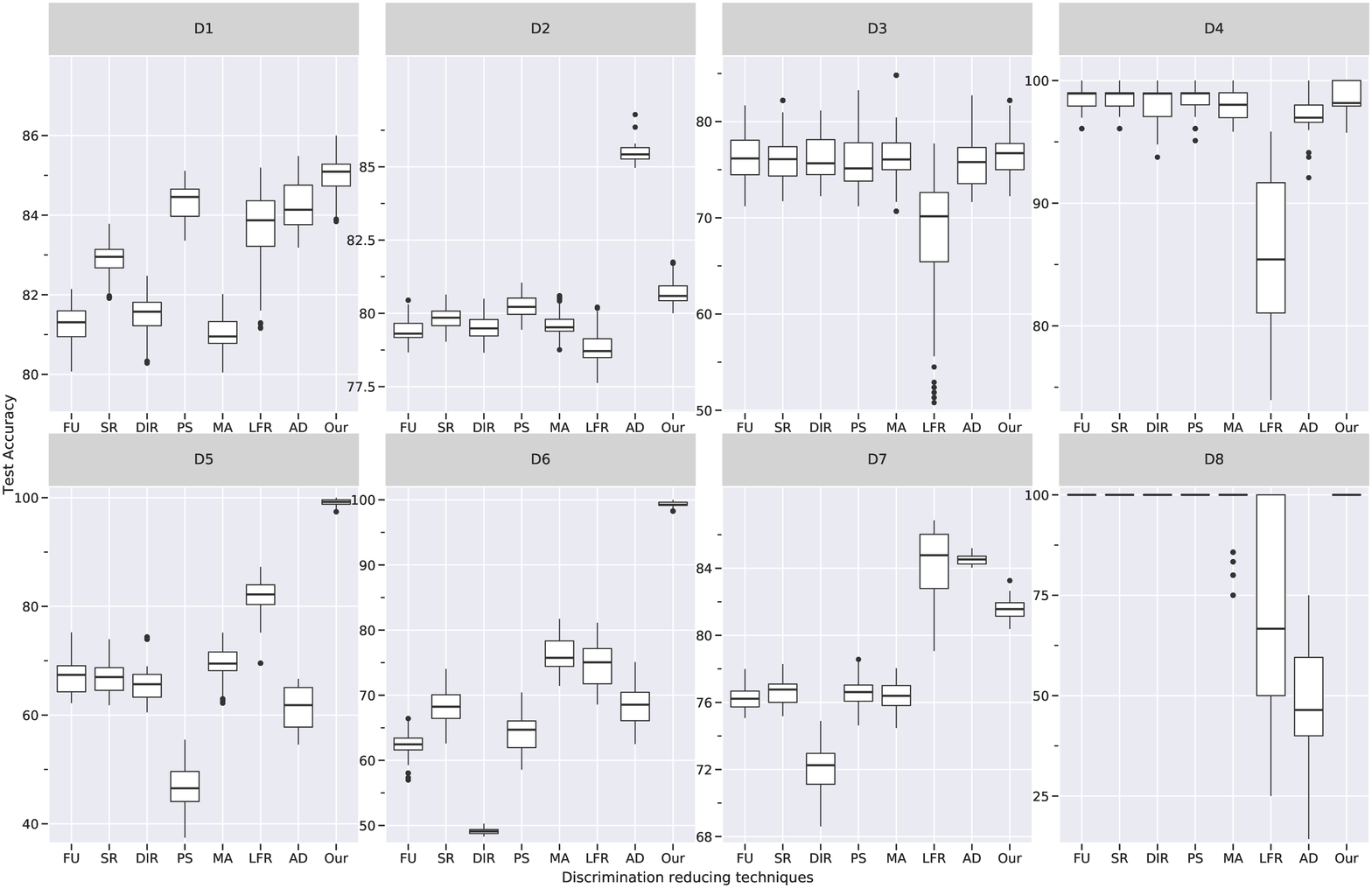}
    \vspace{-10pt}
    \caption{The test accuracy for all 240 hyperparameter choices (higher is
    better).
    Our approach is best or comparable to the best in terms of both accuracy
    and its variance, for all experiments except D2 and D7. \looseness=-1}
   \label{fig:facet_boxplot_accuracy2}
\end{figure*}

\end{document}